%% file: paper.tex
\documentclass{bmvc2k}

\input{math_commands.tex}

\usepackage{enumitem}
\usepackage{algorithm,algpseudocode}
\usepackage{graphicx}
\usepackage{multirow}

\newcommand{\minisection}[1]{\vspace{0mm}\noindent{\textbf{#1}}}

\title{A Probabilistic Hard Attention Model For Sequentially Observed Scenes}

\addauthor{Samrudhdhi B. Rangrej}{samrudhdhi.rangrej@mail.mcgill.ca}{1}
\addauthor{James J. Clark}{james.j.clark@mcgill.ca}{1}

\addinstitution{
 Centre for Intelligent Machines (CIM) \\
 McGill University \\
 Montreal, Quebec, Canada
}

\runninghead{Rangrej, Clark}{A Probabilistic Hard Attention Model}

\begin{document}

\maketitle

\vspace{-0.25cm}
\begin{abstract}
A visual hard attention model actively selects and observes a sequence of subregions in an image to make a prediction. The majority of hard attention models determine the attention-worthy regions by first analyzing a complete image. However, it may be the case that the entire image is not available initially but instead sensed gradually through a series of partial observations. In this paper, we design an efficient hard attention model for classifying such sequentially observed scenes. The presented model never observes an image completely. To select informative regions under partial observability, the model uses Bayesian Optimal Experiment Design. First, it synthesizes the features of the unobserved regions based on the already observed regions. Then, it uses the predicted features to estimate the expected information gain (EIG) attained, should various regions be attended. Finally, the model attends to the actual content on the location where the EIG mentioned above is maximum. The model uses a) a recurrent feature aggregator to maintain a recurrent state, b)  a linear classifier to predict the class label, c) a Partial variational autoencoder to predict the features of unobserved regions. We use normalizing flows in Partial VAE to handle multi-modality in the feature-synthesis problem. We train our model using a differentiable objective and test it on five datasets. Our model gains 2-10\% higher accuracy than the baseline models when both have seen only a couple of glimpses. code: \url{https://github.com/samrudhdhirangrej/Probabilistic-Hard-Attention}.
\end{abstract}

\vspace{-0.25cm}
\section{Introduction}
A deep feedforward network observes an entire scene to achieve state-of-the-art performance. However, observing an entire scene may not be necessary as many times the task-relevant information lies only in a few parts of the scene. A visual hard attention technique allows a recognition system to attend to only the most informative subregions called \textit{glimpses} \cite{mnih2014recurrent,xu2015show}. A hard attention model builds a representation of a scene by sequentially acquiring useful glimpses and fusing the collected information. The representation guides recognition and future glimpse acquisition. Hard attention is useful to reduce data acquisition cost \cite{uzkent2020learning}, to develop interpretable \cite{elsayed2019saccader}, computationally efficient \cite{katharopoulos2019processing} and scalable \cite{papadopoulos2021hard} models.

The majority of hard attention models first analyze a complete image, occasionally at low resolution, to locate the task-relevant subregions \cite{ ba2014multiple, elsayed2019saccader}. However, in practice, we often do not have access to the entire scene. Instead, we only observe parts of the scene as we attend to them. We decide the next attention-worthy location at each step in the process based on the partial observations collected so far. Examples include a self-driving car navigating in unknown territory, time-sensitive aerial imagery for a rescue operation, etc. Here, we develop a hard-attention model for such a scenario. While the majority of existing hard attention models pose glimpse acquisition as a reinforcement learning task to optimize a non-differentiable model \cite{mnih2014recurrent, xu2015show}, we use a fully differentiable model and training objective. We train the model using a combination of discriminative and generative objectives. The former is used for class-label prediction, and the latter is used for the content prediction for unobserved regions. We use the predicted content to find an optimal attention-worthy location as described next.

A hard attention model predicts the class label of an image by attending various informative glimpses in an image. We formulate the problem of finding optimal glimpse-locations as Bayesian Optimal Experiment Design (BOED). Starting from a random location, a sequential model uses BOED to determine the next optimal location. To do so, it estimates the expected information gain (EIG) obtained from observing glimpses at yet unobserved regions of the scene and selects a location with maximum EIG. As the computation of EIG requires the content of regions, the model \textit{synthesizes} the unknown content conditioned on the content of the observed regions. For efficiency reasons, the model predicts the content in the feature space instead of the pixel space. Our model consists of three modules, a recurrent feature aggregator, a linear classifier, and a Partial VAE \cite{ma2018eddi}. Partial VAE synthesizes features of various glimpses in the scene based on partial observations. There may exist multiple possibilities for the content of the unobserved regions, given the content of the observed regions. Hence, we use normalizing flows in Partial VAE to capture the multi-modality in the posterior. Our main contributions are as follows.
\begin{itemize}[noitemsep,topsep=0pt,leftmargin=*]
    \item We develop a hard attention model to classify images using a series of partial observations. The model estimates EIG of the yet unobserved locations and attends a location with maximum EIG.
    \item To estimate EIG of unobserved regions, we synthesize the content of these regions. Furthermore, to improve the efficiency of a model, we synthesize the content in the feature space and use normalizing flows to capture the multi-modality in the problem.
    \item We support our approach with a principled mathematical framework. We implement the model using a compact network and perform experiments on five datasets. Our model achieves 2-10\% higher accuracy compared to the baseline methods when both have seen only a couple of glimpses.
\end{itemize}

\section{Related Works}
\minisection{Hard Attention.}
A hard attention model prioritizes task-relevant regions to extract meaningful features from an input. Early attempts to model attention employed image saliency as a priority map. High priority regions were selected using methods such as winner-take-all \cite{koch1987shifts,itti1998model, itti2000saliency}, searching by throwing out all features but the one with minimal activity \cite{ahmad1992visit}, and dynamic routing of information \cite{olshausen1993neurobiological}.
Few used graphical models to model visual attention. \citet{rimey1991controlling} used augmented hidden Markov models to model attention policy. \citet{larochelle2010learning} used a Restricted Boltzmann Machine (RBM) with third-order connections between attention location, glimpse, and the representation of a scene. Motivated by this, \citet{zheng2015neural} proposed an autoregressive model to compute exact gradients, unlike in an RBM. \citet{tang2014learning} used an RBM as a generative model and searched for informative locations using the Hamiltonian Monte Carlo algorithm.
Many used reinforcement learning to train attention models. \citet{paletta2005q} used Q-learning with the reward that measures the objectness of the attended region. \citet{denil2012learning} estimated rewards using particle filters and employed a policy based on the Gaussian Process and the upper confidence bound. \citet{butko2008pomdp} modeled attention as a partially observable Markov decision process and used a policy gradient algorithm for learning. Later, \citet{butko2009optimal} extended this approach to multiple objects.

Recently, the machine learning community uses the REINFORCE policy gradient algorithm to train non-differentiable hard attention models \cite{mnih2014recurrent, ba2014multiple, xu2015show, elsayed2019saccader, papadopoulos2021hard}. Other recent works use EM-style learning procedure \cite{ranzato2014learning}, wake-sleep algorithm \cite{ba2015learning}, a voting based region selection \cite{alexe2012searching}, and spatial transformer \cite{jaderberg2015spatial, gregor2015draw, eslami2016attend}. Among the recent methods, \cite{ba2014multiple, ranzato2014learning,ba2015learning} look at the low-resolution gist of an input at the beginning, and \cite{xu2015show, elsayed2019saccader, eslami2016attend, gregor2015draw} consume the whole image to predict the locations to attend. In contrast, our model does not look at the entire image at low resolution or otherwise. Moreover, our model is fully differentiable.

\minisection{Image Completion.}
Image completion methods aim at synthesizing unobserved or missing image pixels conditioned on the observed pixels \cite{yu2018generative,pathak2016context,iizuka2017globally,wang2014biggerpicture,yang2019very,wang2019wide}. Image completion is an ill-posed problem with multiple possible solutions for the missing image regions. While early methods use deterministic models, \cite{zhang2020multimodal,cai2020piigan,zhao2020uctgan} used stochastic models to predict multiple instances of a complete image. Many used probabilistic models to ensure that the completions are generated according to their actual probability \cite{zheng2019pluralistic, sohn2015learning, ma2018eddi, garnelo2018neural, garnelo2018conditional}. We use Partial VAE \cite{ma2018eddi} -- a probabilistic model -- to predict the content of the complete image given only a few glimpses. However, unlike other approaches that infer image content in the pixel space, we predict content in the feature space.

\minisection{Patch Selection.}
Many computer vision methods select attention-worthy regions from an image, e.g., region proposal network \cite{ren2015faster}, multi-instance learning \cite{ilse2018attention}, top-$K$ patch selection \cite{angles2018mist, cordonnier2021differentiable}, attention sampling \cite{katharopoulos2019processing}. These approaches observe an entire image and find all attention-worthy patches simultaneously. Our model observes an image only partially and sequentially. Furthermore, it selects a single optimal patch to attend at a given time.

\vspace{-0.2cm}
\section{Model}
\label{sec:model}
In this paper, we consider a recurrent attention model that sequentially captures glimpses from an image $x$ and predicts a label $y$. The model runs for time $t=0$ to $T-1$. It uses a recurrent net to maintain a hidden state $h_{t-1}$ that summarizes glimpses observed until time $t-1$. At time $t$, it predicts coordinates $l_t$ based on the hidden state $h_{t-1}$ and captures a square glimpse $g_t$ centered at $l_t$ in an image $x$, i.e. $g_t = g(x,l_t)$. It uses $g_t$ and $l_t$ to update the hidden state to $h_t$ and predicts the label $y$ based on the updated state $h_t$.

\vspace{-0.2cm}
\subsection{Building Blocks}
\label{sec:architecture}
\begin{figure}
    \centering
    \includegraphics[width=0.8\textwidth]{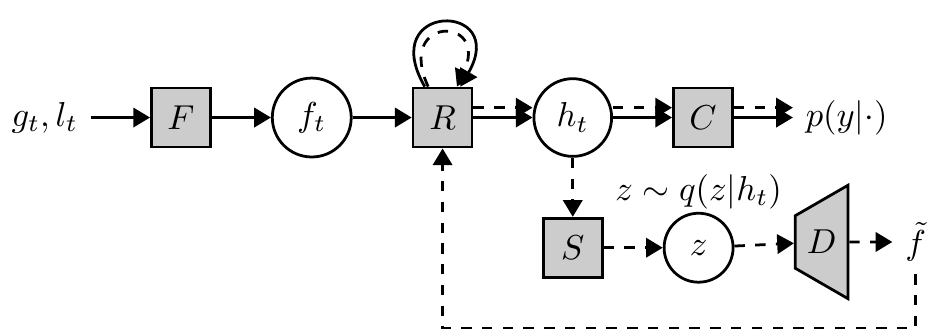}
    \caption{A recurrent attention model sequentially observes glimpses from an image and predicts a class label. At time $t$, the model actively observes a glimpse $g_t$ and its coordinates $l_t$. Given $g_t$ and $l_t$, the feed-forward module $F$ extracts features $f_t$, and the recurrent module $R$ updates a hidden state to $h_t$. Using an updated hidden state $h_t$, the linear classifier $C$ predicts the class distribution $p(y|h_t)$. At time $t+1$, the model assesses various candidate locations $l$ before attending an optimal one. It predicts $p(y|g,l,h_t)$ ahead of time and selects the candidate $l$ that maximizes $\KL[p(y|g,l,h_t)||p(y|h_t)]$. The model synthesizes the features of $g$ using a Partial VAE to approximate $p(y|g,l,h_t)$ without attending to the glimpse $g$. The normalizing flow-based encoder $S$ predicts the approximate posterior $q(z|h_t)$. The decoder $D$ uses a sample $z\sim q(z|h_t)$ to synthesize a feature map $\Tilde{f}$ containing features of all glimpses. The model uses $\Tilde{f}(l)$ as features of a glimpse at location $l$ and evaluates $p(y|g,l,h_t)\approx p(y|\Tilde{f}(l),h_t)$. Dashed arrows show a path to compute the lookahead class distribution $p(y|\Tilde{f}(l),h_t)$.}
    \label{fig:model}
    \vspace{-0.2cm}
\end{figure}

As shown in Figure \ref{fig:model}, the proposed model comprises the following three building blocks. A recurrent feature aggregator ($F$ and $R$) maintains a hidden state $h_t$. A classifier ($C$) predicts the class probabilities $p(y|h_t)$. A normalizing flow-based variational autoencoder ($S$ and $D$) synthesizes a feature map of a complete image given the hidden state $h_t$. Specifically, a flow-based encoder $S$ predicts the posterior of a latent variable $z$ from $h_t$, and a decoder $D$ uses $z$ to synthesize a feature map of a complete image. A feature map of a complete image can also be seen as a map containing features of all glimpses. The BOED, as discussed in section \ref{sec:boed}, uses the synthesized feature map to find an optimal location to attend at the next time step. To distinguish the synthesized feature map from an actual one, let us call the former $\Tilde{f}$ and the latter $f$. Henceforth, we crown any quantity derived from the synthesized feature map with a ( $\Tilde{}$ ). Next, we provide details about the three building blocks of the model, followed by a discussion of the BOED in the context of hard attention.\\

\minisection{Recurrent Feature Aggregator.}
Given a glimpse $g_t$ and its location $l_t$, a feed-forward module extracts features $f_t = F(g_t, l_t)$, and a recurrent network updates a hidden state to $h_t = R(h_{t-1}, f_t)$. We define $F(g,l)=\mathcal{F}_g(g) + \mathcal{F}_l(l)$ and $R(h,f) = LeakyReLU(\mathcal{F}_h(h) + \mathcal{F}_f(f))$. $\mathcal{F}_g$ is a small CNN with receptive-field equal to size of $g$. $\mathcal{F}_l, \mathcal{F}_h, \mathcal{F}_f$ are shallow networks with one linear layer.\\

\minisection{Classifier.}
At each time step $t$, a linear classifier predicts the distribution $p(y|h_t) = C(h_t)$ from a hidden state $h_t$. As the goal of the model is to predict a label $y$ for an image $x$, we learn a distribution $p(y|h_t)$ by minimizing $\KL[p(y|x)||p(y|h_t)]$. Optimization of this KL divergence is equivalent to the minimization of the following cross-entropy loss.
\begin{align}
    \mathcal{L}_{\text{CE}}(t) = - p(y|x) \log(p(y|h_t))
    \label{eq:ce}
\end{align}

\minisection{Partial Variational Autoencoder.}
\label{sec:pvae}
We adapt a variational autoencoder (VAE) to synthesize the feature map of a complete image from the hidden state $h_t$. A VAE learns a joint distribution between the feature map $f$ and the latent variable $z$ given $h_t$, $p(f,z|h_t)=p(f|z)p(z|h_t)$. An encoder approximates the posterior $q(z|f,h_t)$, and a decoder infers the likelihood $p(f|z)$. The optimization of VAE requires calculation of $\KL[q(z|f,h_t)||p(z|h_t)]$ \cite{kingma2013auto}. As the hard attention model does not observe the complete image, it cannot estimate $q(z|f,h_t)$. Hence, we cannot incorporate the standard VAE directly into a hard attention framework and use the following approach.

At the time $t$, let's separate an image $x$ into two parts, $o_t$ --- the set of regions observed up to $t$, and $u_t$ --- the set of regions as yet unobserved. \citet{ma2018eddi} observed that in a VAE, $o_t$ and $u_t$ are conditionally independent given $z$, i.e. $p(x|z)=p(u_t|z) p(o_t|z)$. They synthesize $u_t$ independently from the sample $z\sim q(z|o_t)$, while learning the posterior $q(z|o_t)$ by optimizing ELBO on $\log(p(o_t))$. They refer to the resultant VAE as a Partial VAE.

The BOED, as discussed in section \ref{sec:boed}, requires features of $o_t$ and $u_t$. \citet{ma2018eddi} consider $o_t$ and $u_t$ in pixel space. Without the loss of generality, we consider $o_t$ and $u_t$ in the feature space. Synthesizing $o_t$ and $u_t$ in the feature space serves two purposes. First, the model does not have to extract features of $o_t$ and $u_t$ for the BOED as they are readily available. Second, Partial VAE does not have to produce unnecessary details that may later be thrown away by the feature extractor, such as the exact color of a pixel. Recall that the features $f_{1:t}$ and the hidden state $h_t$ calculated by our attention model correspond to the glimpses observed up to $t$, which is equivalent to $o_t$, the set of observed regions. Hence, we write $q(z|o_t)$ as $q(z|h_t)$ and $p(o_t|z)$ as $p(f_{1:t}|z)$ in the ELBO of Partial VAE.
\begin{align}
    \mathcal{L}_{\text{PVAE}}(t) = -ELBO &= -\Big\{ \mathbb{E}_{q(z|o_t)} \log(p(o_t|z)) - \KL[q(z|o_t)||p(z)]\Big\} \nonumber \\ &= -\Big\{\mathbb{E}_{q(z|h_t)} \log(p(f_{1:t}|z)) - \KL[q(z|h_t)||p(z)]\Big\}
    \label{eq:pvae}
\end{align}

In \eqref{eq:pvae}, the prior $p(z)$ is a Gaussian distribution with zero mean and unit variance. To obtain expressive posterior $q(z|h_t)$, we use normalizing flows in Partial VAE \cite{kingma2016improved}. Specifically, we use auto-regressive Neural Spline Flows (NSF) \cite{durkan2019neural}. Between the two flow layers, we flip the input \cite{dinh2016density} and normalize it using ActNorm \cite{kingma2018glow}. Refer to supp. material for brief overview of NSF. In Figure \ref{fig:model}, the flow-based encoder $S$ infers the posterior $q(z|h_t) = S(h_t)$.

In a Partial VAE, $p(f|z) = p(f_{1:t}|z) p(f^c_{1:t}|z)$; where $f^c_{1:t}$ are the features of the glimpses other than the ones observed up to $t$ and $f=f_{1:t} \cup f^c_{1:t}$. We implement a decoder $D$ that synthesizes a feature map containing features of all glimpses in an image given the sample $z\sim q(z|h_t)$, i.e. $\Tilde{f} = D(z)$. Let $m_t$ be a binary mask with value \textbf{1} for the glimpses observed by the model up to $t$ and \textbf{0} otherwise; hence, $f_{1:t} = m_t \odot f$, where $\odot$ is an element-wise multiplication. We assume a Gaussian likelihood and evaluate the log-likelihood in \eqref{eq:pvae} using the mask $m_t$ as follows.
\begin{align}
   - 
   \log(p(f_{1:t}|z)) \propto \frac{1}{2}\sum \frac{|m_t\odot \Tilde{f} - m_t\odot f|^2}{\sigma^2} + \log(\sigma^2) 
   \label{eq:xliklihood}
\end{align}
Where $\sigma$ is a model parameter. The BOED uses $\Tilde{f}$ to find an optimal location to attend.

\subsection{Bayesian Optimal Experiment Design (BOED)}
\label{sec:boed}
The BOED evaluates the optimality of a set of experiments by measuring information gain in the interest parameter due to the experimental outcome \cite{chaloner1995bayesian}. In the context of hard attention, an experiment is to attend a location $l$ and observe a corresponding glimpse $g=g(x,l)$. An experiment of attending a location $l$ is optimal if it gains maximum information about the class label $y$. We can evaluate optimality of attending a specific location by measuring several metrics such as feature variance \cite{huang2018active}, uncertainty in the prediction \cite{melville2004active}, expected Shannon information \cite{lindley1956measure}. For a sequential model, information gain $ \KL[p(y|g, l, h_{t-1})||p(y|h_{t-1})]$ is an ideal metric. It measures the change in the entropy of the class distribution from one time step to the next due to observation of a glimpse $g$ at location $l$ \cite{bernardo1979expected,ma2018eddi}.

The model has to find an optimal location to attend at time $t$ before observing the corresponding glimpse. Hence, we consider an expected information gain (EIG) over the generating distribution of $g$. An EIG is also a measure of Bayesian surprise \cite{itti2006bayesian,schwartenbeck2013exploration}.
\begin{align}
    EIG(l) &= \mathbb{E}_{p(g|l, h_{t-1})} \KL\big[p(y|g, l, h_{t-1})||p(y|h_{t-1})\big] \\ &= \mathbb{E}_{p(f(l)| h_{t-1})} \KL\big[p(y|f(l), h_{t-1})||p(y|h_{t-1})\big]
    \label{eq:EIG}
\end{align}

Where $f(l)$ are features of a glimpse located at $l$, i.e. $f(l)=F(g,l)$. Inspired by \cite{harvey2019near}, we define $p(f(l)| h_{t-1})$ as follows.
\begin{align}
    p(f(l)| h_{t-1}) &= \mathbb{E}_{q(z|h_{t-1})}  p(f(l)|z)  =  \mathbb{E}_{q(z|h_{t-1})} \delta(D(z)(l)) = \mathbb{E}_{q(z|h_{t-1})} \delta(\Tilde{f}(l))
    \label{eq:p_of_gt}
\end{align}
Here, $\delta(\cdot)$ is a delta distribution. As discussed in the section \ref{sec:pvae}, the flow-based encoder $S$ predicts the posterior $q(z|h_{t-1})$ and the decoder $D$ predicts the feature map containing features of all glimpses in an image, $\Tilde{f}=D(z)$. Combining \eqref{eq:EIG} and \eqref{eq:p_of_gt} yields,
\begin{align}
    EIG(l) &= \mathbb{E}_{q(z|h_{t-1})} \KL\big[p(y|\Tilde{f}(l), h_{t-1})||p(y|h_{t-1})\big] 
    \label{eq:EIG_x}
\end{align}
To find an optimal location to attend at time $t$, the model compares various candidates for $l_t$. It predicts $EIG(l)$ for each candidate $l$ and selects an optimal candidate as $l_t$, i.e. $l_t = \argmax_l EIG(l)$.
When the model is considering a candidate $l$, it uses $\Tilde{f}(l)$ to calculate $\Tilde{h}_t = R(h_{t-1}, \Tilde{f}(l))$ and $p(y|\Tilde{h}_{t}) = C(\Tilde{h}_t)$. It uses the distribution $p(y|\Tilde{h}_{t}) = p(y|\Tilde{f}(l), h_{t-1})$ to calculate $EIG$ in \eqref{eq:EIG_x}. We refer to $p(y|\Tilde{h}_{t})$ as the \textit{lookahead} class distribution computed by anticipating the content at the location $l$ ahead of time. In Figure \ref{fig:model}, the dashed arrows show a lookahead step.
Furthermore, to compute $EIG$ for all locations simultaneously, we implement all modules of our model with convolution layers. The model computes $EIG$ for all locations as a single activation map in a single forward pass. An optimal location is equal to the coordinates of the pixel with maximum value in the $EIG$ map.

\section{Experiments}
\minisection{Datasets.} We evaluate our model on SVHN \cite{netzer2011reading}, CINIC-10 \cite{darlow2018cinic}, CIFAR-10 \cite{krizhevsky2009learning}, CIFAR-100 \cite{krizhevsky2009learning}, and TinyImageNet \cite{tinyimagenet} datasets. These datasets consist of real-world images categorized into 10, 10, 10, 100 and 200 classes respectively. Images in TinyImageNet are of size $64\times64$ and images in the remaining dataset are of size $32\times32$.

\minisection{Training and Testing.}
We train our model in three phases. In the first phase, we pre-train modules $F$, $R$, and $C$ with a random sequence of $T$ glimpses using $\sum_{t=0}^{T-1} \mathcal{L}_{CE}(t)$ as a training objective. In the second phase, we introduce $S$ and $D$. We pre-train $S$ and $D$ while keeping $F$, $R$, $C$ frozen. Again, we use a random sequence of $T$ glimpses and train $S$ and $D$ using $\sum_{t=0}^{T-1} \mathcal{L}_{\text{PVAE}}(t)$ training criterion. To produce the target $f$ used in equation \ref{eq:xliklihood}, we feed a complete image and a grid of all locations to the pretrained $F$, which computes features of all glimpses as a single feature map. Pre-training $(F, R, C)$ and $(S,D)$ separately ensures that the Partial VAE receives a stable target feature map $f$ in equation \ref{eq:xliklihood}. Finally, in the third phase, we fine-tune all modules end-to-end using the training objective $\mathcal{L} = \sum_{t=0}^{T-1} \alpha \mathcal{L}_{\text{PVAE}}(t) + \beta \mathcal{L}_{CE}(t)$, where $\alpha$ and $\beta$ are hyperparameters. In the finetuning stage, we sample an optimal sequence of glimpses using the BOED framework. We use only one sample $z\sim q(z|h_t)$ to estimate the $EIG$ map during the training, which leads to exploration. In the test phase, we achieve exploitation by using $P$ samples of $z$ to estimate the $EIG$ accurately. The test procedure is shown in Algorithm \ref{algo:pred_all}. Refer to supp. material (SM) for more details.

\setlength{\textfloatsep}{5pt}
\begin{algorithm}[t]
  \caption{Test procedure for the model shown in Figure \ref{fig:model}}
  \begin{algorithmic}[1]
      \State Randomly sample $l_0$; Capture $g_0$ at $l_0$, compute $f_0$, $h_0$ and $p(y|h_0)$
      \For{$t \in \{1,\dots,T-1\}$}\Comment{T is the time budget}
        \State Sample $z_i \sim q(z|h_{t-1})$
        and predict $\Tilde{f}^{i}; i \in \{0,\dots,P-1\}$\Comment{P is the sample budget}
        \State Compute $\Tilde{h}_{t}^{i}$, $p(y|\Tilde{h}_t^i)$ and $EIG = \frac{1}{P} \sum_i \KL[p(y|\Tilde{h}_t^i)||p(y|h_{t-1})]$ \Comment{\eqref{eq:EIG_x}}
        \State $l_{t} = \argmax \{EIG\}$ 
        \State Capture $g_t$ at $l_t$; Compute $f_t$, $h_t$ and $p(y|h_t)$
      \EndFor
  \end{algorithmic}
  \label{algo:pred_all}
\end{algorithm}

\minisection{Hyperparameters.}  We implement our model with a small number of parameters, and it runs for $T=7$ time steps. It senses glimpses of size $16\times16$ overlapping with stride $n=8$ for TinyImageNet and senses glimpses of size $8\times8$ overlapping with stride $n=4$ for the remaining datasets. The Partial VAE predicts $\Tilde{f}$ and $EIG$ for a set of glimpses separated with stride equal to $n$. We do not allow our model to revisit glimpses attended in the past. We choose the hyperparameters $\alpha$ and $\beta$ such that the two loss-terms contribute equally. The sample budget $P$ is 20 for all experiments. Refer to SM for more details.

\subsection{Baseline Comparison}
\begin{figure}[!t]
    \centering
    \hfill
    \begin{minipage}[c]{0.32\linewidth}
    \includegraphics[width = \textwidth]{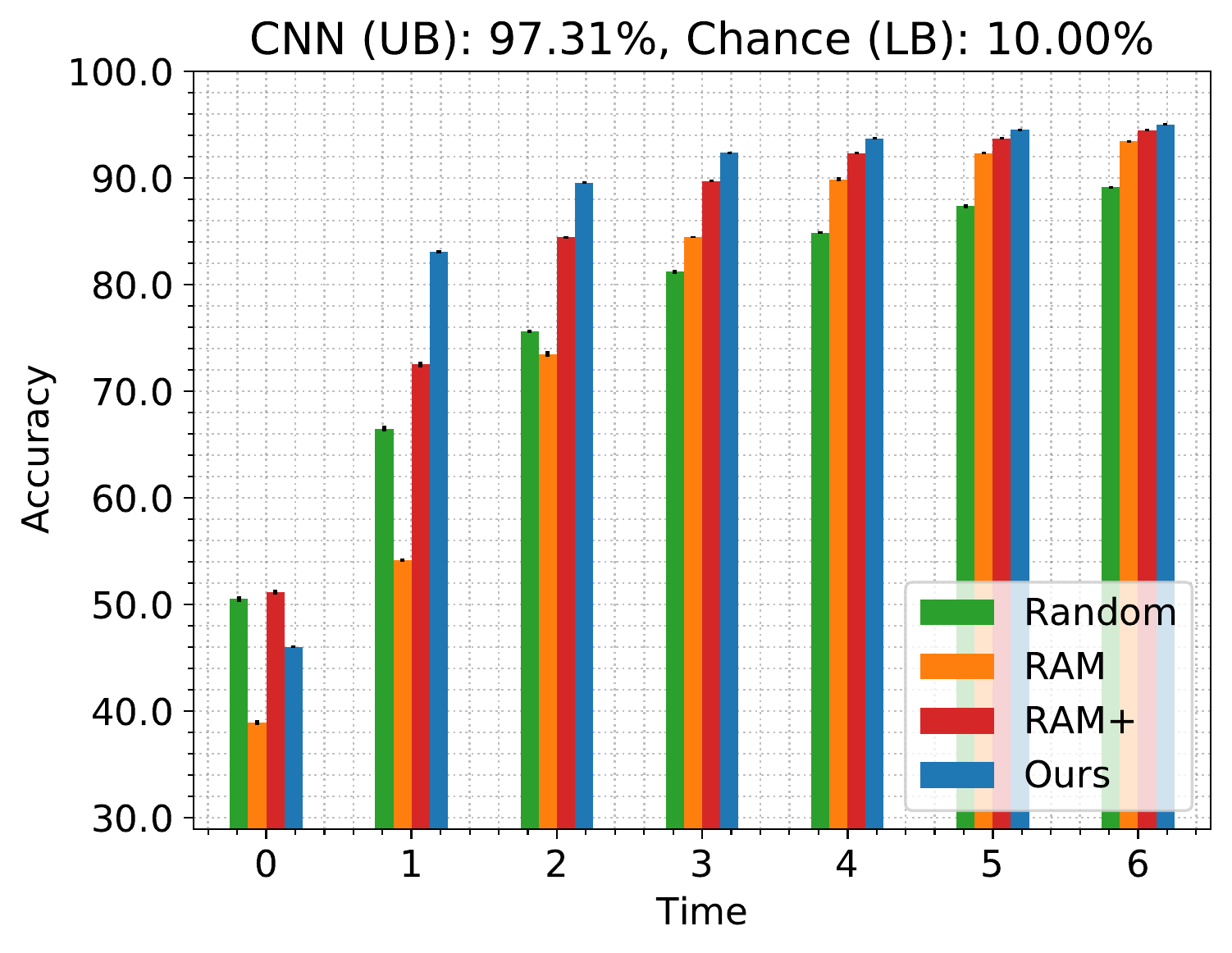}
    \centering{(a)}
    \end{minipage}
    \hfill
    \begin{minipage}[c]{0.32\linewidth}
    \includegraphics[width = \textwidth]{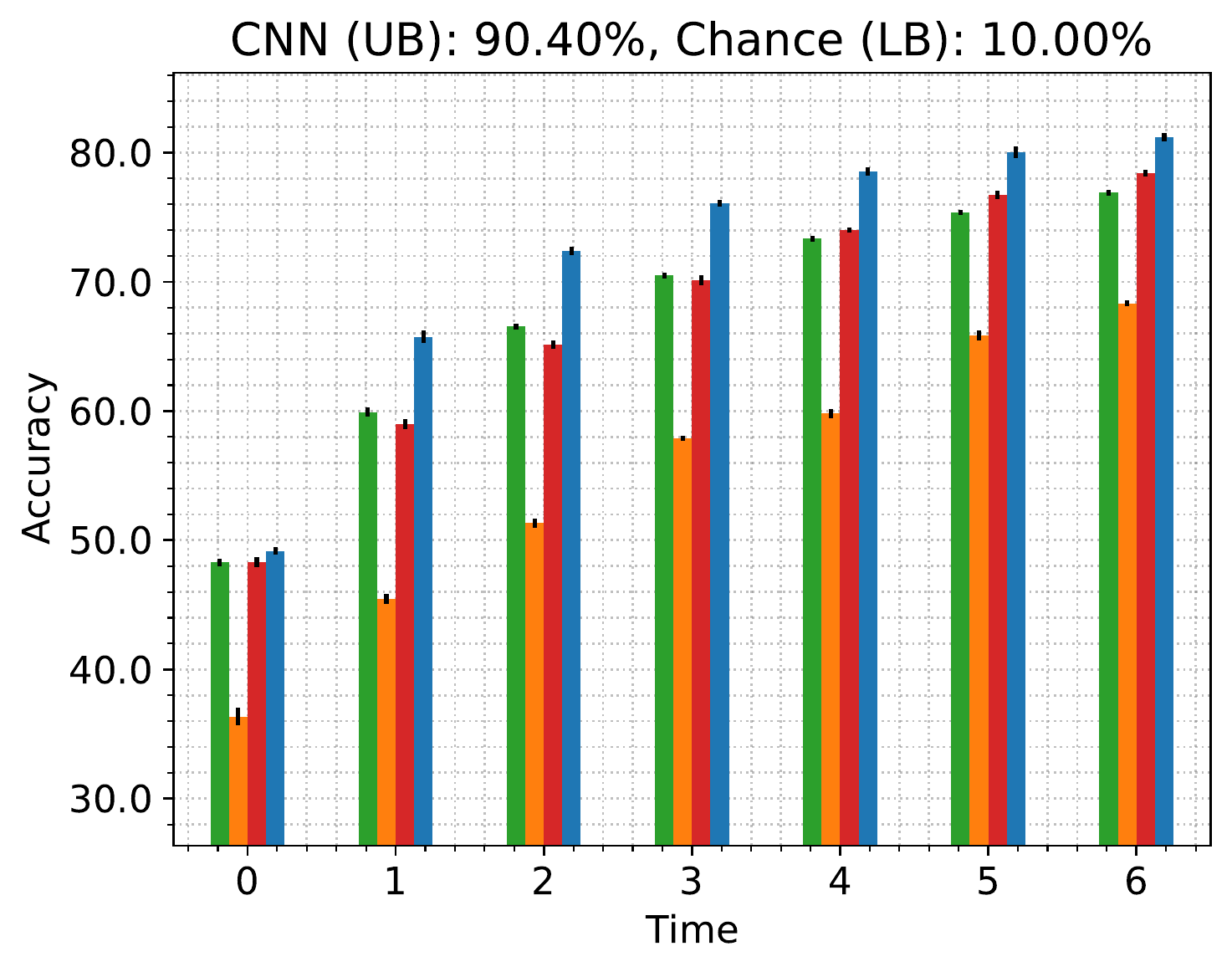}
    \centering{(b)}
    \end{minipage}
    \hfill
    \begin{minipage}[c]{0.32\linewidth}
    \includegraphics[width = \textwidth]{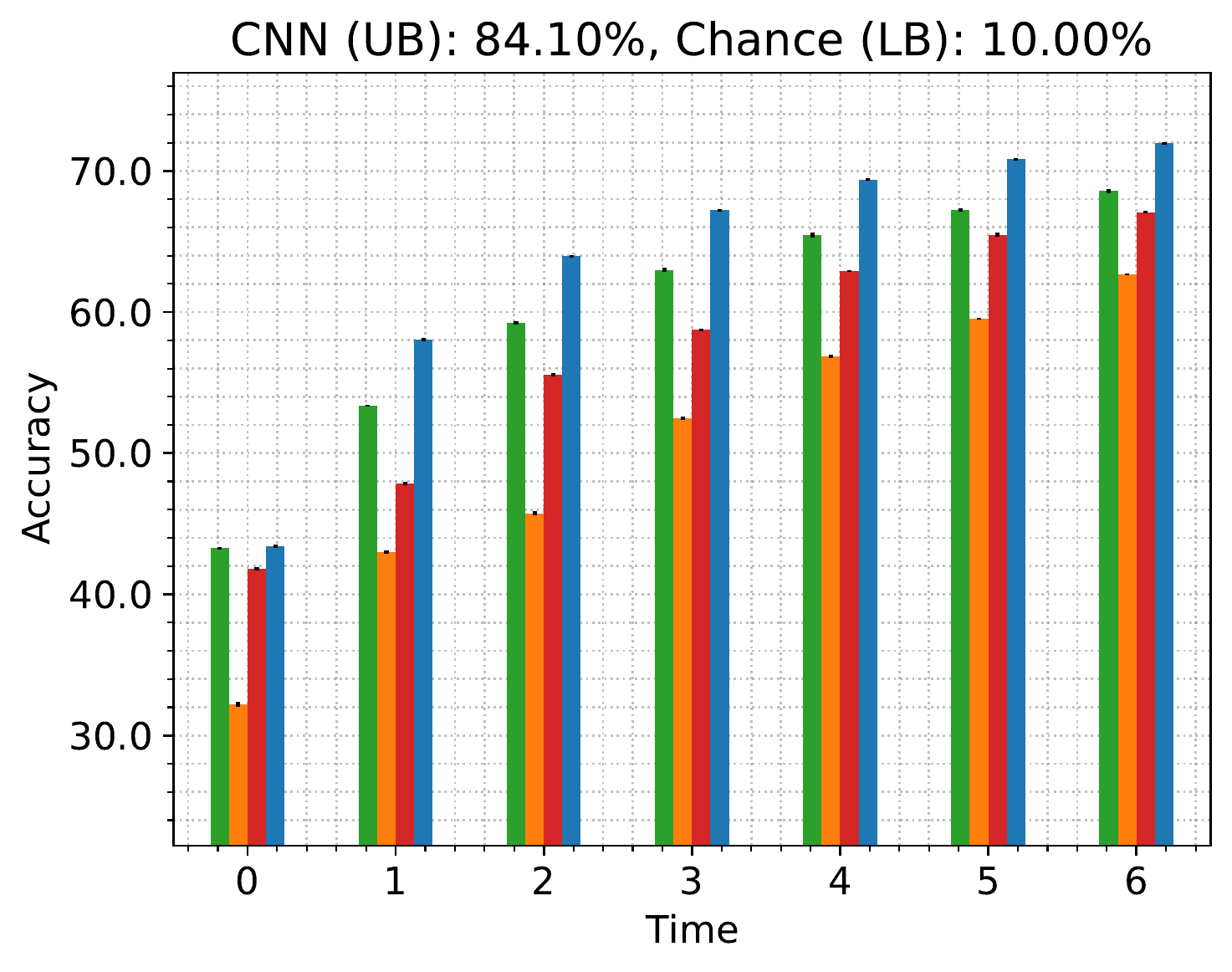}
    \centering{(c)}
    \end{minipage}
    \hfill
    \vfill
    \hfill
    \begin{minipage}[c]{0.32\linewidth}
    \includegraphics[width = \textwidth]{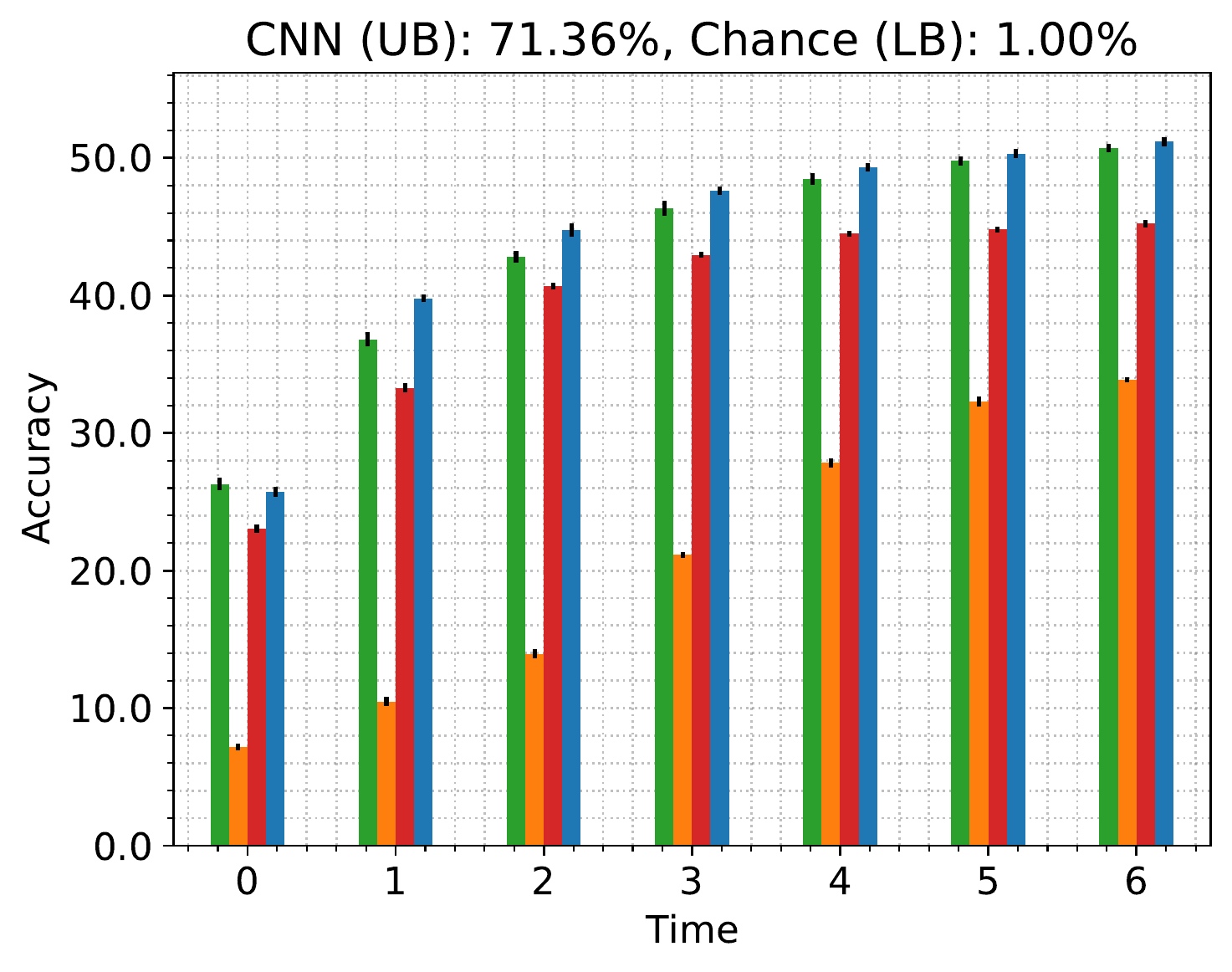}
    \centering{(d)}
    \end{minipage}
    \hfill
    \begin{minipage}[c]{0.32\linewidth}
    \includegraphics[width = \textwidth]{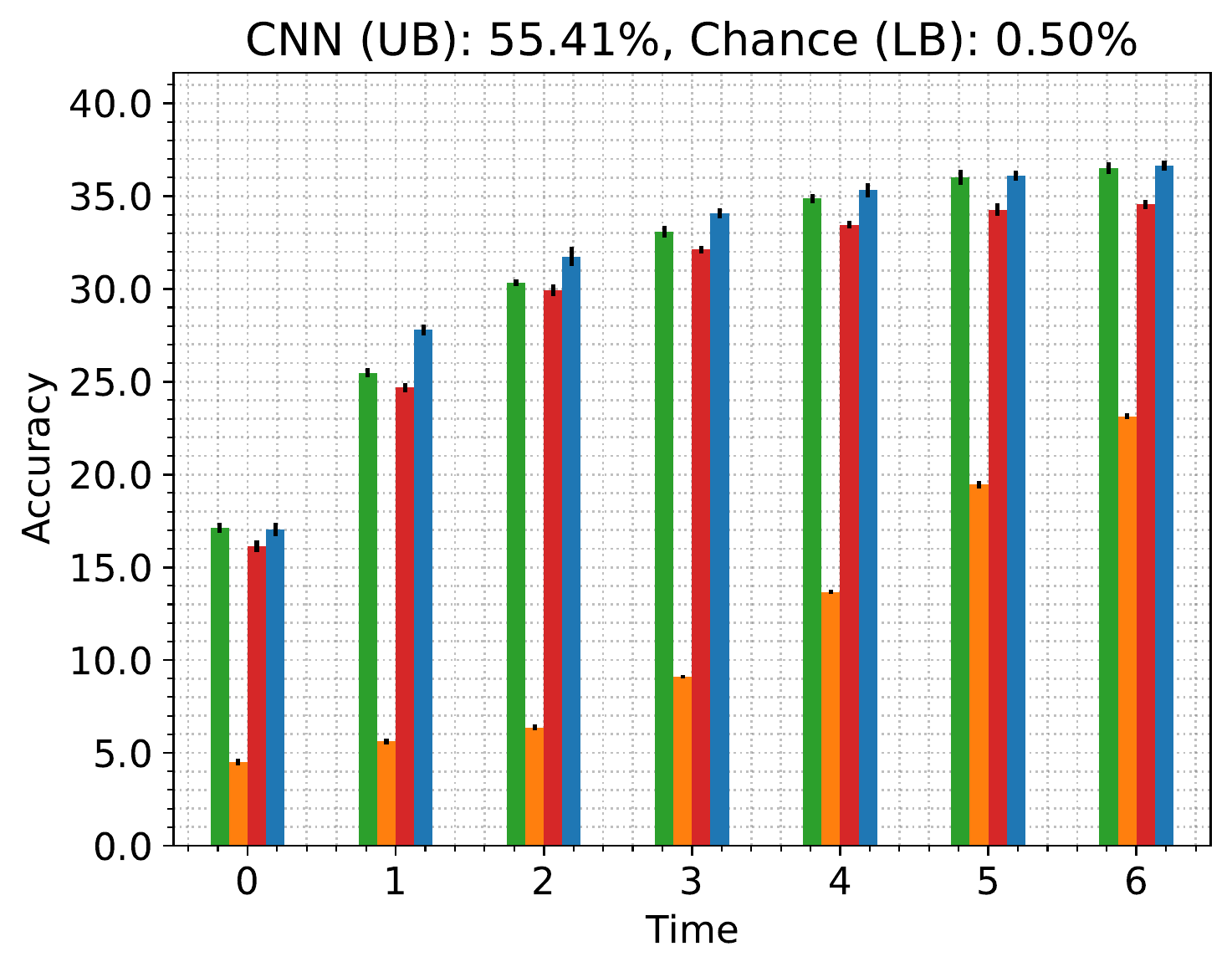}
    \centering{(e)}
    \end{minipage}
    \hfill
    \begin{minipage}[c]{0.32\linewidth}
    \includegraphics[width = \textwidth]{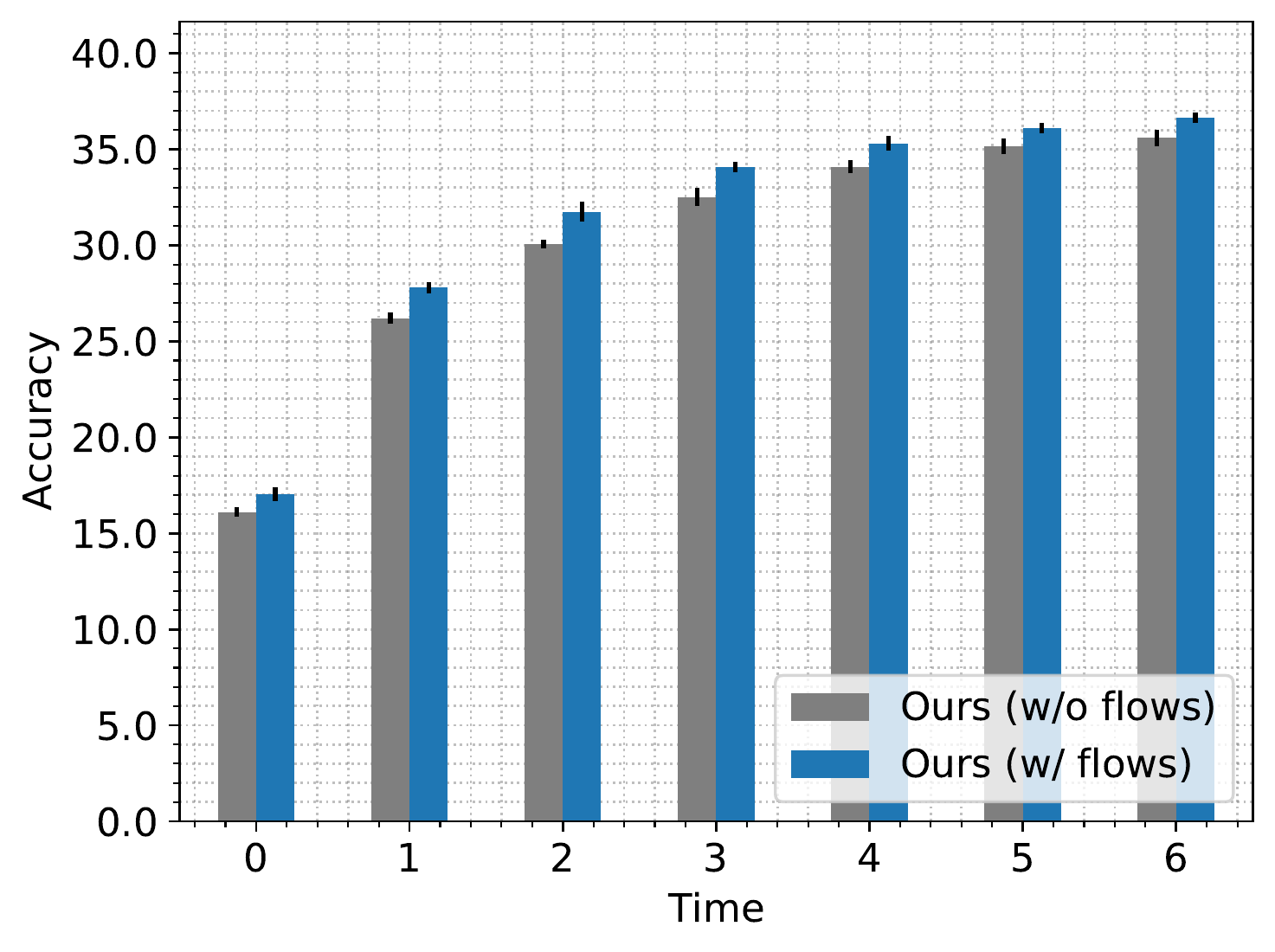}
    \centering{(f)}
    \end{minipage}
    \hfill
    \caption{\textit{Baseline Comparison on (a) SVHN (b) CIFAR-10 (c) CINIC-10 (d) CIFAR-100 (e) TinyImageNet. Ablation study on normalizing flows on (f) TinyImageNet.} (a-e) We compare various methods for $t=0$ to $6$. All results are averaged over ten different runs. Accuracy of a CNN and chance accuracy, displayed on top of each plot, serve as upper and lower bounds for the accuracy of glimpse-based methods. A CNN observes an entire image, whereas glimpse-based methods observe $<43.75\%$ area of the image by $t=6$. When compared with baseline methods, our method achieves the highest accuracy. (f) The use of normalizing flows improves accuracy.}
    \label{fig:compare}
    \vspace{-0.45cm}
\end{figure}

We compare our model with four baselines in Figure \ref{fig:compare}(a-e). RAM is a state-of-the-art hard attention model that observes images partially and sequentially to predict the class labels \cite{mnih2014recurrent}. We implement RAM using the same structure as our model. Instead of the Partial VAE, RAM has a controller that learns a Gaussian attention policy. \citet{mnih2014recurrent} minimize $\mathcal{L}_{CE}$ at the end of $T$ steps. Following \cite{li2017dynamic}, we improve RAM by minimizing $\mathcal{L}_{CE}$ at all $T$ steps. We refer to this baseline as RAM+. We also consider a baseline model that attends glimpses on random locations. The Random baseline does not have a controller or a Partial VAE. Our model and the three baselines described so far observe the image only partially through a series of glimpses. Additionally, we train a feed-forward CNN that observes the entire image to predict the class label.

For the SVHN dataset, the Random baseline outperforms RAM for initial time steps. However, with time, RAM outperforms the Random baseline by attending more useful glimpses. RAM+ outperforms RAM and the Random baselines at all time steps. We observe a different trend for non-digit datasets. RAM+ consistently outperforms RAM on non-digit datasets; however, RAM+ falls behind the Random baseline. In RAM and RAM+, the classifier shares latent space $h_t$ with the controller, while the Random baseline dedicates an entire latent space to the classifier. We speculate that the dedicated latent space in the Random baseline is one of the reasons for its superior performance on complex datasets.

\begin{figure}[h!]
    \hfill
    \begin{minipage}[c]{0.32\linewidth}
    \includegraphics[width = \textwidth]{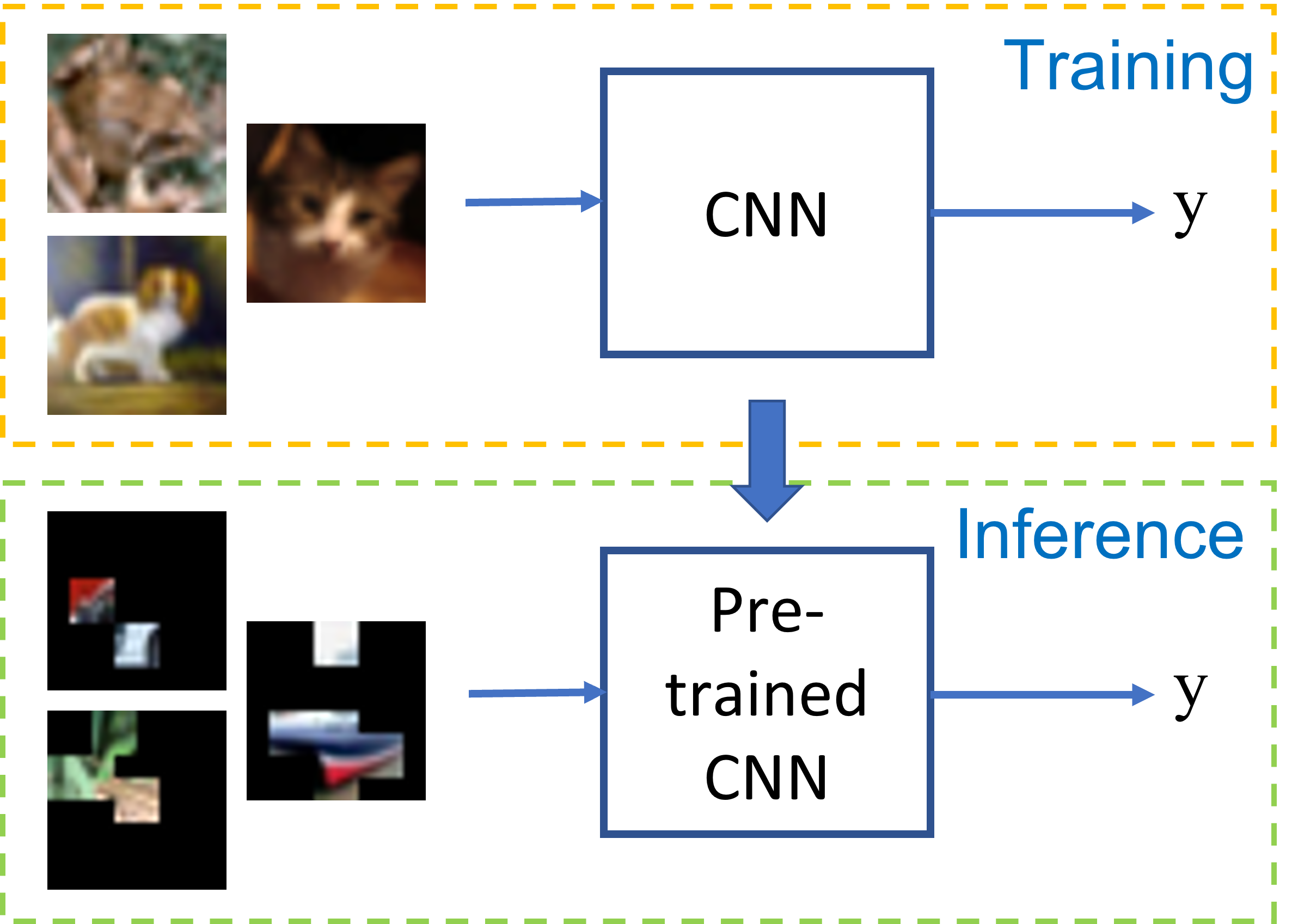}
    \centering{(a)}
    \end{minipage}
    \hfill
    \begin{minipage}[c]{0.32\linewidth}
    \includegraphics[width = \textwidth]{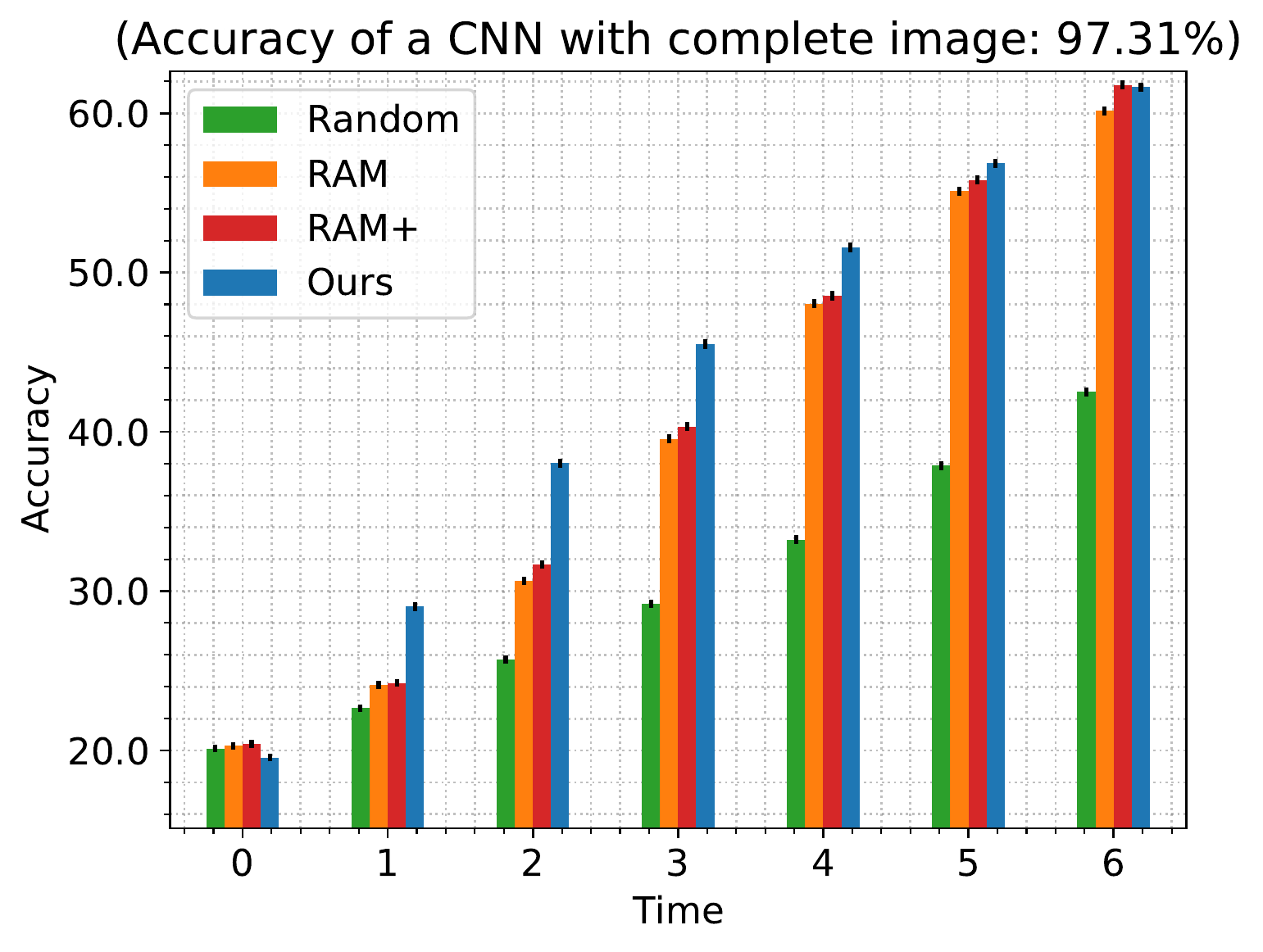}
    \centering{(b)}
    \end{minipage}
    \hfill
    \begin{minipage}[c]{0.32\linewidth}
    \includegraphics[width = \textwidth]{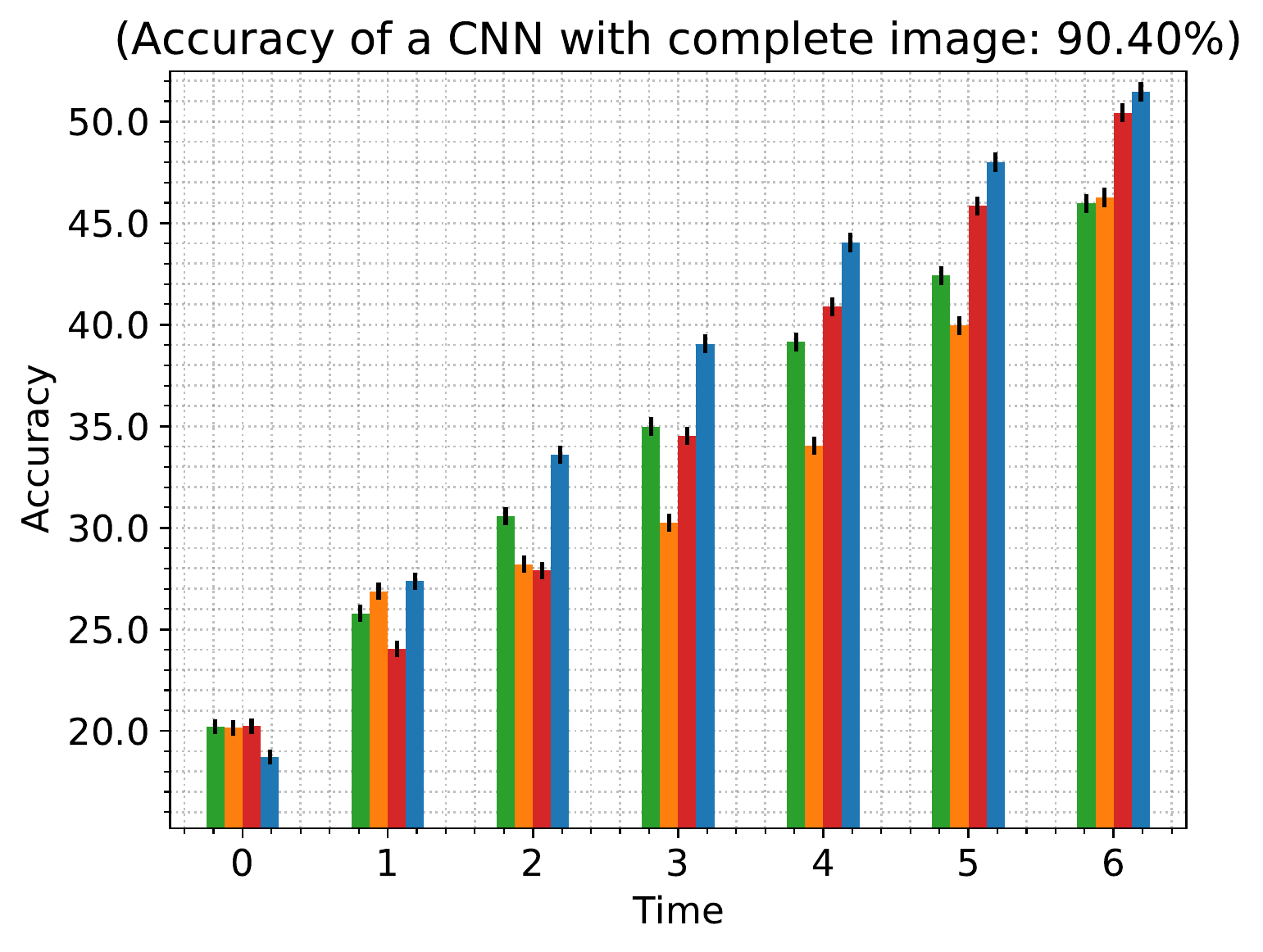}
    \centering{(c)}
    \end{minipage}
    \hfill
    \vfill
    \hfill
    \begin{minipage}[c]{0.32\linewidth}
    \includegraphics[width = \textwidth]{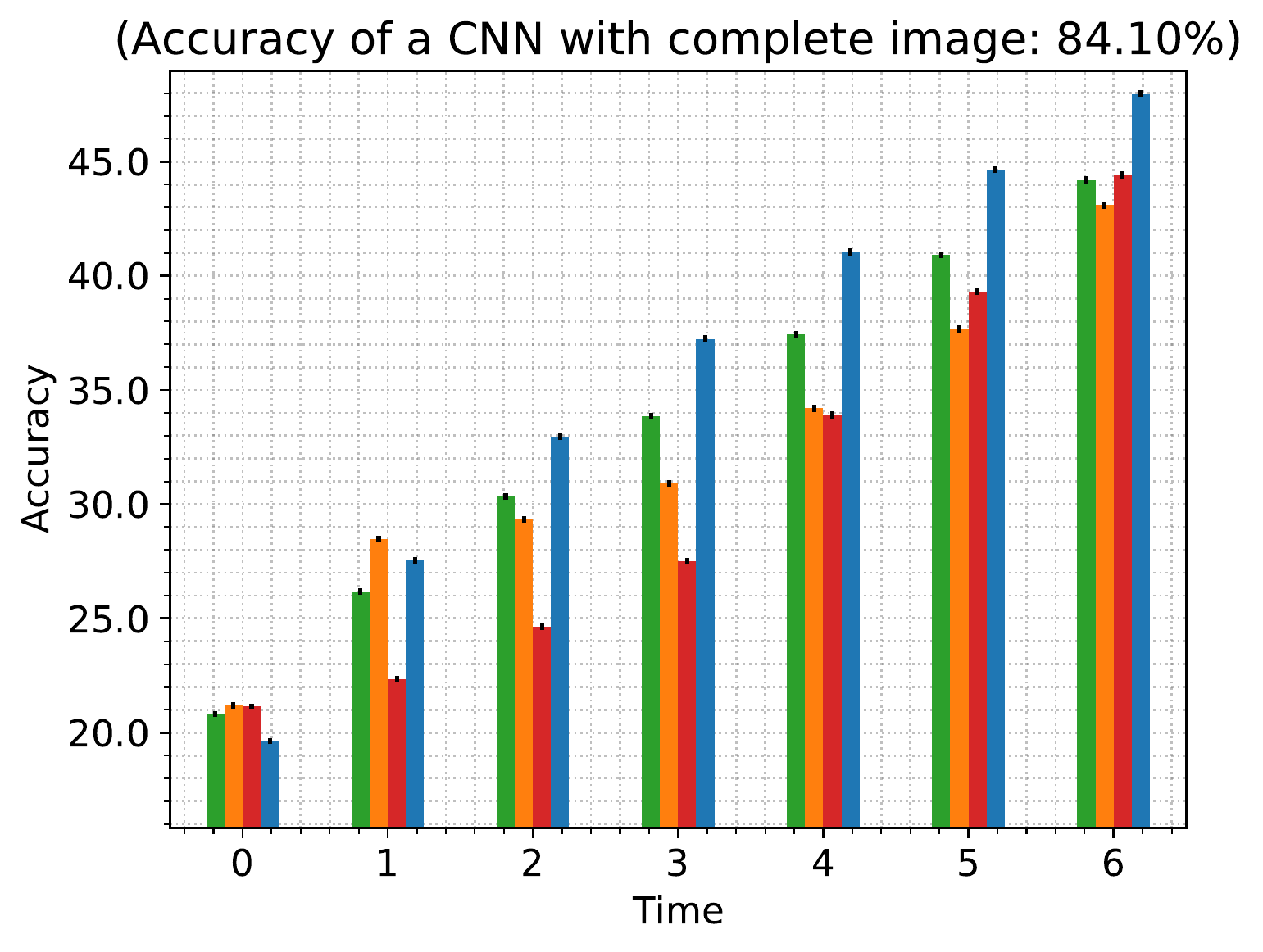}
    \centering{(d)}
    \end{minipage}
    \hfill
    \begin{minipage}[c]{0.32\linewidth}
    \includegraphics[width = \textwidth]{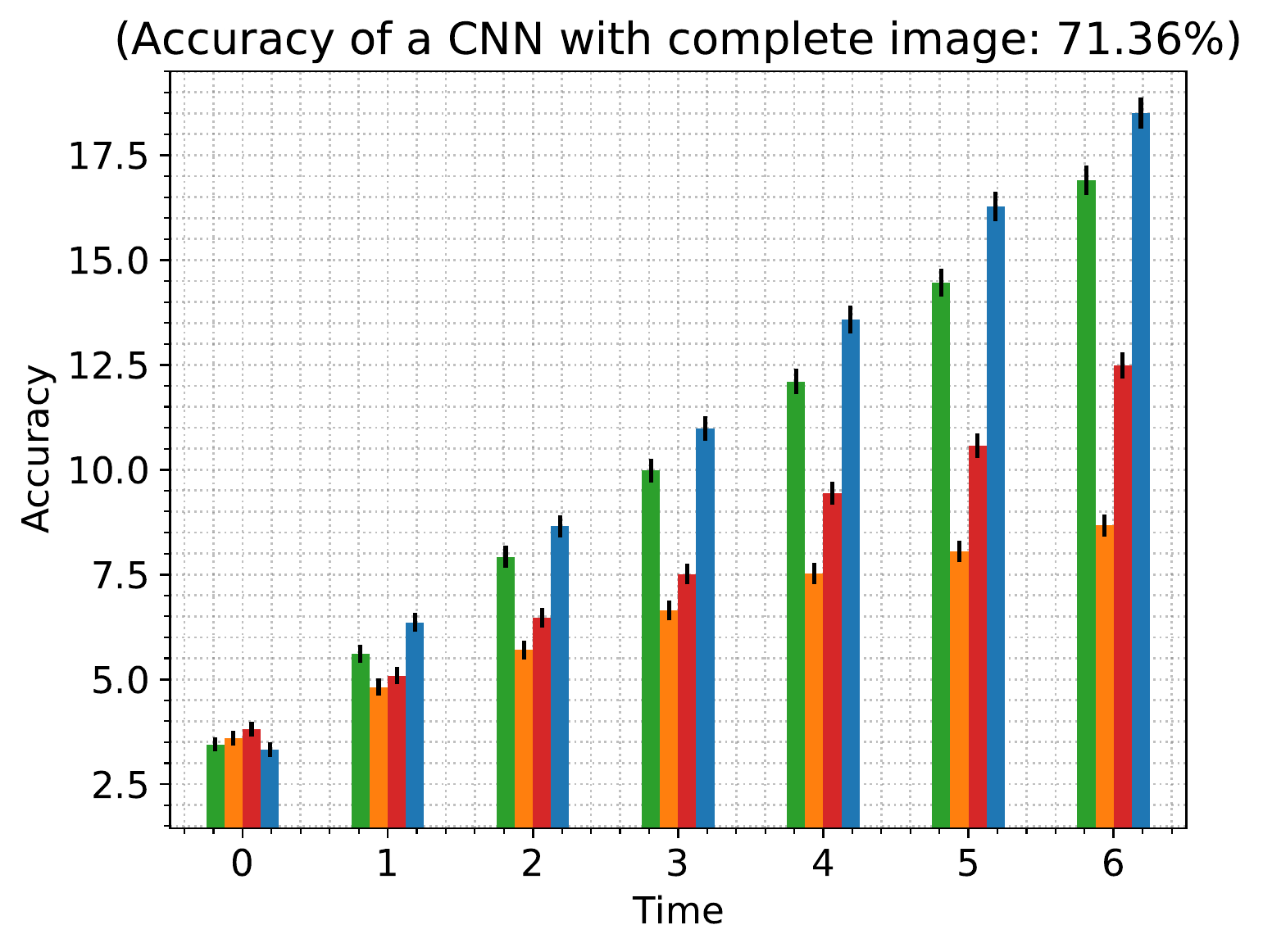}
    \centering{(e)}
    \end{minipage}
    \hfill
    \begin{minipage}[c]{0.32\linewidth}
    \includegraphics[width = \textwidth]{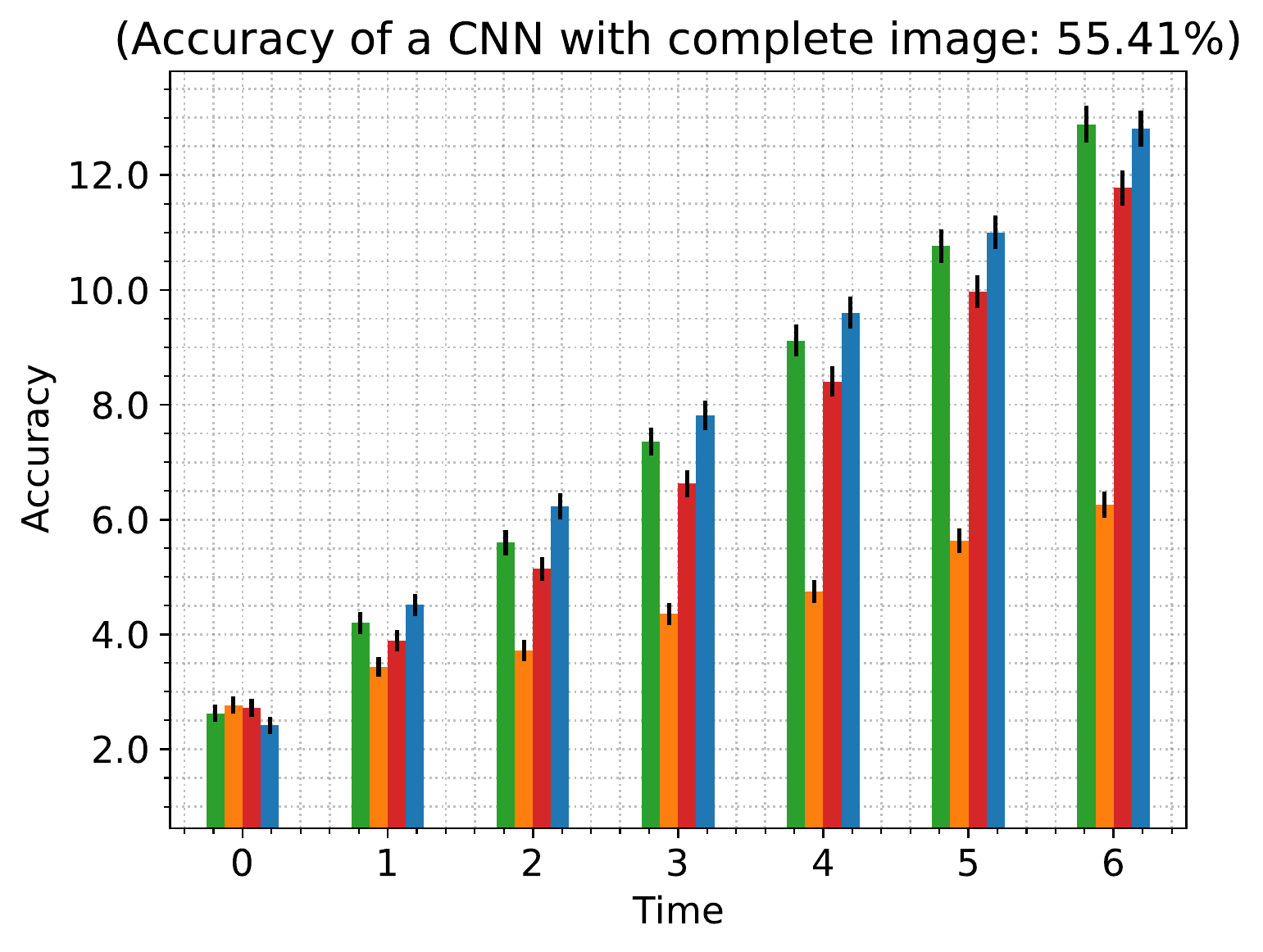}
    \centering{(f)}
    \end{minipage}
    \hfill
    \caption{\textit{Accuracy of a CNN when only the attended glimpses are observed}. (a) Experiment setup: we train a CNN using complete images and test it on masked images with only the glimpses attended by various methods made visible. Results on (b) SVHN (c) CIFAR-10 (d) CINIC-10 (e) CIFAR-100 (f) TinyImageNet. All results are averaged over three runs.}
    \label{fig:cnnobserved}
\end{figure}

\begin{figure}[!t]
    \hfill
    \begin{minipage}[c]{0.48\linewidth}
    \includegraphics[width = \textwidth]{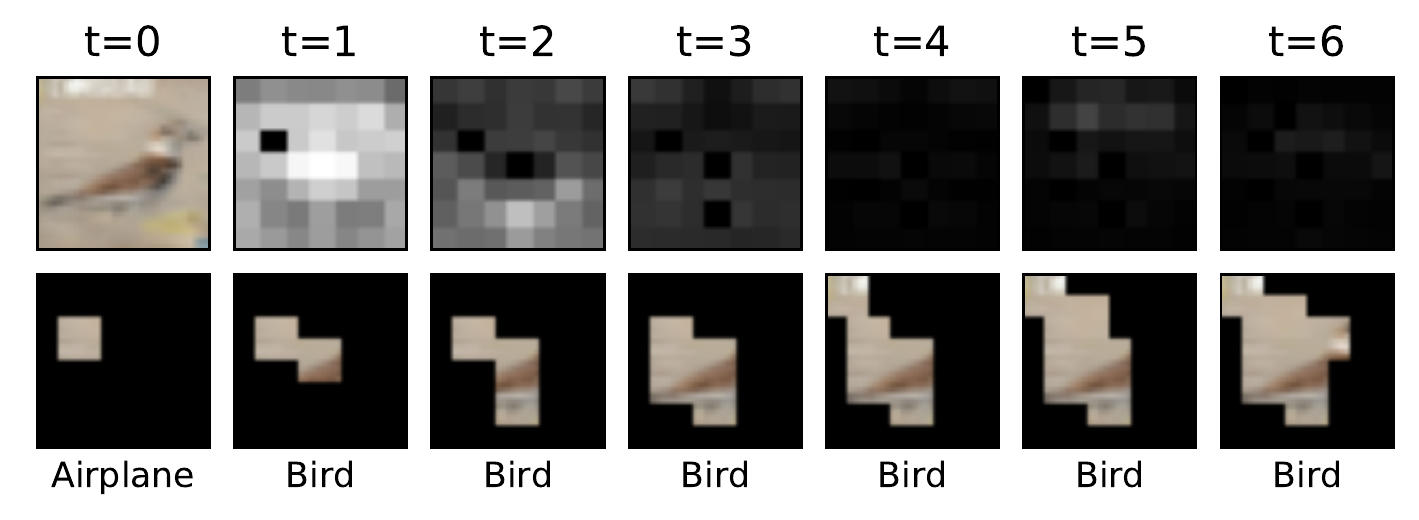}
    \centering{(a)}
    \end{minipage}
    \hfill
    \begin{minipage}[c]{0.48\linewidth}
    \includegraphics[width = \textwidth]{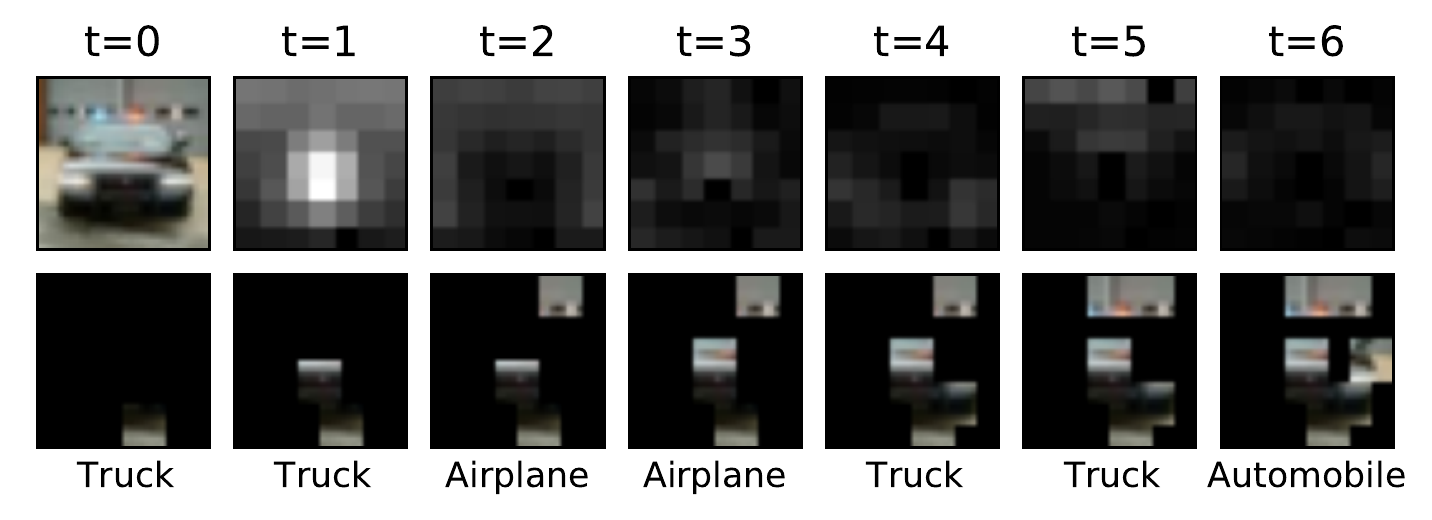}
    \centering{(b)}
    \end{minipage}
    \hfill
    \vfill
    \hfill
    \begin{minipage}[c]{0.48\linewidth}
    \includegraphics[width = \textwidth]{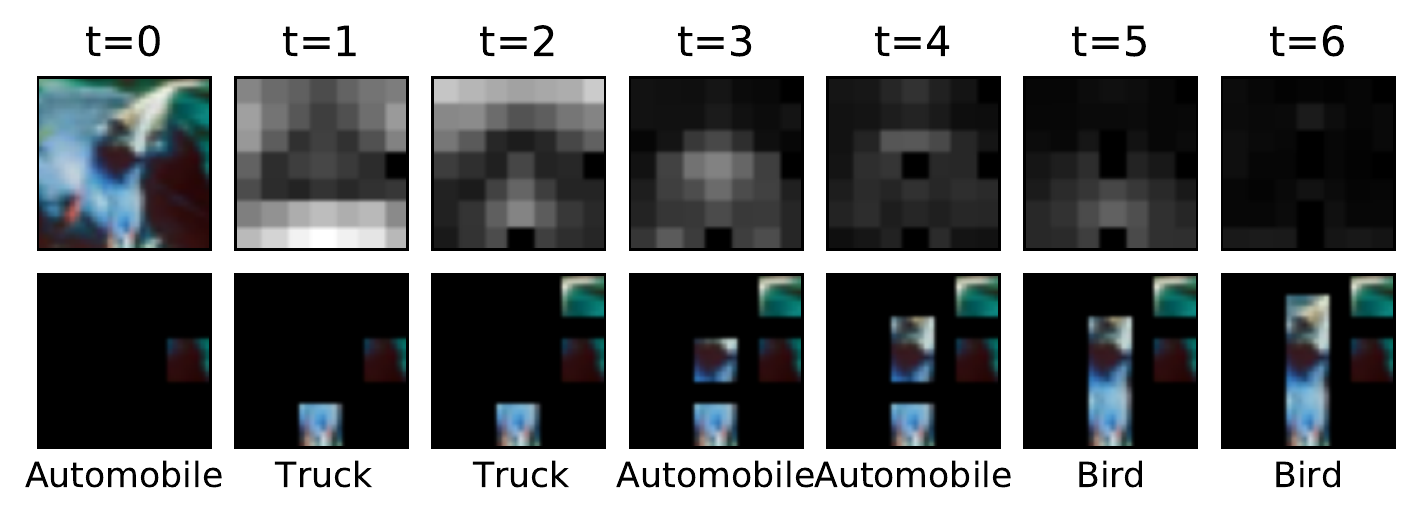}
    \centering{(c)}
    \end{minipage}
    \hfill
    \begin{minipage}[c]{0.48\linewidth}
    \includegraphics[width = \textwidth]{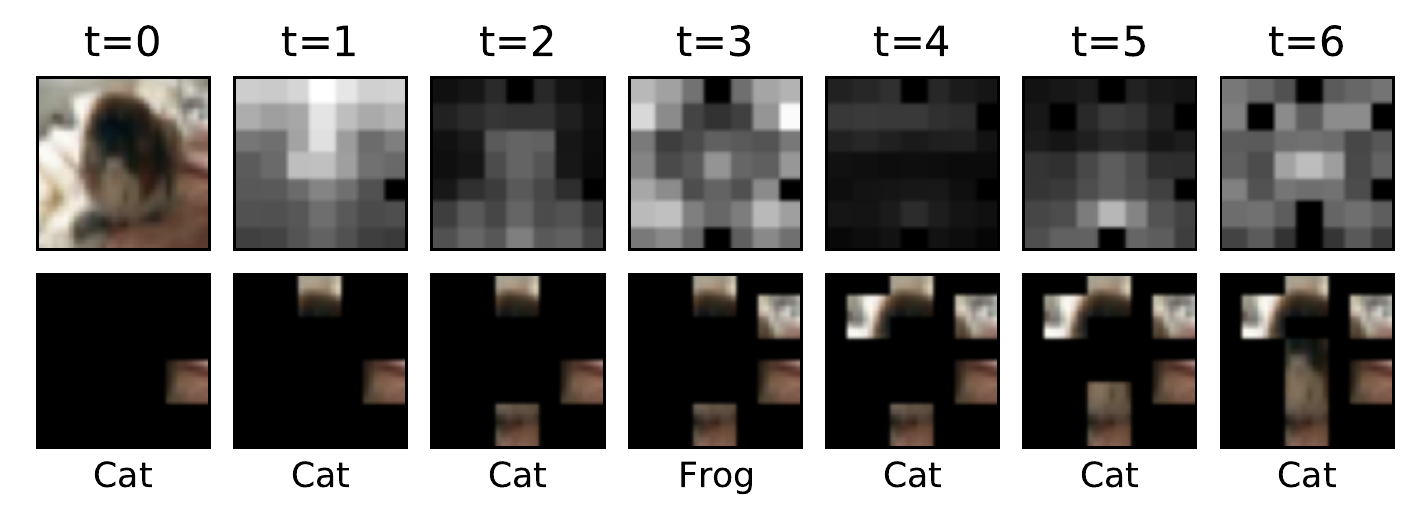}
    \centering{(d)}
    \end{minipage}
    \hfill
    \caption{\textit{Visualization of the $EIG$ maps and the glimpses observed by our model on CIFAR-10 images.}  The top rows in each plot show the entire image and the $EIG$ maps for $t=1$ to 6. The bottom rows in each plot show glimpses attended by our model. The model observes the first glimpse at a random location. Our model observes a glimpse of size $8\times 8$. The glimpses overlap with the stride of 4, resulting in a $7\times 7$ grid of glimpses. The EIG maps are of size $7 \times 7$ and are upsampled for the display. We display the entire image for reference; our model never observes the whole image. (a-c) success cases (d) failure case.}
    \label{fig:visual}
\end{figure}

Our model consistently outperforms all attention baselines on all datasets. The performance gap between the highest performing baseline and our model reduces with many glimpses, as one can expect. Predicting an optimal glimpse-location is difficult for early time-steps as the models have access to minimal information about the scene so far. Compared to the highest performing baseline at $t=1$, our model achieves around 10\% higher accuracy on SVHN, around 5-6\% higher accuracy on CIFAR-10 and CINIC-10, and around 2-3\% higher accuracy on CIFAR-100 and TinyImageNet. Note that CIFAR-100 and TinyImageNet are more challenging datasets compared to SVHN, CINIC-10, and CIFAR-10. Similar to RAM and RAM+, the classifier and the Partial VAE share a common latent space in our model. Hence, our model achieves a lower gain over the Random baseline for complex datasets. We attribute the small but definite gain in the accuracy of our model to a better selection of glimpses.

The CNN has the highest accuracy as it observes a complete image. The accuracy of the CNN serves as the upper bound for the hard attention methods. Unlike CNN, the hard attention methods observe less than half of the image through small glimpses, each uncovering only 6.25\% area of an image. On the CIFAR-10 dataset, the CNN predicts correct class labels for approximately 9 out of 10 images after observing each image completely. Remarkably, our model predicts correct class labels for 8 out of 10 images after observing less than half of the total area in each image.

\vspace{-0.25cm}
\subsubsection{Comparison of Attention Policies using a common CNN}
\label{sec:cnn_acc}

Above, we compared the attention policies of various methods using their respective classifiers. However, each model attains different discriminative power due to different training objectives. While RAM, RAM+, and our model are trained jointly for two different tasks, i.e., classification and glimpse-location prediction or feature-synthesis, the Random baseline is trained for only one task, i.e., classification. Consequently, the Random baseline attains higher discriminative power than others and achieves high accuracy despite using a sub-optimal attention policy. To make a fair comparison of the attention policies irrespective of the discriminative power of the models, we perform the following experiment.

We mask all image regions except for the ones observed by the attention model so far and let the baseline CNN predict a class label from this masked image (see Figure \ref{fig:cnnobserved}). RAM+ consistently outperforms RAM, suggesting that the former has learned a better attention policy than the latter. As RAM+ is trained using $\mathcal{L}_{CE}$ at all time-steps, it achieves higher accuracy, and ultimately, higher reward during training with REINFORCE \cite{mnih2014recurrent}. RAM and RAM+ outperform the Random baseline for the SVHN dataset. However, they fall short on natural image datasets. In contrast, our method outperforms all baselines with a significant margin on all datasets, suggesting that the glimpses selected by our model are more informative about the image class than the ones chosen by the baselines.

RAM and RAM+ struggle on images with many objects and repeated structure \cite{sermanet2014attention}, as often the case with natural image datasets. For example, TinyImageNet includes many images with multiple objects (e.g. beer bottle), repeated patterns (e.g. spider web), and dispersed items (e.g. altar). Note that a random policy can perform competitively in such scenarios, especially when the location of various objects in an image is unknown due to partial observability. Yet, our method can learn policies that are superior to the Random baseline. Refer to SM for additional analyses.

\vspace{-0.06cm}
\subsection{Ablation study on Normalizing Flows}
We inspect the necessity of a flexible distribution for the posterior $q(z|h_t)$ and, therefore, the necessity of normalizing flows in the encoder $S$. To this end, we model the posterior with a unimodal Gaussian distribution and let $S$ output mean and diagonal covariance of a Gaussian. We do not use flow layers in this case. Figure \ref{fig:compare}(f) shows the result for TinyImageNet dataset. We observe that modeling a complex posterior using normalizing flows improves accuracy. Ideally, the Partial VAE should predict all possibilities of $\Tilde{f}$ consistent with the observed region $o_t$. When the model observes a small region, a complex posterior helps determine multiple plausible feature maps. A unimodal posterior fails to cover all possibilities. Hence, the EIG estimated with the former is more accurate, leading to higher performance. Refer to SM for visualization of $q(z|h_t)$ estimated with and without normalizing flows.

\vspace{-0.06cm}
\subsection{Visualization}
We visualize a few interesting examples of sequential recognition from the CIFAR-10 dataset in Figure \ref{fig:visual}. Refer to SM for additional examples. In Figure \ref{fig:visual}(a), activity in the $EIG$ map reduces as the model settles on a class `Bird'. In Figure \ref{fig:visual}(b), the model takes a long time to decide the true class of the image. Though incorrect, it predicts classes that are types of vehicles. After discovering parts like headlight and rear-view mirror, it predicts the true class `Automobile' at $t=6$. Figure \ref{fig:visual}(c) shows a difficult example. The model decides the true class `Bird' after covering parts of the bird at $t=5$. Notice a high amount of activity in the $EIG$ maps up to $t=5$ and reduced activity at $t=6$. Figure \ref{fig:visual}(d) shows a failure case with enduring activity in $EIG$ maps. Finally, observe that the $EIG$ maps are often multimodal.

\vspace{-0.06cm}
\section{Conclusions}
We presented a hard attention model that uses BOED to find the optimal locations to attend when the image is observed only partially. To find an optimal location without observing the corresponding glimpse, the model uses Partial VAE to synthesize the content of the glimpse in the feature space. Synthesizing features of unobserved regions is an ill-posed problem with multiple solutions. We use normalizing flows in Partial VAE to capture a complex distribution of unobserved glimpse features, which leads to improved performance. The synthesized features enable the model to evaluate and compare the expected information gain (EIG) of various candidate locations, from which the model selects a candidate with optimal EIG. The predicted EIG maps are often multimodal. Consequentially, the attention policy used by our model is multimodal. Our model achieves superior performance compared to the baseline methods that use unimodal attention policy, proving the effectiveness of multimodal policies in hard attention. When all models have seen only a couple of glimpses, our model achieves 2-10\% higher accuracy than the baselines. 

\newpage
\bibliography{egbib}

\newpage
\appendix
\section{Architecture}
Table \ref{tab:arch} shows the architecture of various components of our model. Note that all modules use a small number of layers. $F_g$ uses the least possible layers to achieve the effective receptive field equal to the area of a single glimpse. The decoder $D$ uses the smallest number of $ConvTranspose$ and $Conv$ layers to generate the feature maps of required spatial dimensions and refine them based on the global context. The encoder $S$ uses flow layers according to the complexity of the dataset. All other modules use a single linear layer. We implement linear layers using $1\times 1$ convolution layers.
The dimensionality of features $f$, hidden state $h$ and latent representation $z$ for various datasets are mentioned in Table \ref{tab:space}. 
\begin{table}[h]
    \centering
    \begin{tabular}{|p{1.5cm}|c|c|}
    \hline
         & Module & Architecture \\
    \hline
    \multirow{4}{1.5cm}{Recurrent feature aggregator} & \rule{0pt}{10pt}$\mathcal{F}_l$ & $Conv_{k=1}(\cdot)$ \\
         \cline{2-3}
         &\rule{0pt}{10pt}$\mathcal{F}_g$ & $n_g\times\{BN(LeakyReLU(Conv_{k=3}(\cdot))\}$ \\
         &&$Conv_{k=2}(\cdot)$ \\
         \cline{2-3}
         &\rule{0pt}{10pt}$F(g, l)$ & $\mathcal{F}_g(g) + \mathcal{F}_l(l)$\\
         \cline{2-3}
         &\rule{0pt}{10pt}$\mathcal{F}_h$ & $Conv_{k=1}(\cdot)$ \\
         \cline{2-3}
         &\rule{0pt}{10pt}$\mathcal{F}_f$ & $Conv_{k=1}(BN(LeakyReLU(\cdot)))$ \\
         \cline{2-3}
         &\rule{0pt}{10pt}$R(h,f)$ & $LN(LeakyReLU(\mathcal{F}_h(h) + \mathcal{F}_f(f)))$ \\
         \hline
    Classifier &\rule{0pt}{10pt}$C$ & $Softmax(Conv_{k=1}(Dropout_{p=0.5}(\cdot)))$\\
         \hline
    \multirow{3}{1.5cm}{Partial VAE}& \rule{0pt}{10pt}$S$ & $n_s\times(ActNorm(Flip(NSF(\cdot))))$\\
         \cline{2-3}
         &\rule{0pt}{10pt} $D$ & $3\times\{LN(LeakyReLU(ConvTranspose_{k=3}(\cdot)\}$\\
         && $5\times\{LN(LeakyReLU(Conv_{k=3}^{p=1}(\cdot)\}$\\
         && $Conv_{k=3}^{p=1}(\cdot)$\\
    \hline
    \end{tabular}
    \caption{Architecture of our model. $k = kernel\_size$, $p = padding$. $n_g$ is set to 3 for SVHN, CINIC-10, CIFAR-10 and CIFAR-100, and set to 7 for TinyImageNet. $n_s$ is set to 4 for SVHN, CINIC-10 and CIFAR-10, and set to 6 for CIFAR-100 and TinyImageNet. BN = Batch Normalization \cite{ioffe2015batch}. LN = Layer Normalization \cite{ba2016layer}. NSF = Neural Spline Flows \cite{durkan2019neural}. ActNorm layer is presented in \cite{kingma2018glow}.}
    \label{tab:arch}
\end{table}

\begin{table}[h]
    \centering
    \begin{tabular}{|l|c|c|c|}
    \hline
              & $f$ & $h$ & $z$ \\
    \hline
         SVHN & 128 & 512 & 256 \\ 
         CINIC-10 & 128 & 512 & 256 \\
         CIFAR-10 & 128 & 512 & 256 \\
         CIFAR-100 & 512 & 2048 & 1024 \\
         TinyImageNet & 512 & 2048 & 1024 \\
    \hline
    \end{tabular}
    \caption{Dimensionality of features $f$, hidden state $h$ and latent representation $z$.}
    \label{tab:space}
\end{table}

\section{Optimization Details}
We train our model in three stages until convergence. In the first stage we pre-train $F$, $R$ and $C$. In the second stage, we pre-train $S$ and $D$ while keeping $F$, $R$, and $C$ frozen. In the third stage, we finetune all modules end-to-end. Unless stated otherwise, we use the same setting for all three stages. We trained our models on a single Tesla P100 GPU with 12GB of memory or a single Tesla V100 GPU with 16GB of memory.\\

\minisection{Data Preparation.} We augment training images using random crop, scale, horizontal flip and color jitter transformations, and map pixel values in range $[-1,1]$. We use a batch-size of 64. We use the same scheme for all datasets.\\

\minisection{Loss Function.} We compute an average loss across batch and time. We set the hyperparameters $\alpha$ and $\beta$ of the loss function as follows. The hyperparameter $\alpha$ is set to the inverse of dimensionality of the latent representation $z$. The hyperparameter $\beta$ is set to 32, 16, 16, 8 and 8 for SVHN, CINIC-10, CIFAR-10, CIFAR-100 and TinyImageNet respectively.\\

\minisection{Optimizer.} We use Adam optimizer \cite{kingma2014adam} with the default setting of $(\beta_1, \beta_2)=(0.9, 0.999)$. In the first training stage, we use a learning rate of 0.001 for all datasets. In the second and third training stages, we use a learning rate of 0.001 for SVHN, CIFAR-10, and CINIC-10, and a learning rate of 0.0001 for CIFAR-100 and TinyImageNet. We divide the learning rate by 0.5 at a plateau.

\begin{figure}[t]
    \hfill
    \begin{minipage}[c]{0.32\linewidth}
    \includegraphics[width = \textwidth]{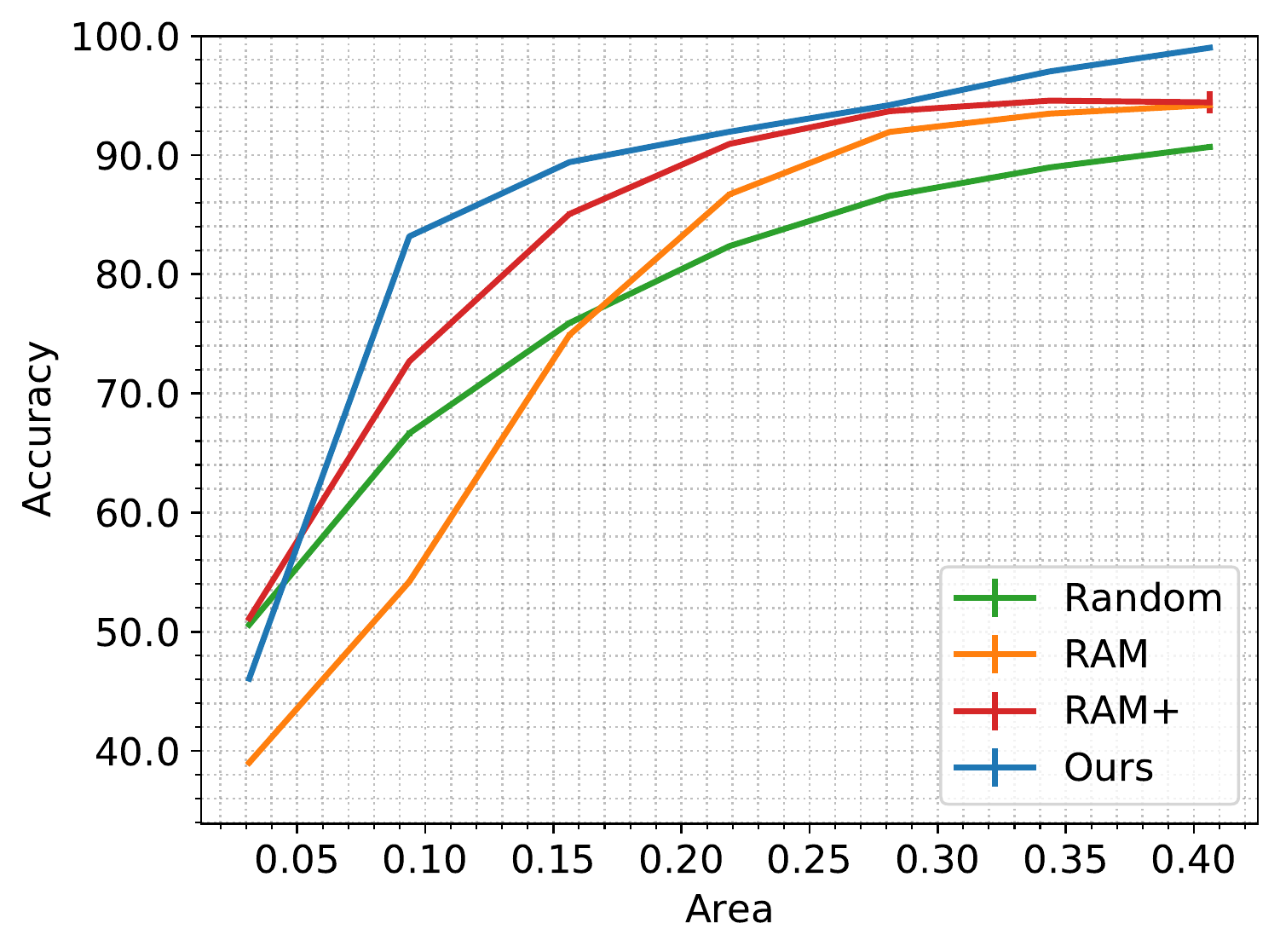}
    \centering{(a)}
    \end{minipage}
    \hfill
    \begin{minipage}[c]{0.32\linewidth}
    \includegraphics[width = \textwidth]{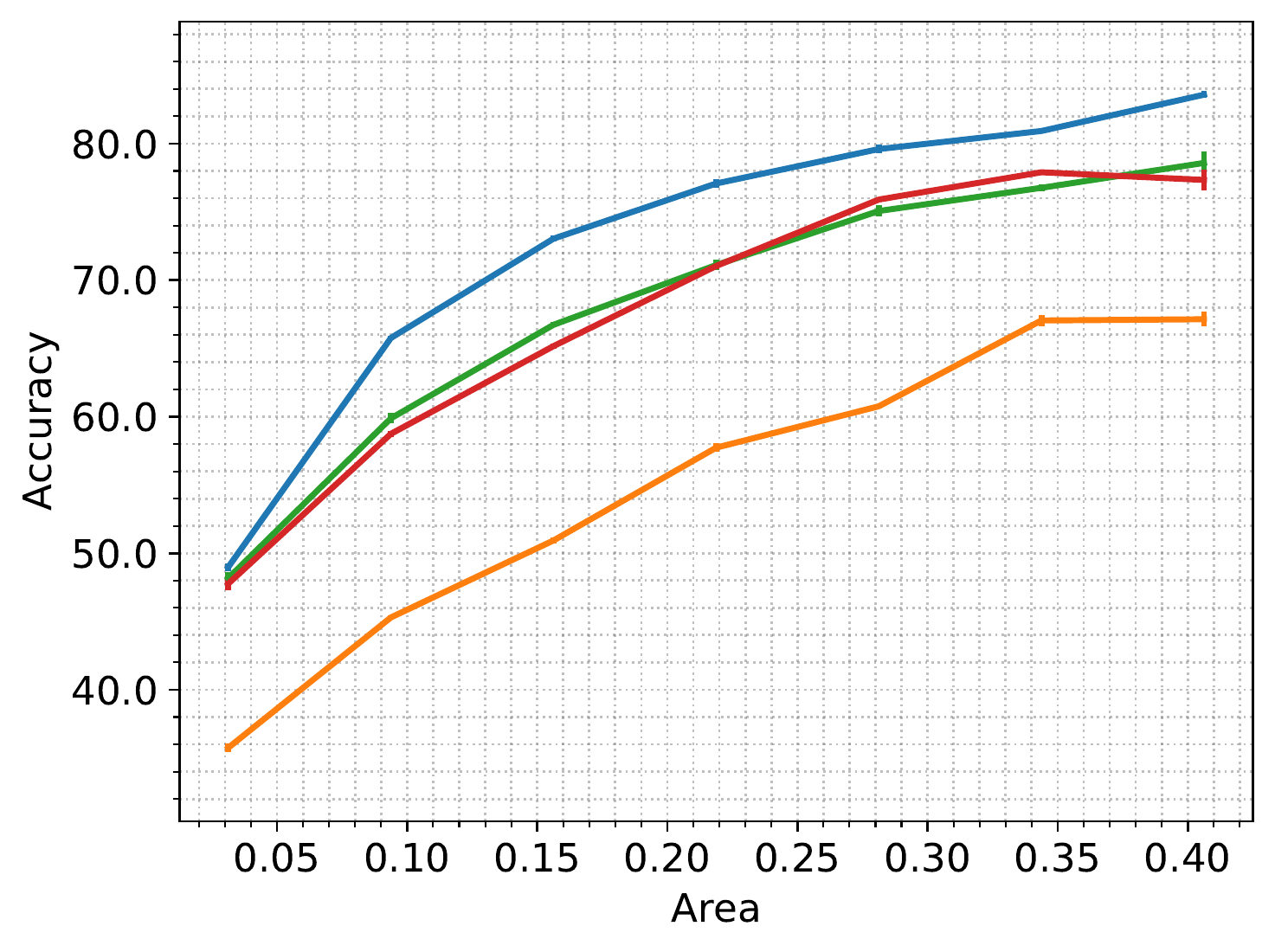}
    \centering{(b)}
    \end{minipage}
    \hfill
    \begin{minipage}[c]{0.32\linewidth}
    \includegraphics[width = \textwidth]{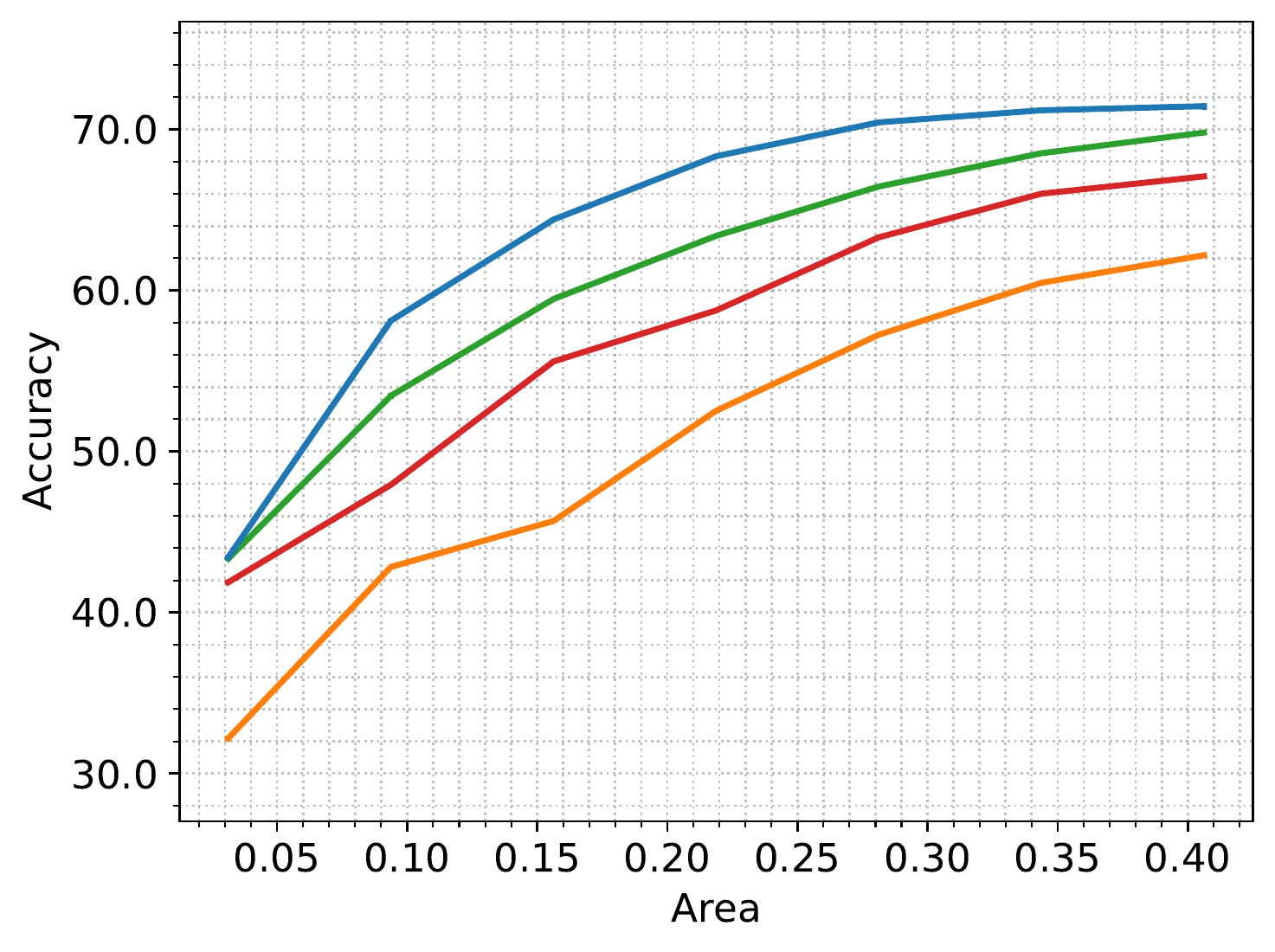}
    \centering{(c)}
    \end{minipage}
    \hfill
    \vfill
    \hfill
    \begin{minipage}[c]{0.32\linewidth}
    \includegraphics[width = \textwidth]{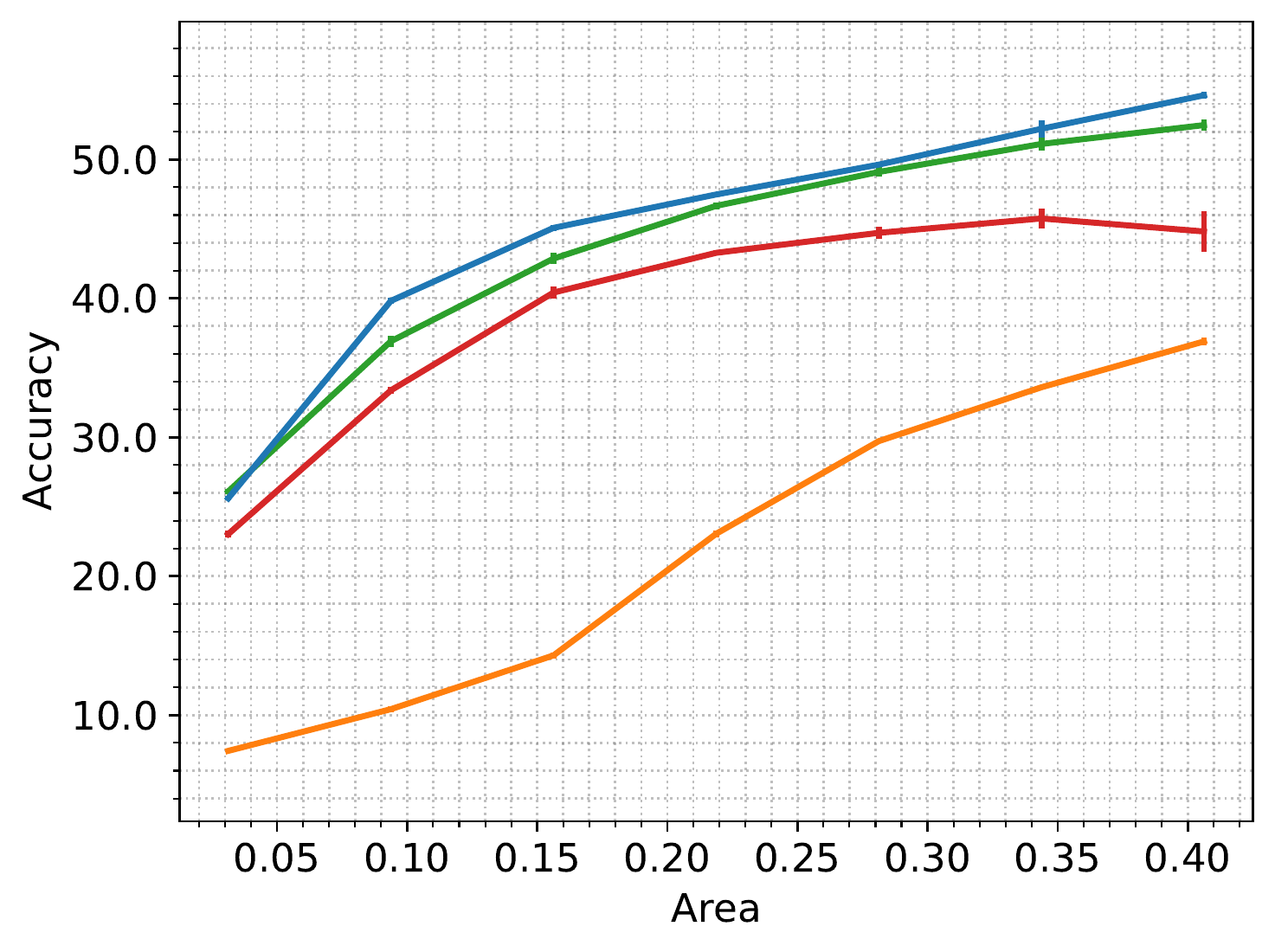}
    \centering{(d)}
    \end{minipage}
    \hfill
    \begin{minipage}[c]{0.32\linewidth}
    \includegraphics[width = \textwidth]{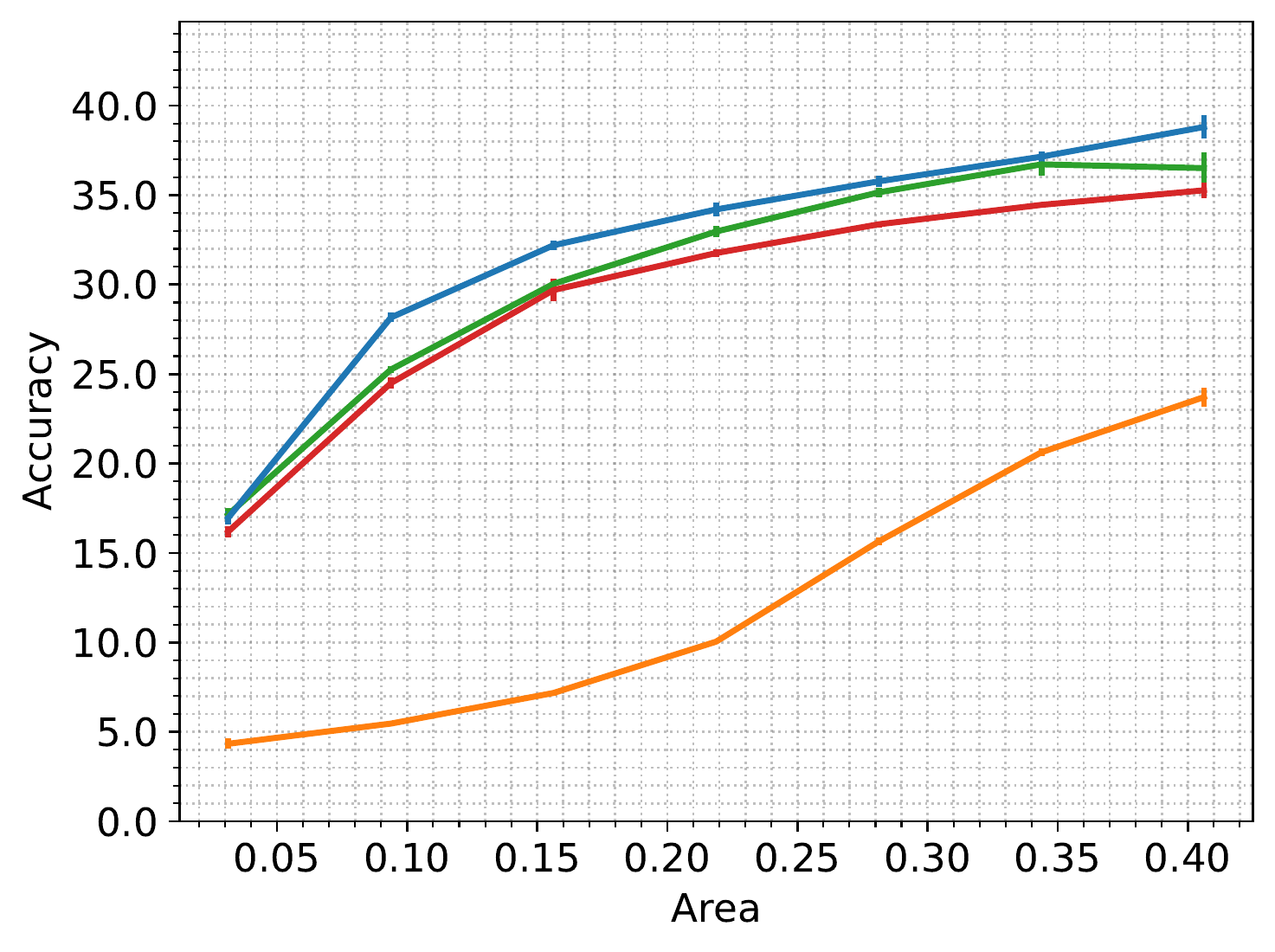}
    \centering{(e)}
    \end{minipage}
    \hfill
    \caption{\textit{Accuracy as a function of area observed in an image}. (a) SVHN (b) CIFAR-10 (C) CINIC-10 (d) CIFAR-100 (e) TinyImageNet. Hard attention models observe only a fraction of area in the image through multiple glimpses. Accuracy of the models increase as they observe more area. Our model achieves the highest accuracy for any given value of the observed area. Results are averaged over three independent runs.}
    \label{fig:areaacc}
\end{figure}
\begin{figure}[h!]
    \hfill
    \begin{minipage}[c]{0.32\linewidth}
    \includegraphics[width = \textwidth]{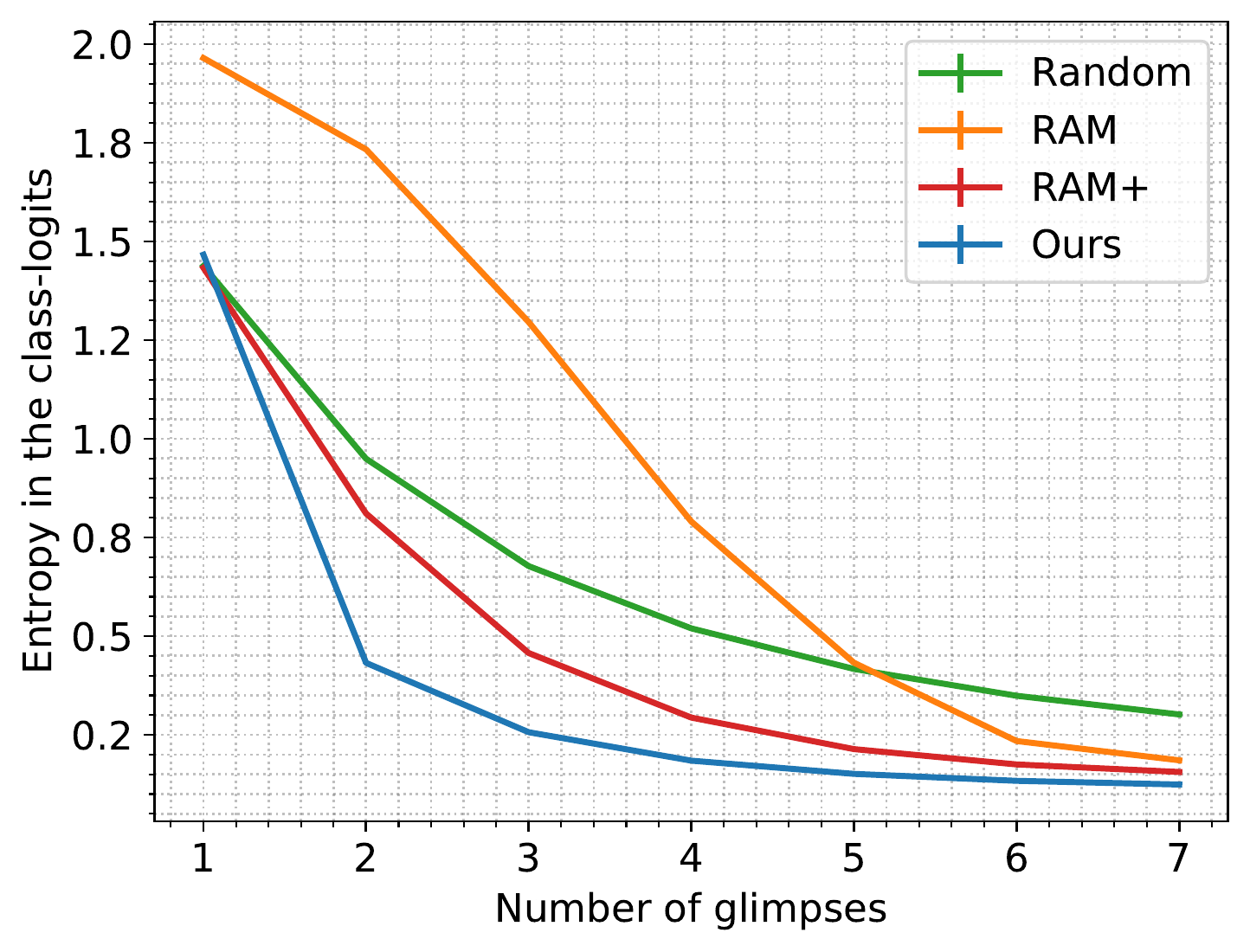}
    \centering{(a)}
    \end{minipage}
    \hfill
    \begin{minipage}[c]{0.32\linewidth}
    \includegraphics[width = \textwidth]{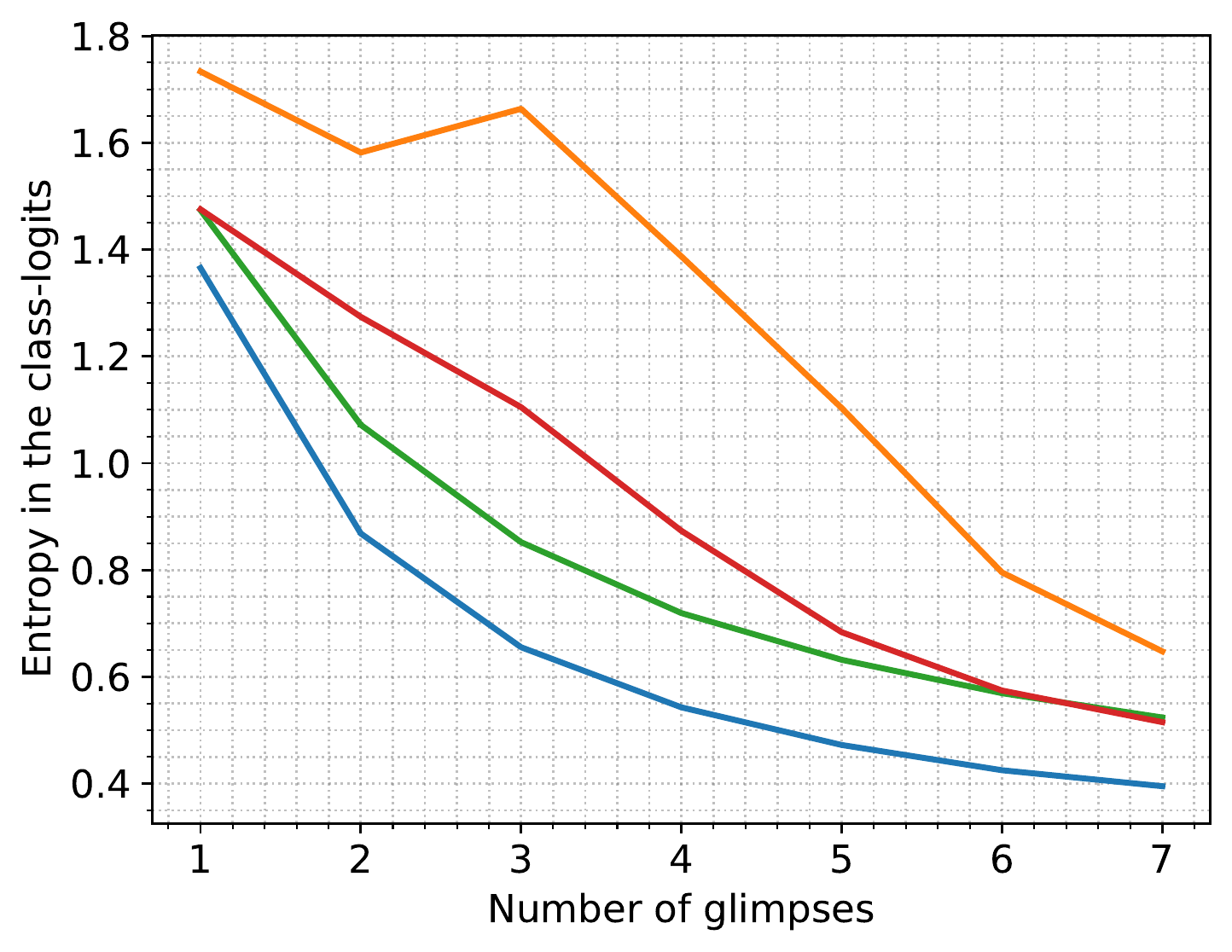}
    \centering{(b)}
    \end{minipage}
    \hfill
    \begin{minipage}[c]{0.32\linewidth}
    \includegraphics[width = \textwidth]{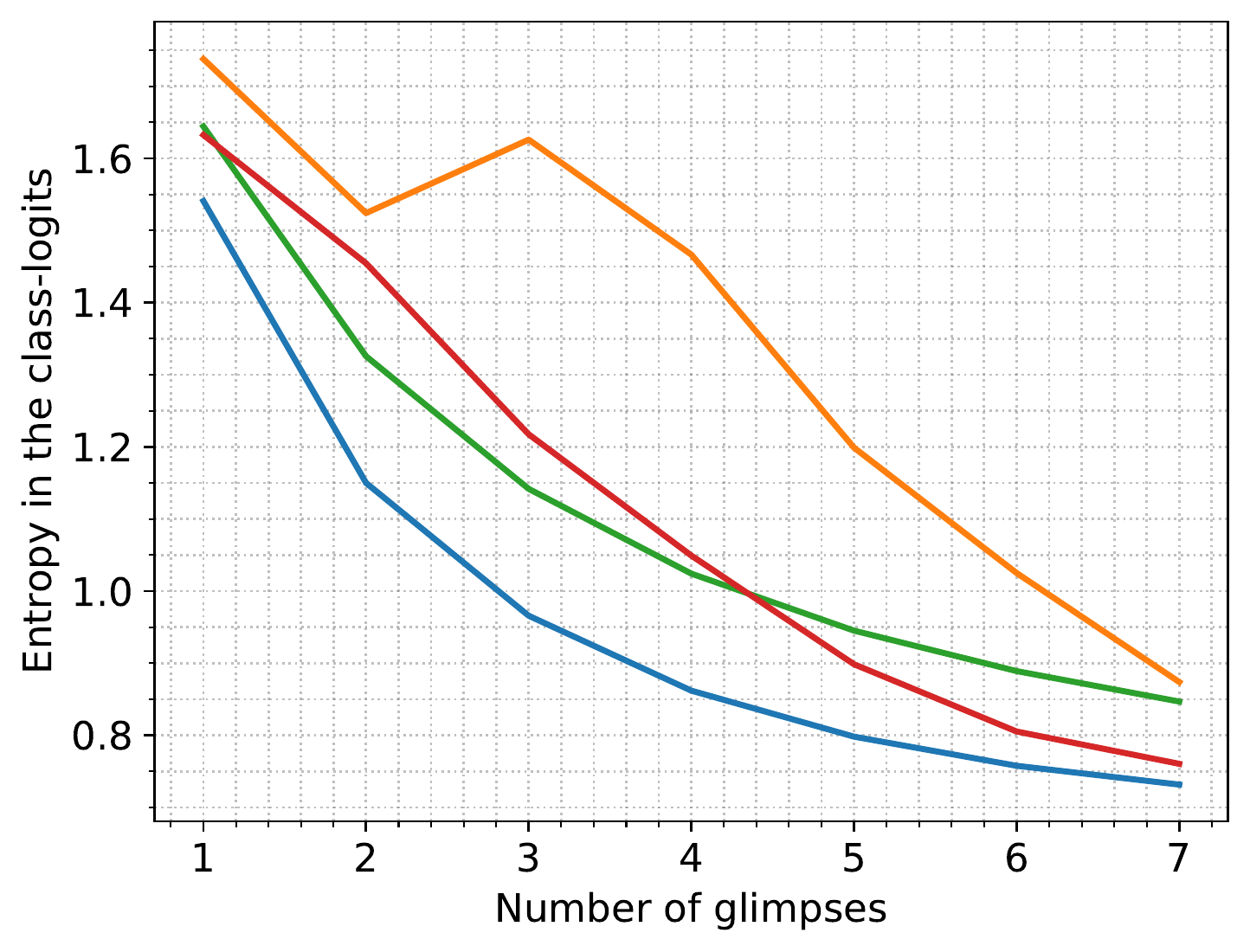}
    \centering{(c)}
    \end{minipage}
    \hfill
    \vfill
    \hfill
    \begin{minipage}[c]{0.32\linewidth}
    \includegraphics[width = \textwidth]{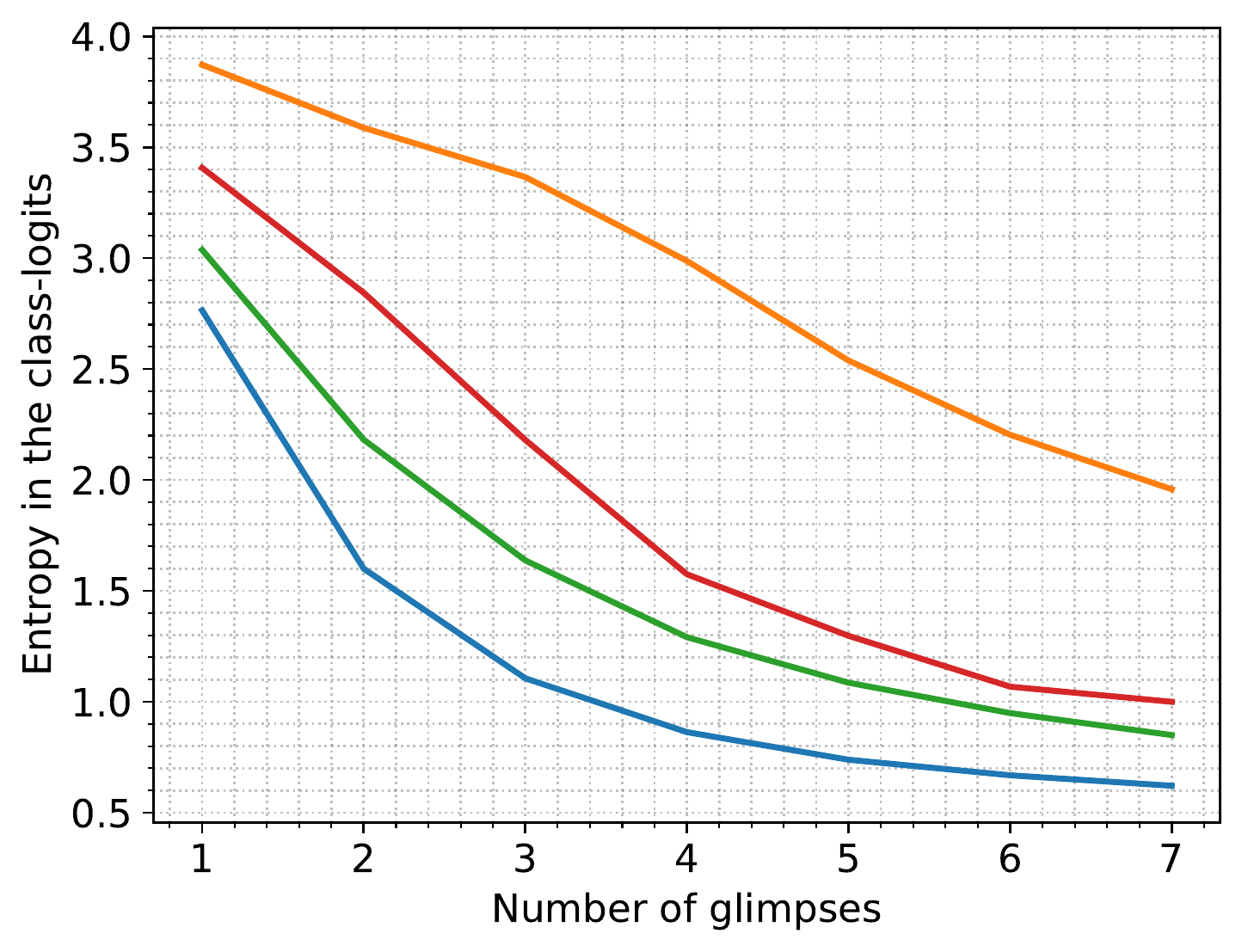}
    \centering{(d)}
    \end{minipage}
    \hfill
    \begin{minipage}[c]{0.32\linewidth}
    \includegraphics[width = \textwidth]{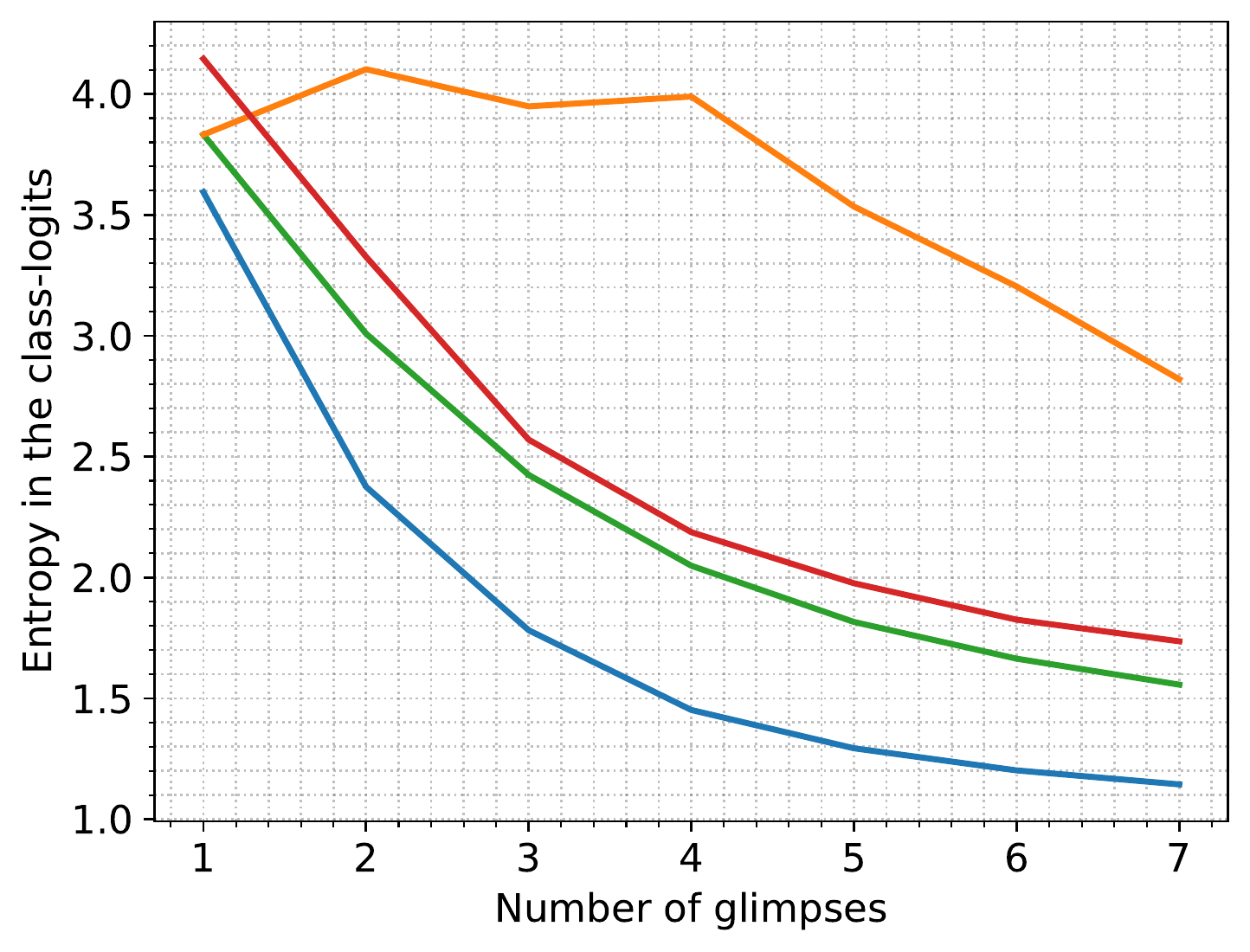}
    \centering{(e)}
    \end{minipage}
    \hfill
    \caption{\textit{Entropy in the class-logits vs number of glimpses}. (a) SVHN (b) CIFAR-10 (C) CINIC-10 (d) CIFAR-100 (e) TinyImageNet. Entropy in the class-logits decreases as the hard attention models acquire more glimpses. Our model achieves the lowest entropy in the predictions at all times. Results are averaged over three independent runs.}
    \label{fig:conf}
\end{figure}
\begin{figure}[h!]
    \hfill
    \begin{minipage}[c]{0.32\linewidth}
    \includegraphics[width = \textwidth]{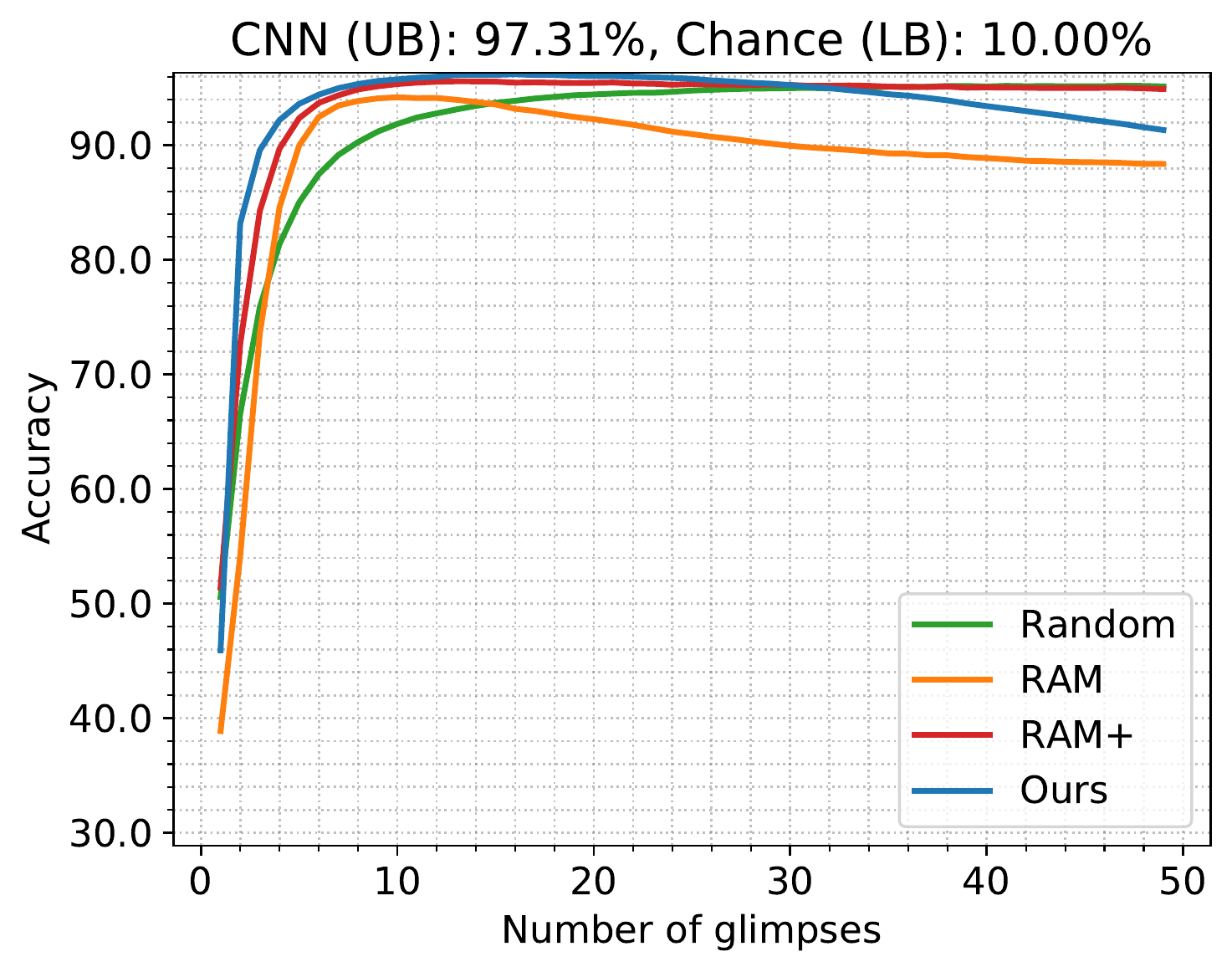}
    \centering{(a)}
    \end{minipage}
    \hfill
    \begin{minipage}[c]{0.32\linewidth}
    \includegraphics[width = \textwidth]{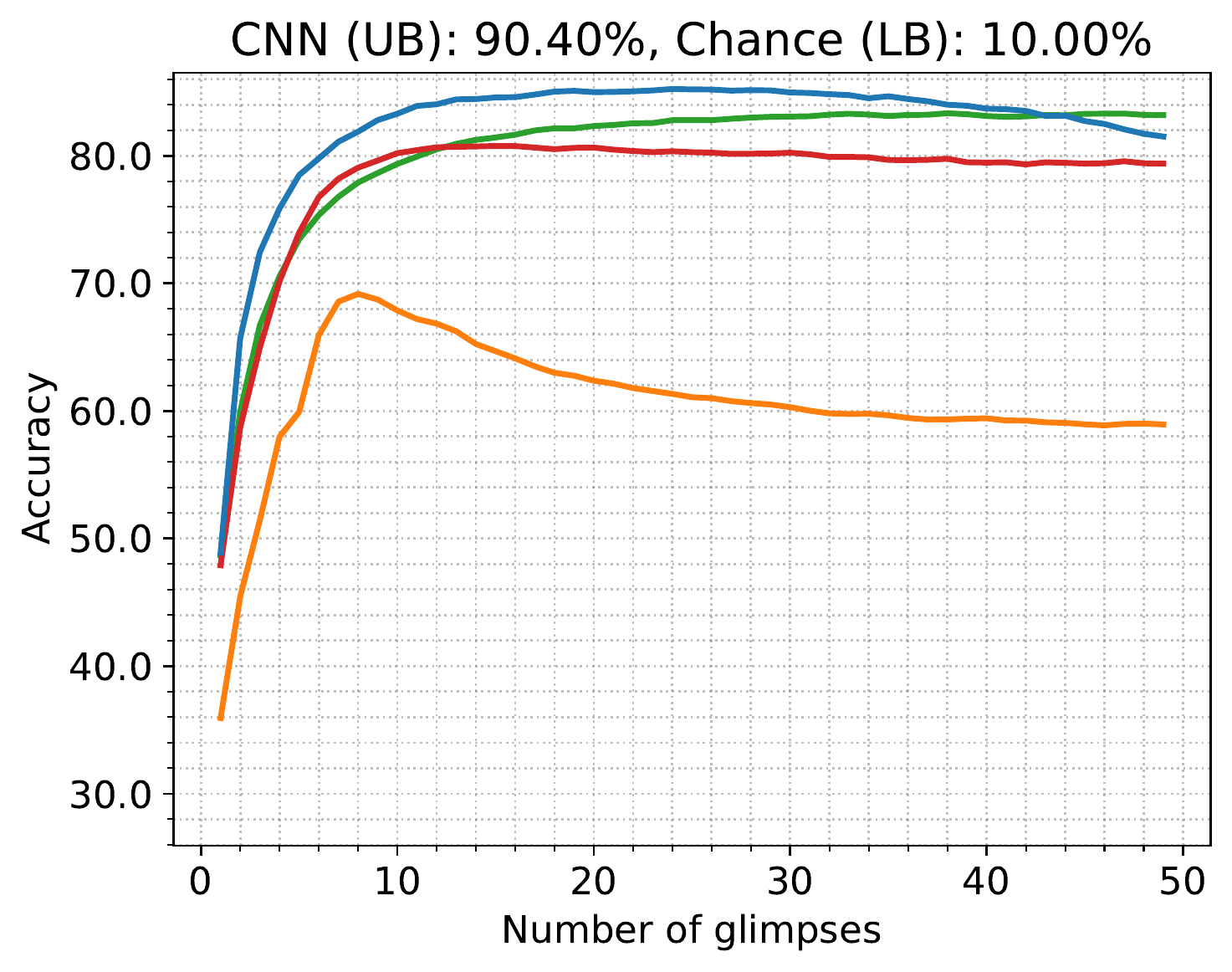}
    \centering{(b)}
    \end{minipage}
    \hfill
    \begin{minipage}[c]{0.32\linewidth}
    \includegraphics[width = \textwidth]{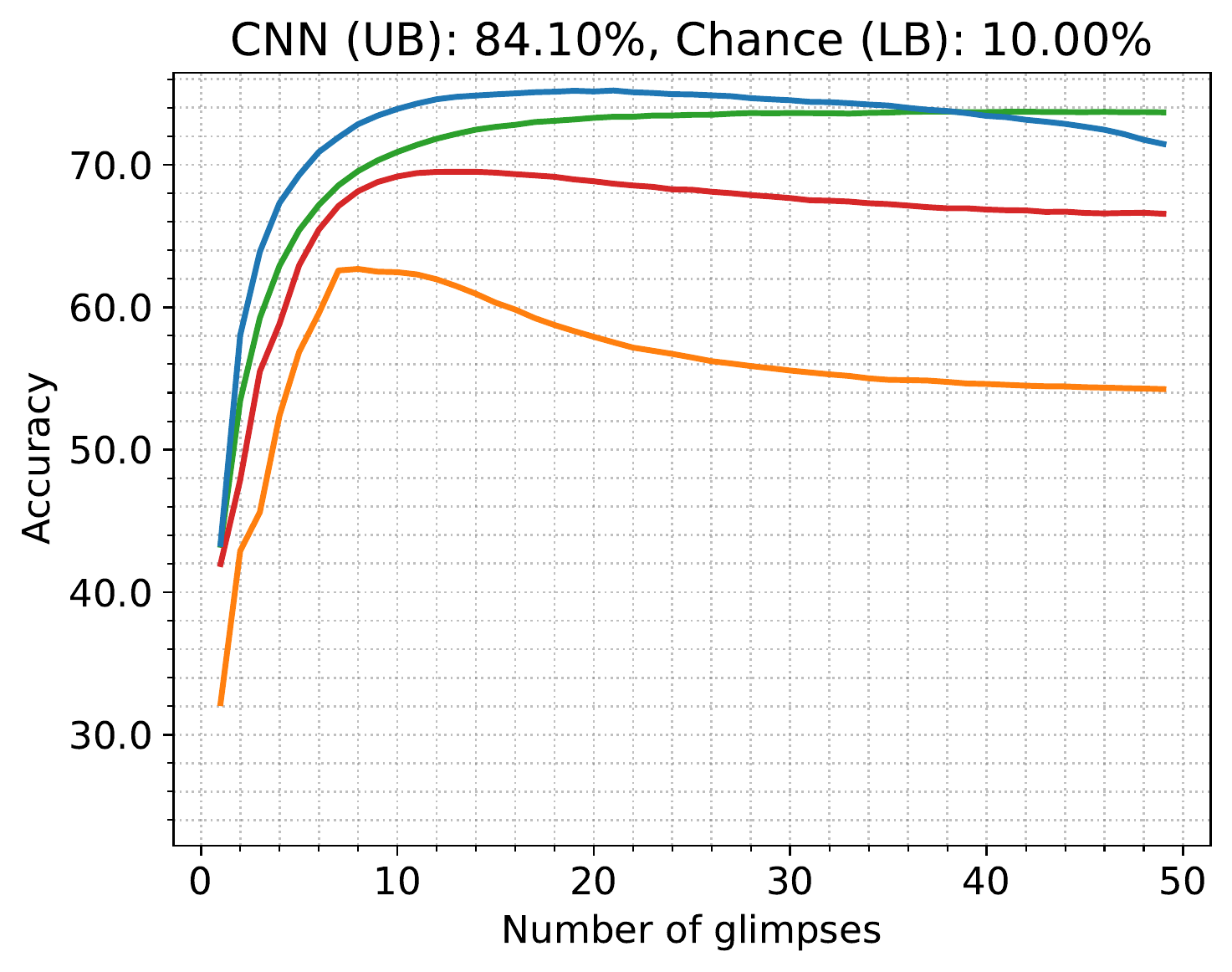}
    \centering{(c)}
    \end{minipage}
    \hfill
    \vfill
    \hfill
    \begin{minipage}[c]{0.32\linewidth}
    \includegraphics[width = \textwidth]{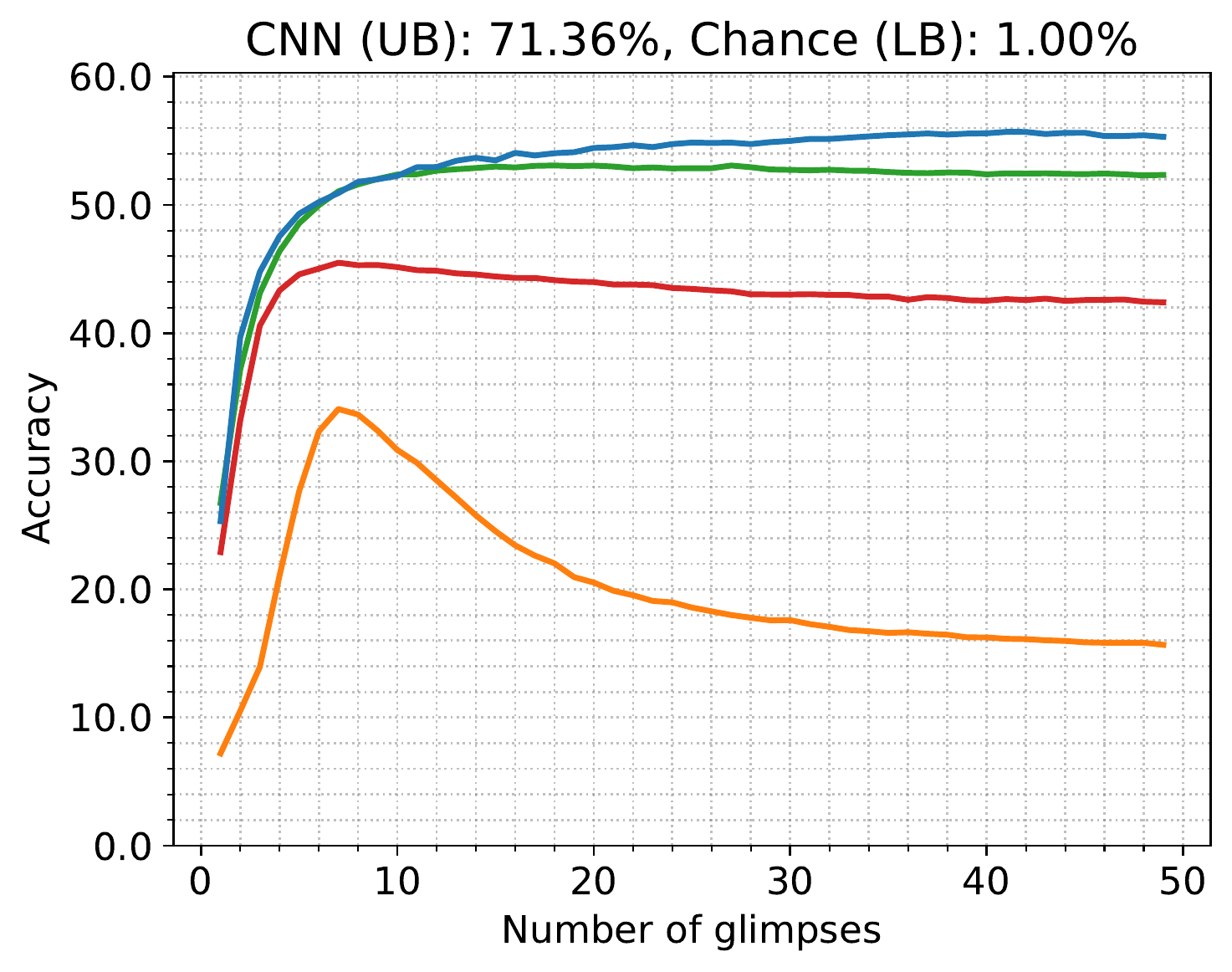}
    \centering{(d)}
    \end{minipage}
    \hfill
    \begin{minipage}[c]{0.32\linewidth}
    \includegraphics[width = \textwidth]{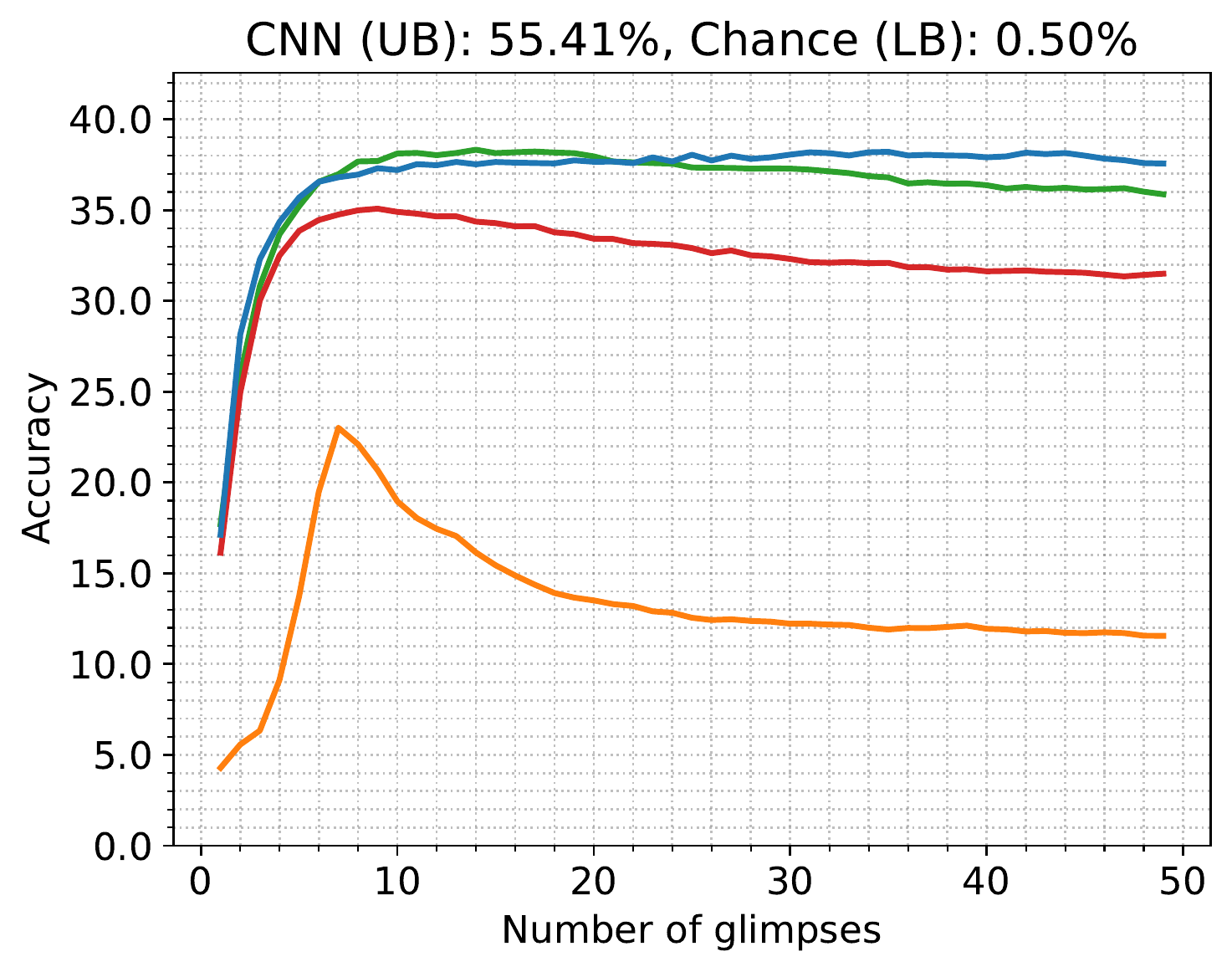}
    \centering{(e)}
    \end{minipage}
    \hfill
    \caption{\textit{Baseline comparison for a large number of glimpses}. (a) SVHN (b) CIFAR-10 (C) CINIC-10 (d) CIFAR-100 (e) TinyImageNet. Results are averaged over three runs. Our model generalizes to a large number of glimpses and achieves the highest accuracy having seen an optimal number of glimpses.}
    \label{fig:largeT}
\end{figure}
\begin{figure}[t]
    \begin{minipage}[t]{0.13\linewidth}
    \vfill
    \includegraphics[width = 0.8\textwidth]{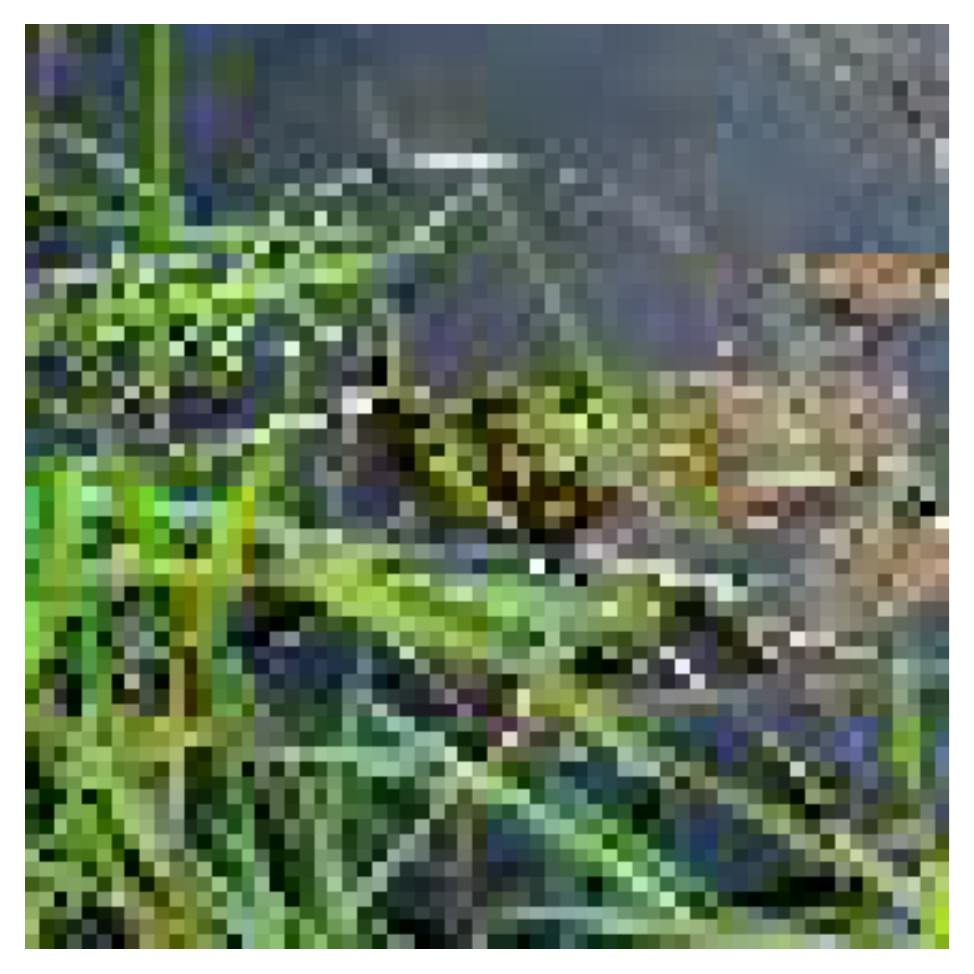}
    \end{minipage}
    \hfill
    \begin{minipage}[t]{0.13\linewidth}
    \centering{t=0}
    \begin{minipage}[c]{\linewidth}
    \hfill
    \includegraphics[width = 0.8\textwidth]{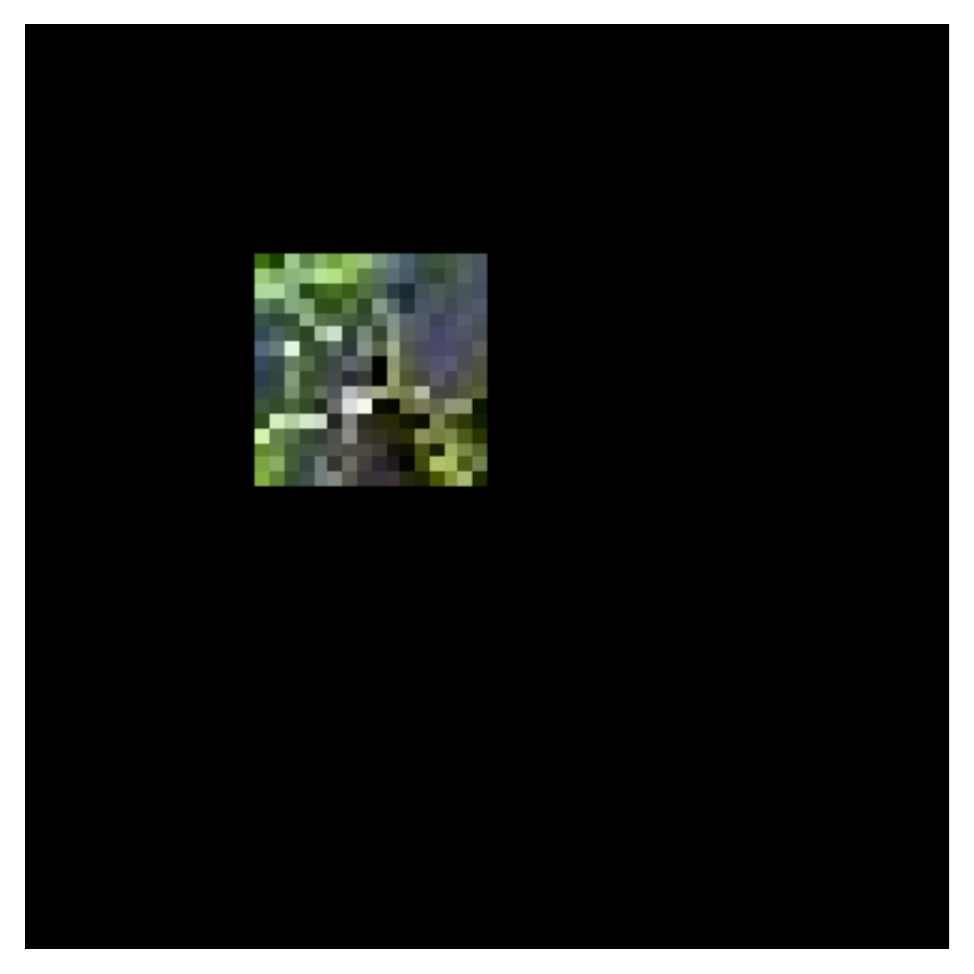}
    \hfill\hfill
    \end{minipage}
    \vfill
    \begin{minipage}[c]{\linewidth}
    \includegraphics[width = \textwidth]{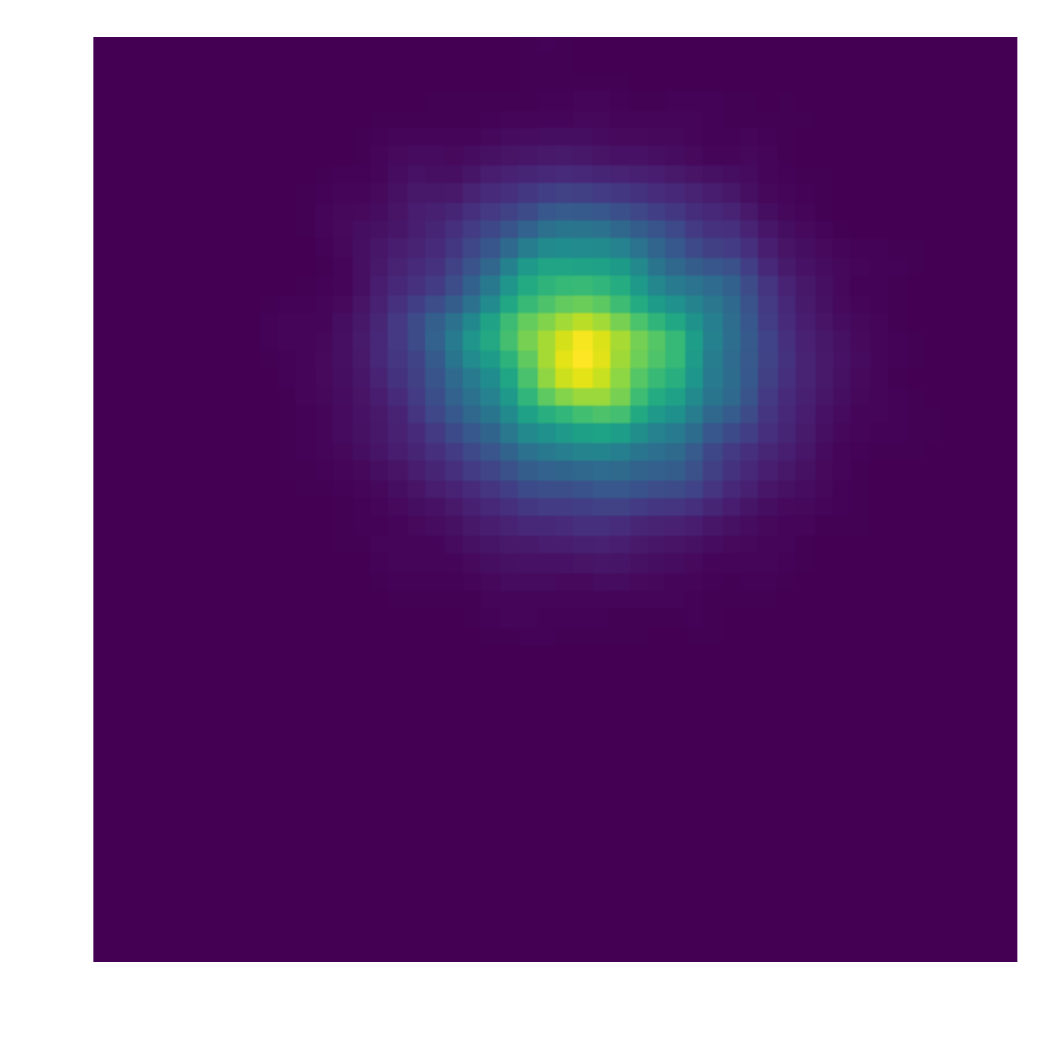}
    \end{minipage}
    \vfill
    \begin{minipage}[c]{\linewidth}
    \includegraphics[width = \textwidth]{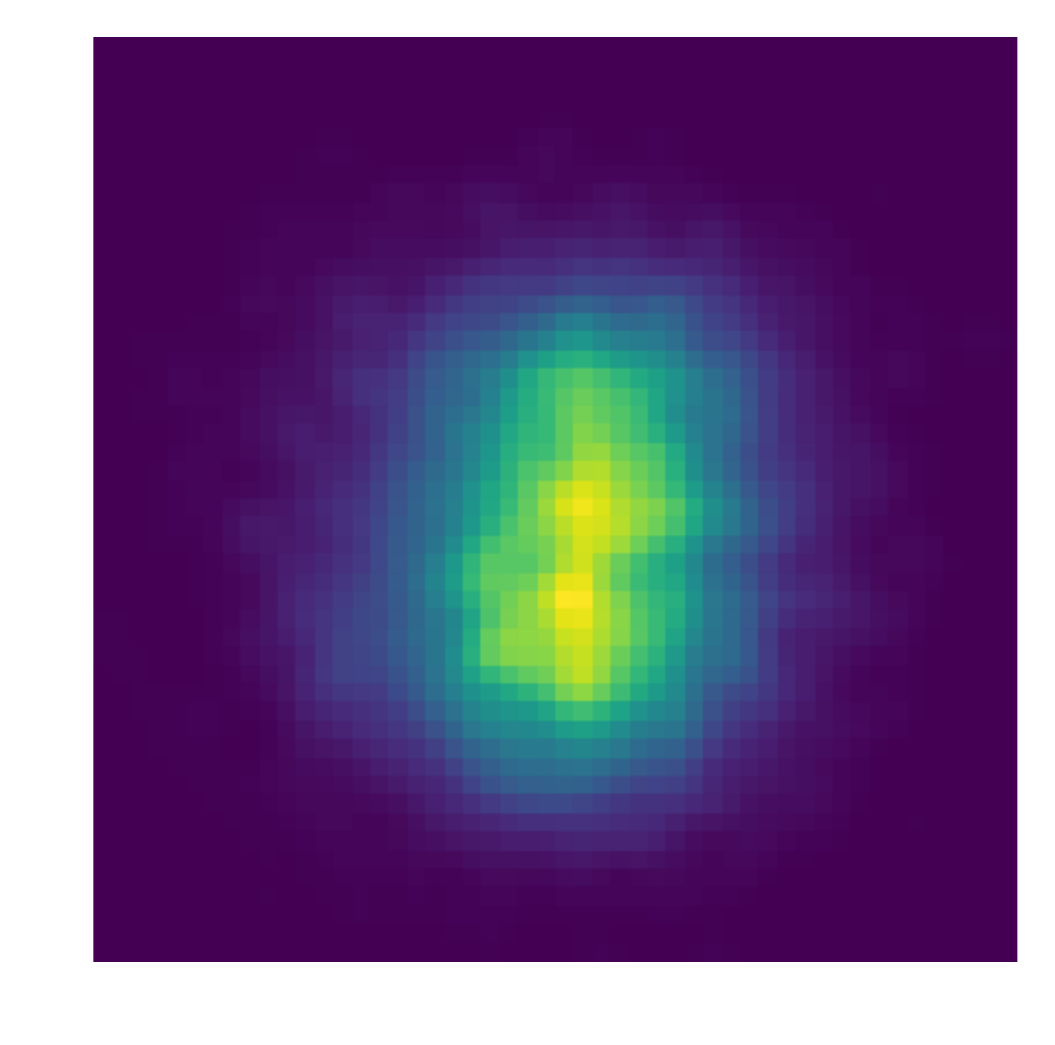}
    \end{minipage}
    \end{minipage}
    \hfill
    \begin{minipage}[t]{0.13\linewidth}
    \centering{t=1}
    \begin{minipage}[c]{\linewidth}
    \hfill
    \includegraphics[width = 0.8\textwidth]{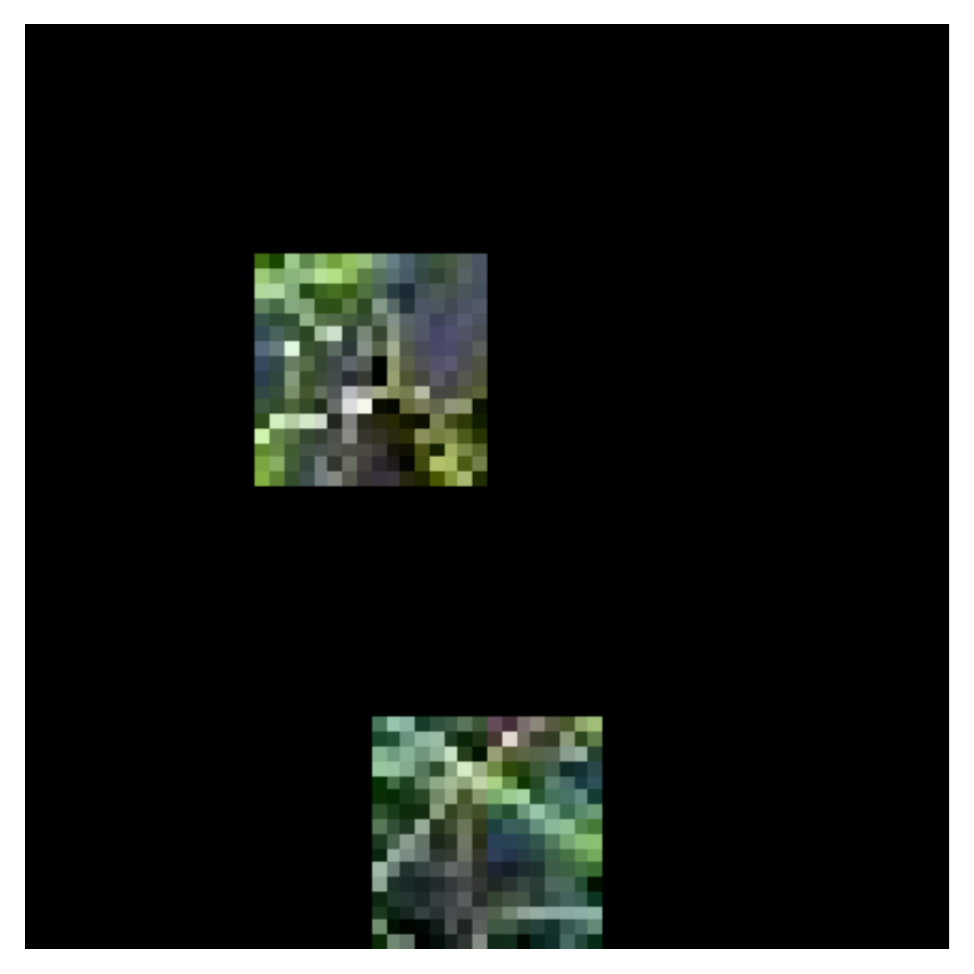}
    \hfill\hfill
    \end{minipage}
    \vfill
    \begin{minipage}[c]{\linewidth}
    \includegraphics[width = \textwidth]{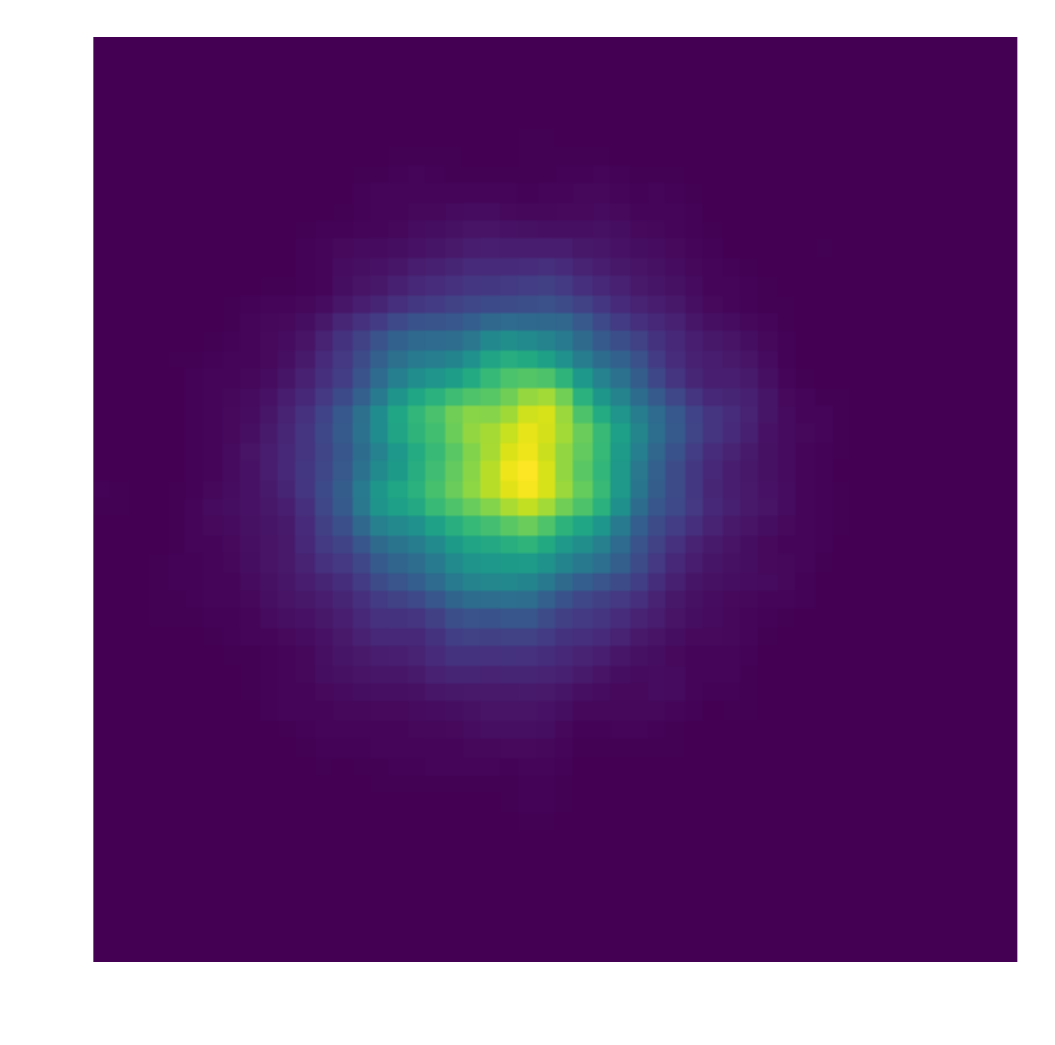}
    \end{minipage}
    \vfill
    \begin{minipage}[c]{\linewidth}
    \includegraphics[width = \textwidth]{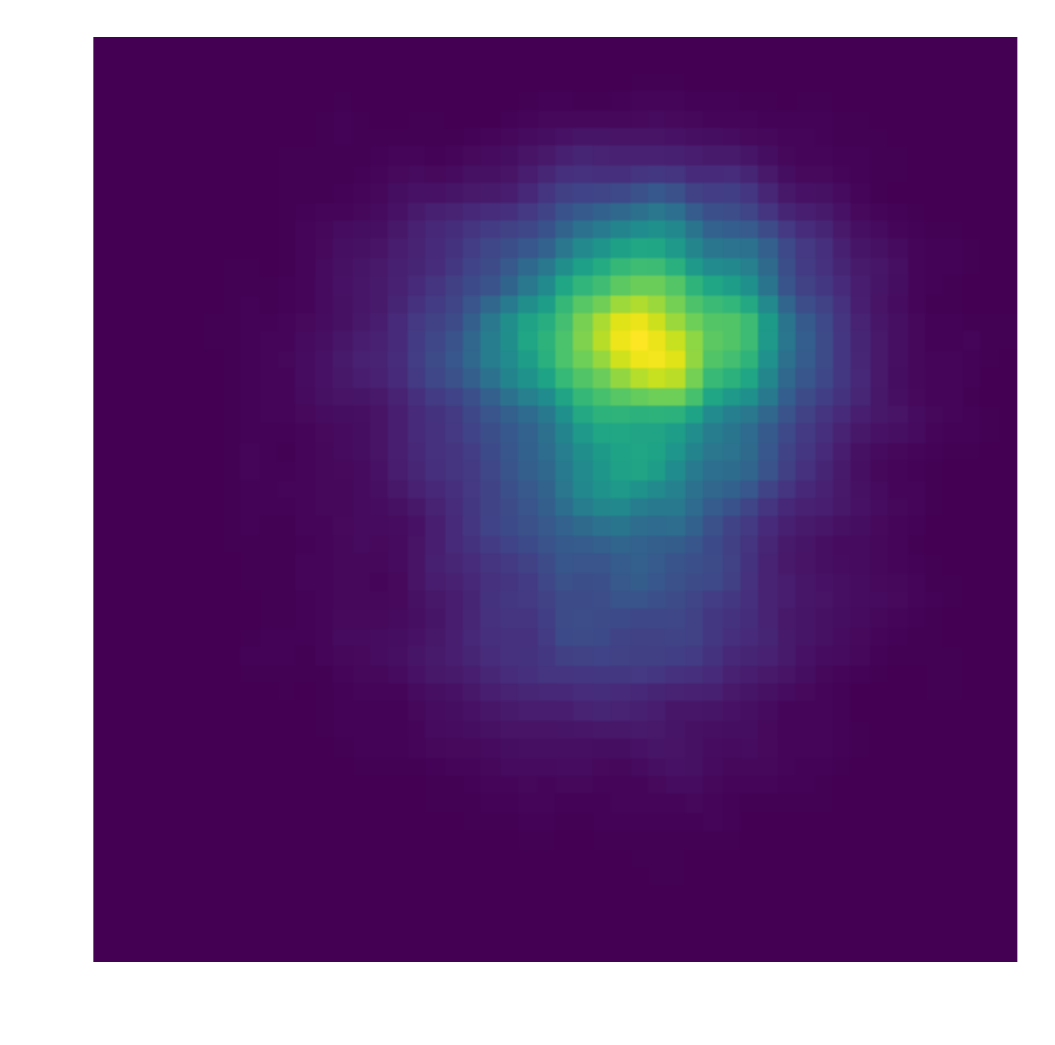}
    \end{minipage}
    \end{minipage}
    \hfill
    \begin{minipage}[t]{0.13\linewidth}
    \centering{t=2}
    \begin{minipage}[c]{\linewidth}
    \hfill
    \includegraphics[width = 0.8\textwidth]{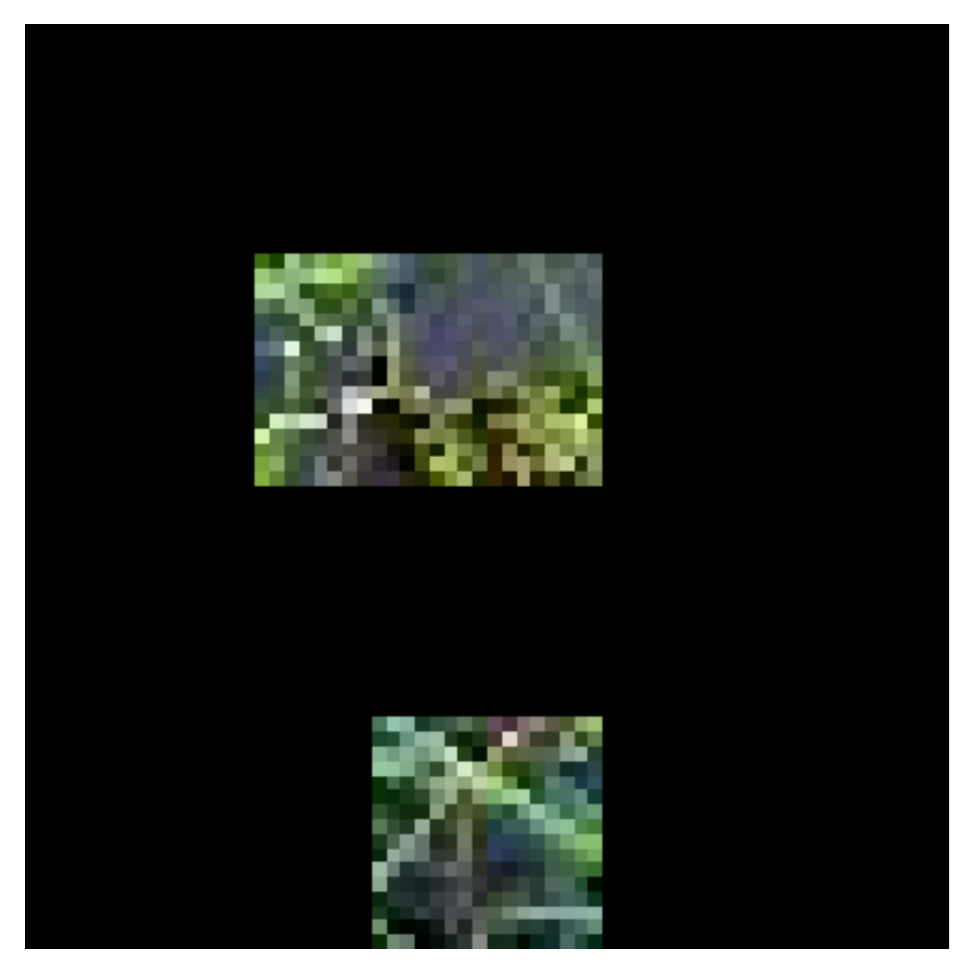}
    \hfill\hfill
    \end{minipage}
    \vfill
    \begin{minipage}[c]{\linewidth}
    \includegraphics[width = \textwidth]{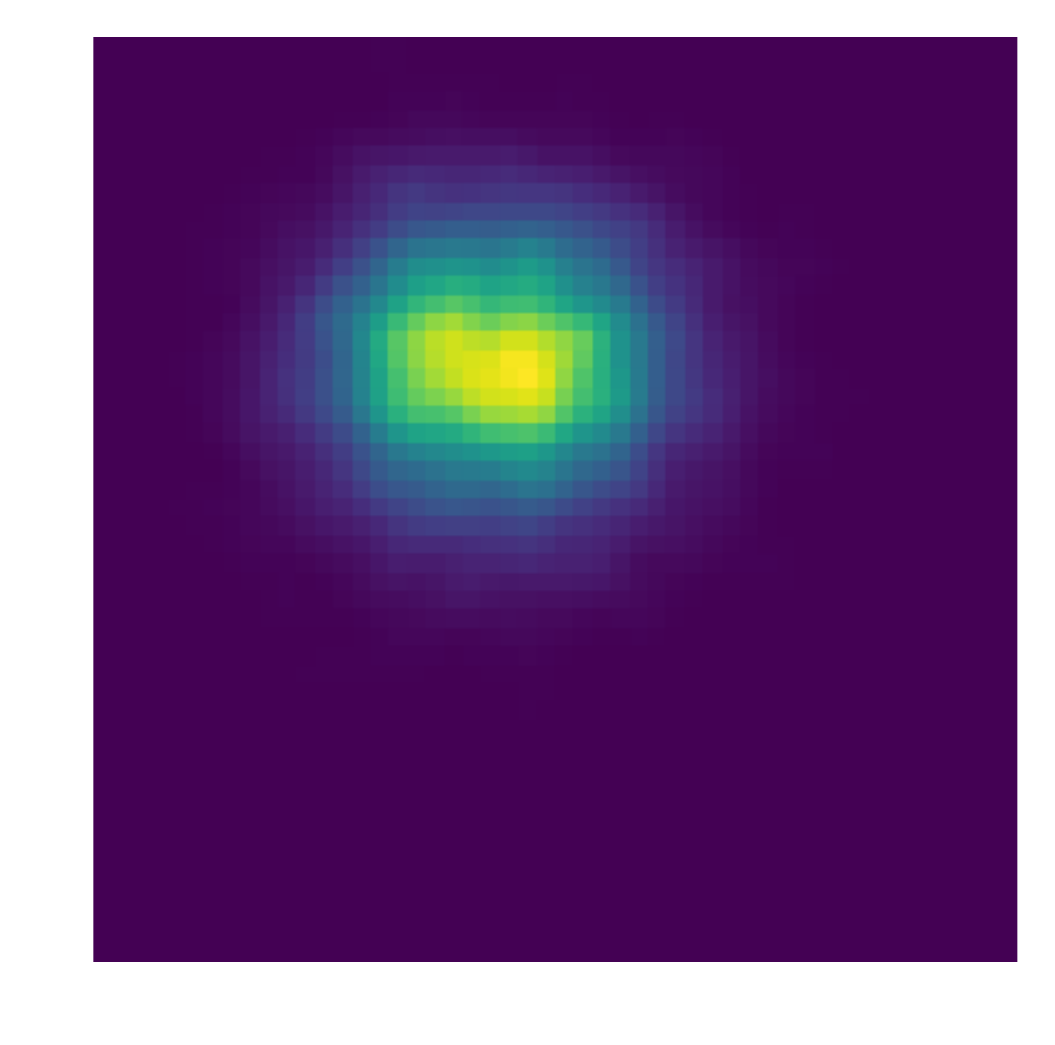}
    \end{minipage}
    \vfill
    \begin{minipage}[c]{\linewidth}
    \includegraphics[width = \textwidth]{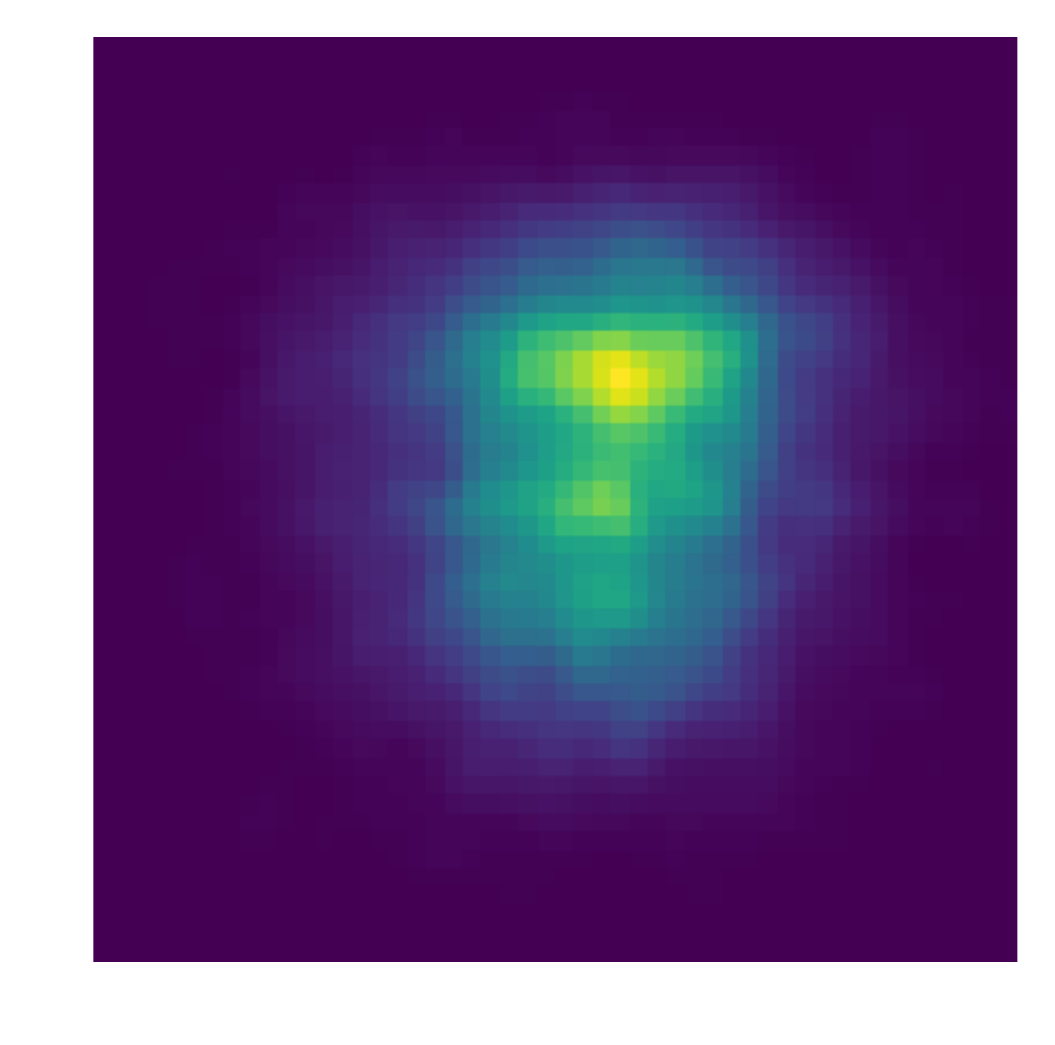}
    \end{minipage}
    \end{minipage}
    \hfill
    \begin{minipage}[t]{0.13\linewidth}
    \centering{t=3}
    \begin{minipage}[c]{\linewidth}
    \hfill
    \includegraphics[width = 0.8\textwidth]{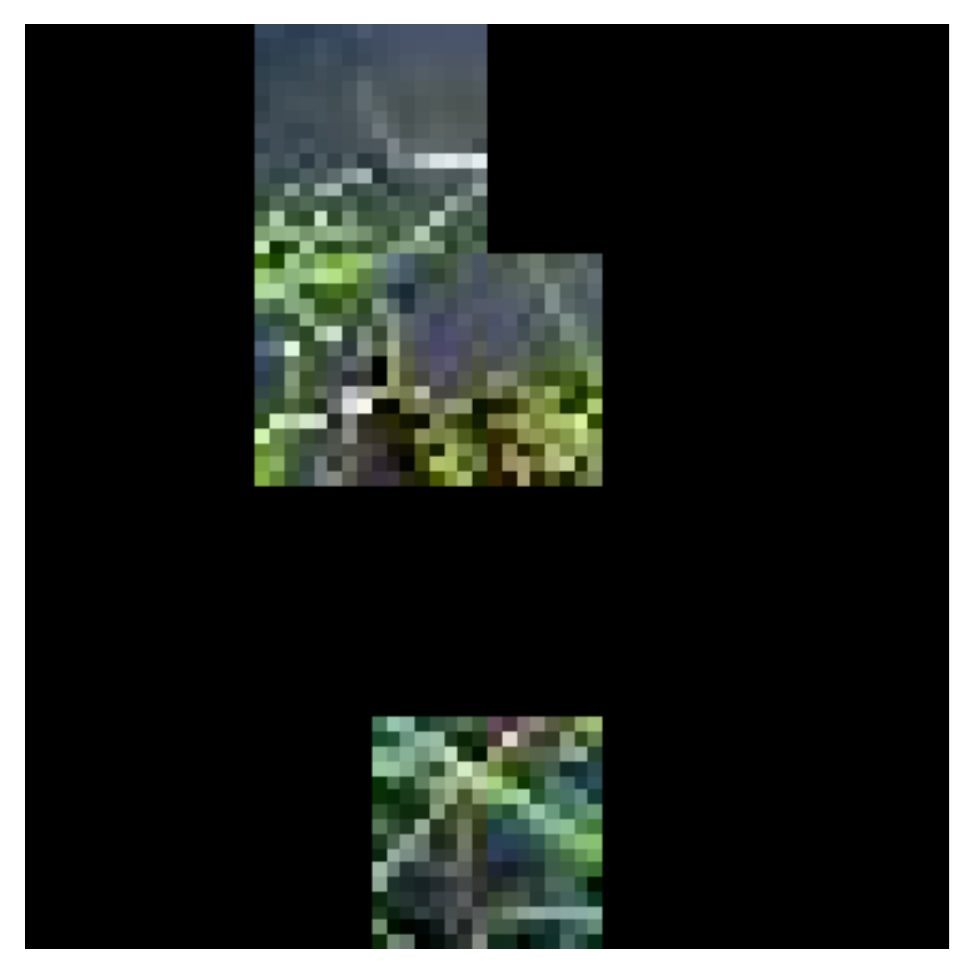}
    \hfill\hfill
    \end{minipage}
    \vfill
    \begin{minipage}[c]{\linewidth}
    \includegraphics[width = \textwidth]{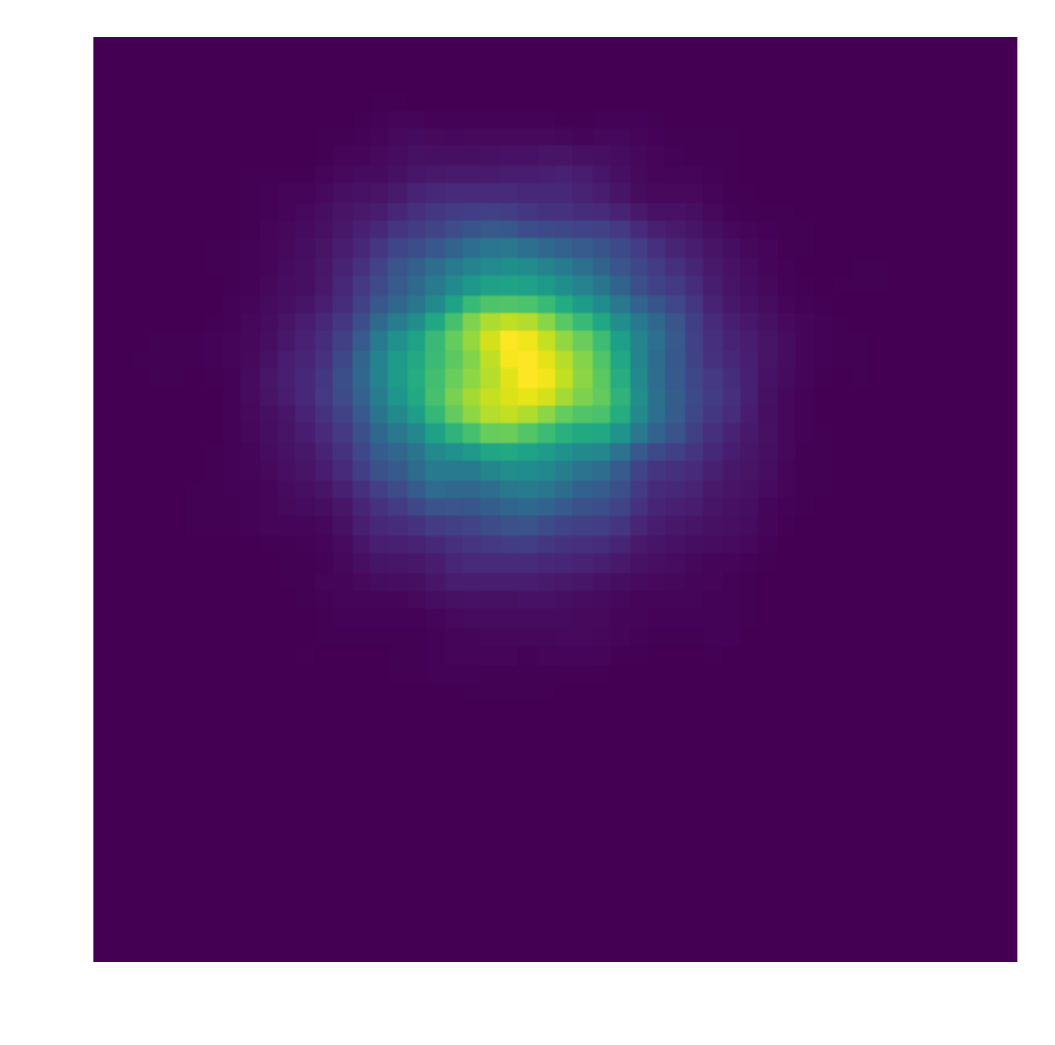}
    \end{minipage}
    \vfill
    \begin{minipage}[c]{\linewidth}
    \includegraphics[width = \textwidth]{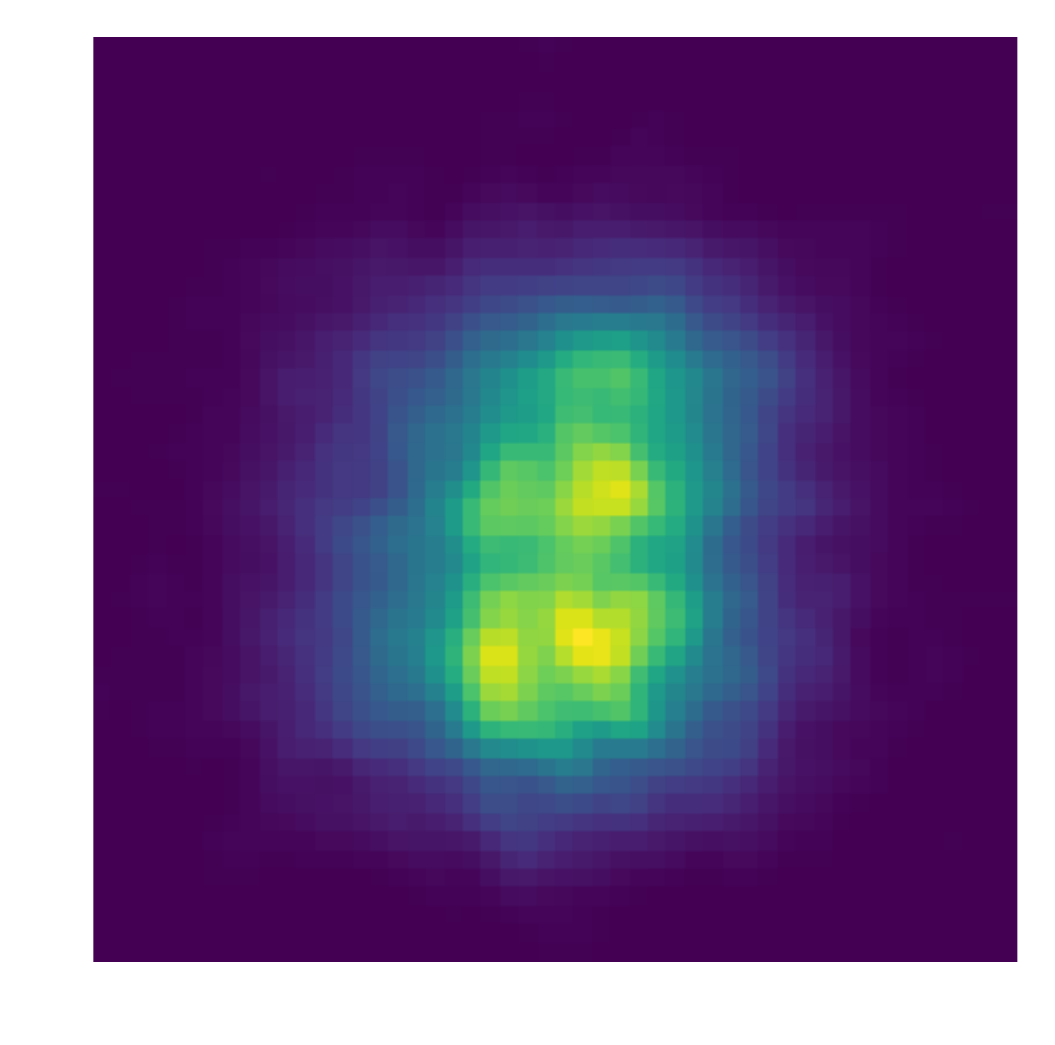}
    \end{minipage}
    \end{minipage}
    \hfill
    \begin{minipage}[t]{0.13\linewidth}
    \centering{t=4}
    \begin{minipage}[c]{\linewidth}
    \hfill
    \includegraphics[width = 0.8\textwidth]{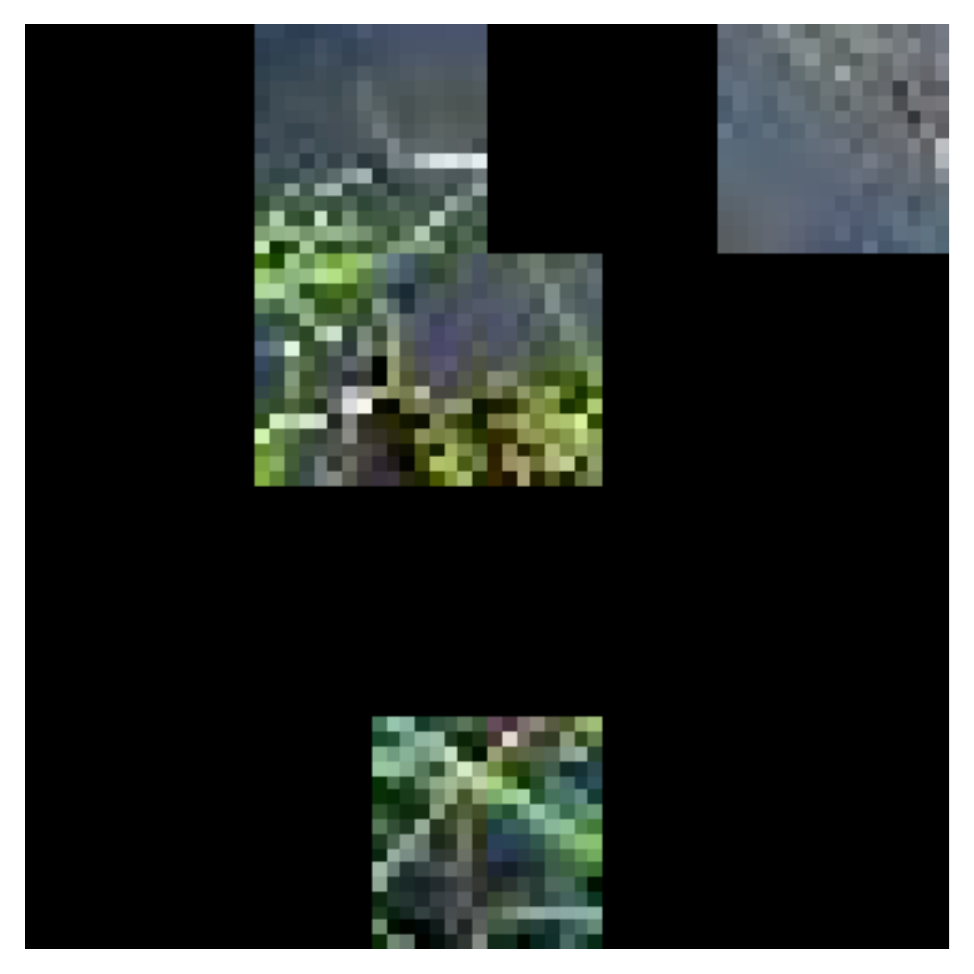}
    \hfill\hfill
    \end{minipage}
    \vfill
    \begin{minipage}[c]{\linewidth}
    \includegraphics[width = \textwidth]{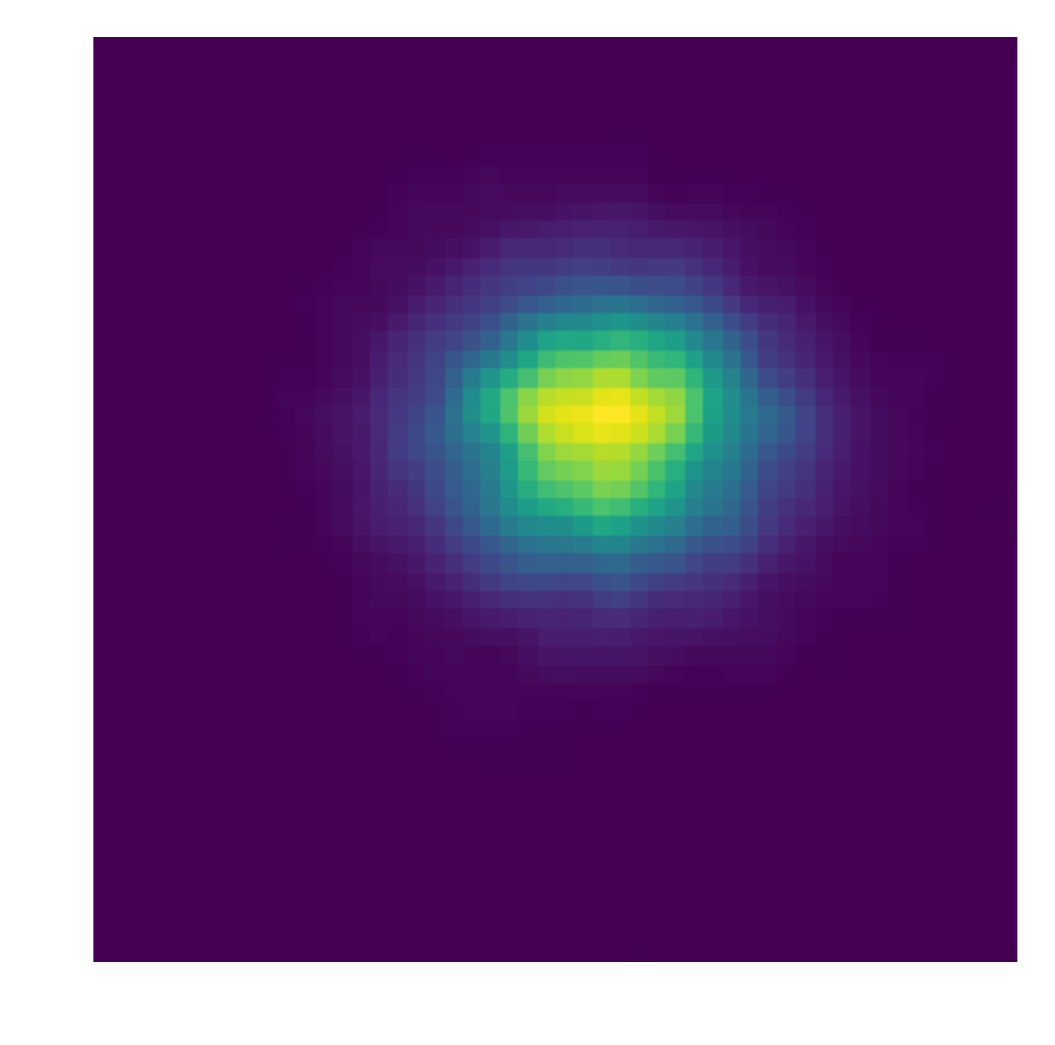}
    \end{minipage}
    \vfill
    \begin{minipage}[c]{\linewidth}
    \includegraphics[width = \textwidth]{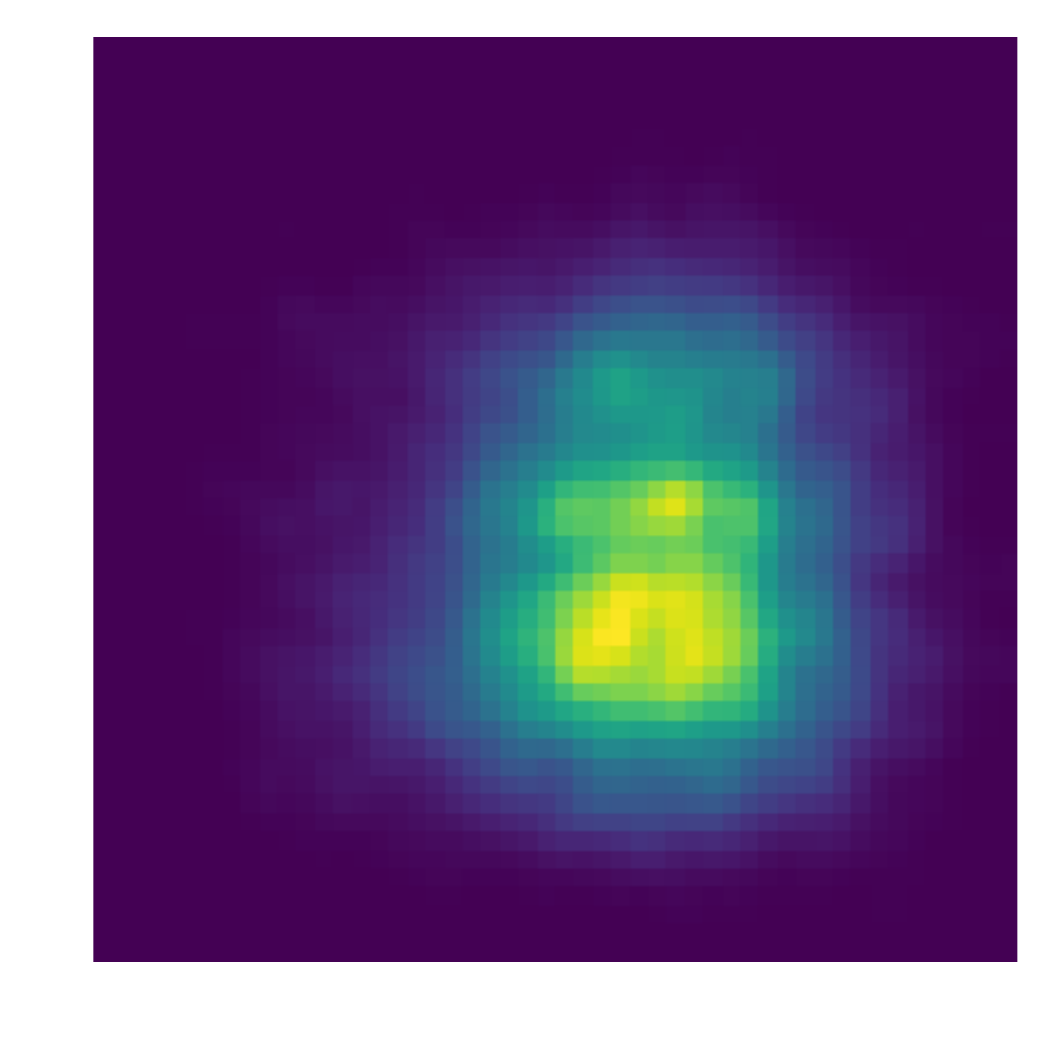}
    \end{minipage}
    \end{minipage}
    \hfill
    \begin{minipage}[t]{0.13\linewidth}
    \centering{t=5}
    \begin{minipage}[c]{\linewidth}
    \hfill
    \includegraphics[width = 0.8\textwidth]{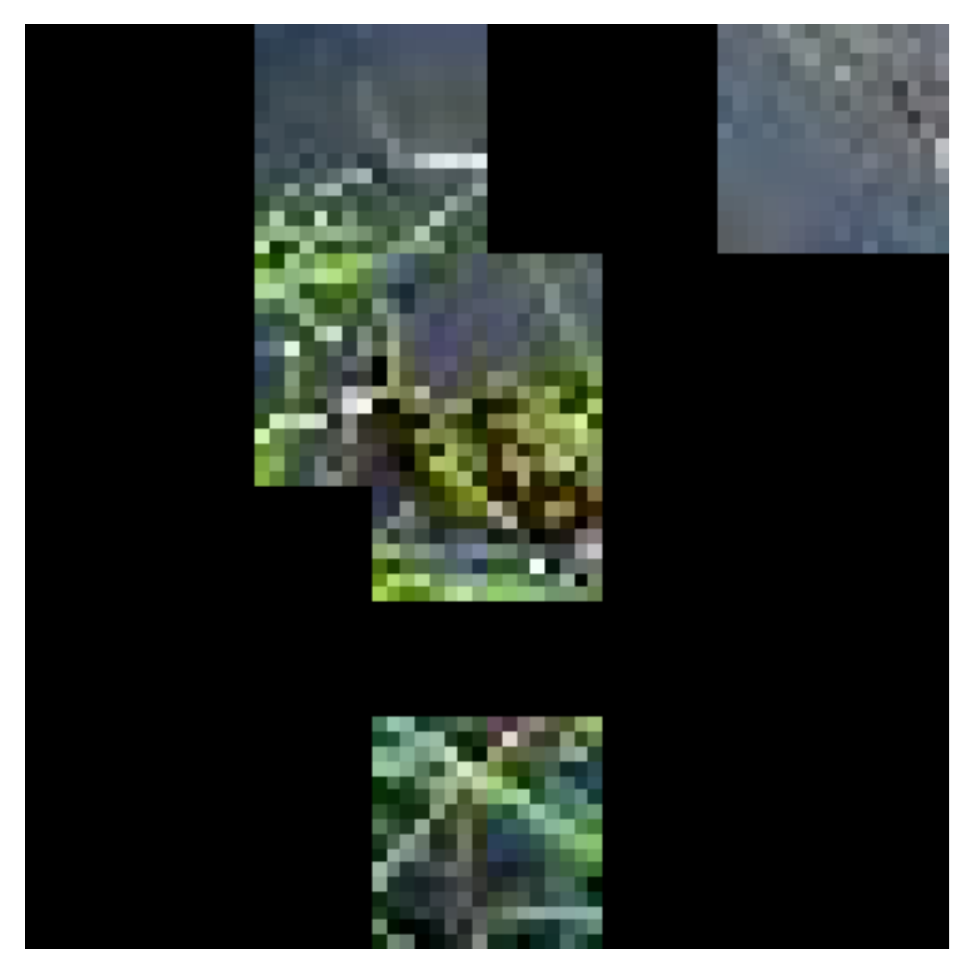}
    \hfill\hfill
    \end{minipage}
    \vfill
    \begin{minipage}[c]{\linewidth}
    \includegraphics[width = \textwidth]{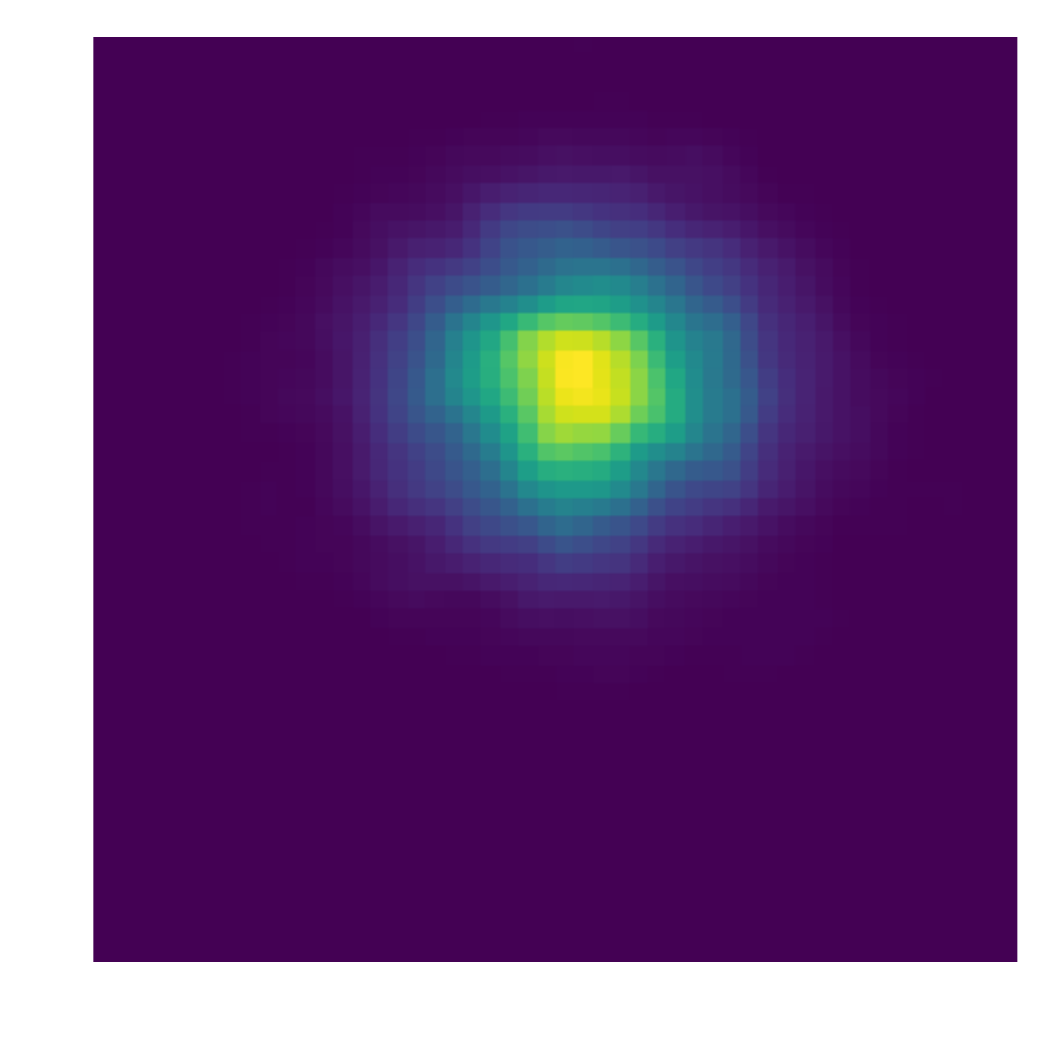}
    \end{minipage}
    \vfill
    \begin{minipage}[c]{\linewidth}
    \includegraphics[width = \textwidth]{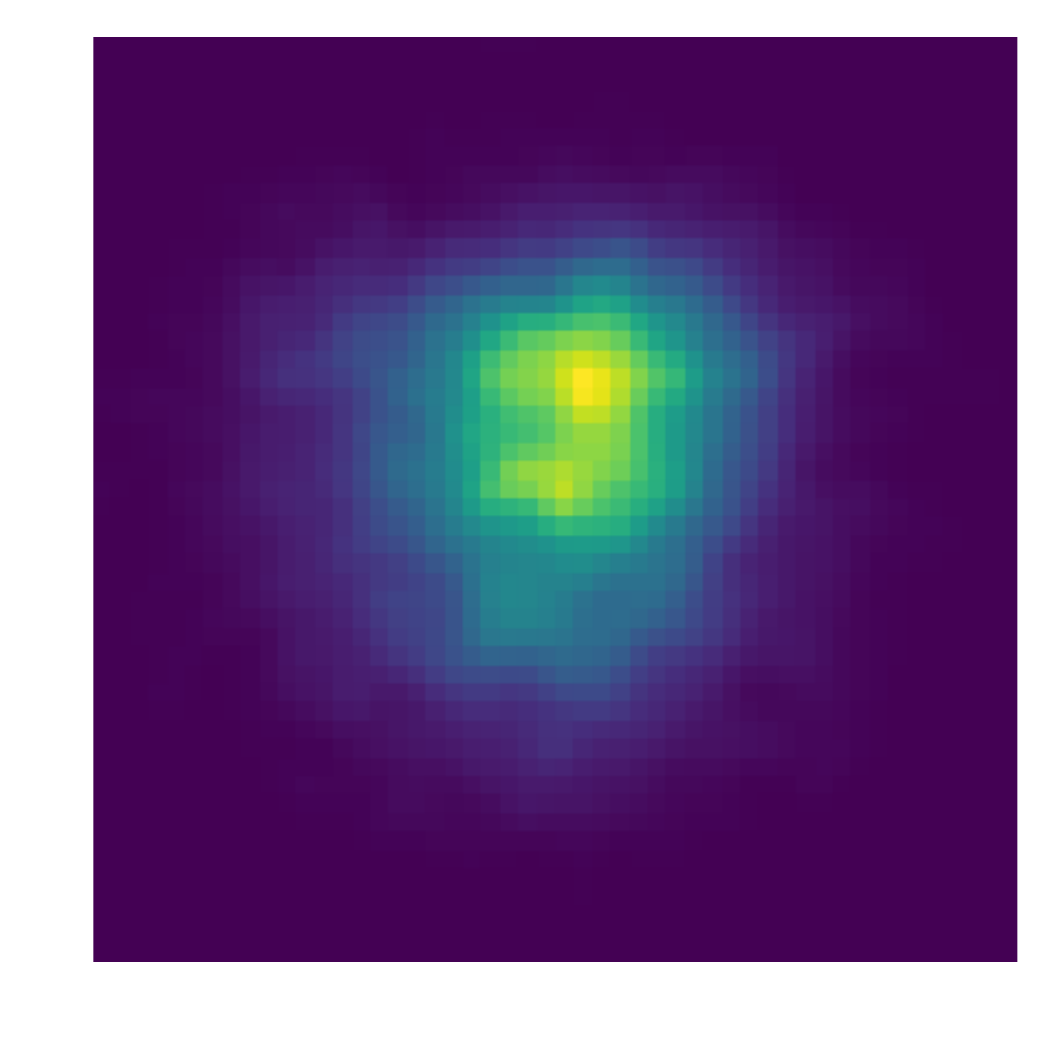}
    \end{minipage}
    \end{minipage}
    \caption{\textit{TSNE projection of $q(z|h_t)$ estimated with and without normalizing flows for an example image from TinyImageNet dataset}. A complete image is shown in the first column for reference. Our model never observes a complete image. In columns two to seven: (top) glimpses observed by the models to compute $h_t$ (middle) TSNE projection of $q(z|h_t)$ estimated without using normalizing flows. (bottom) TSNE projection of $q(z|h_t)$ estimated using normalizing flows. Normalizing flows capture a complex multimodal posterior.}
    \label{fig:tsne}
\end{figure}

\begin{figure}[t!]
    \hfill
    \begin{minipage}[c]{0.40\linewidth}
    \includegraphics[width = \textwidth]{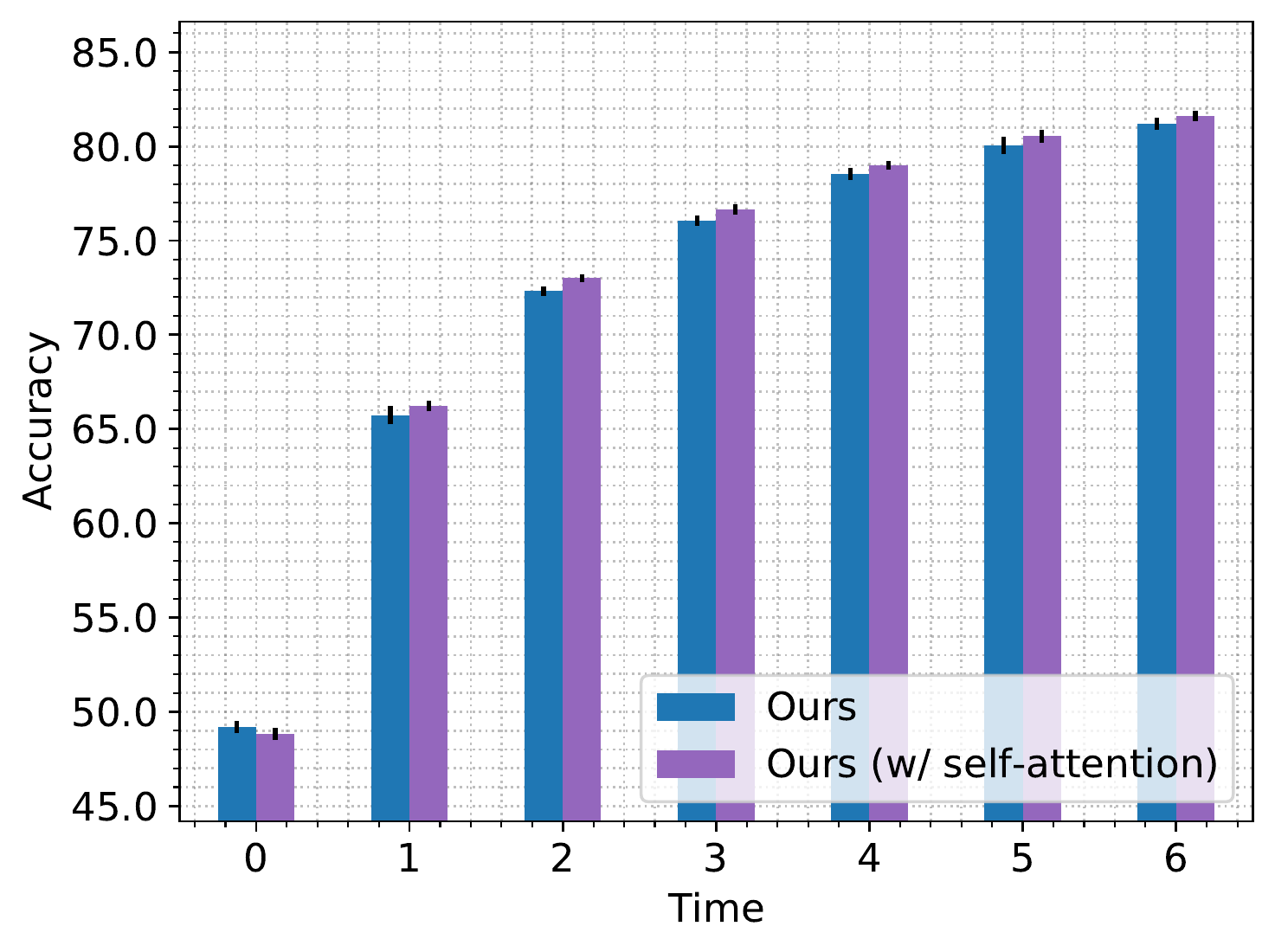}
    \centering{(a)}
    \end{minipage}
    \hfill
    \begin{minipage}[c]{0.40\linewidth}
    \includegraphics[width = \textwidth]{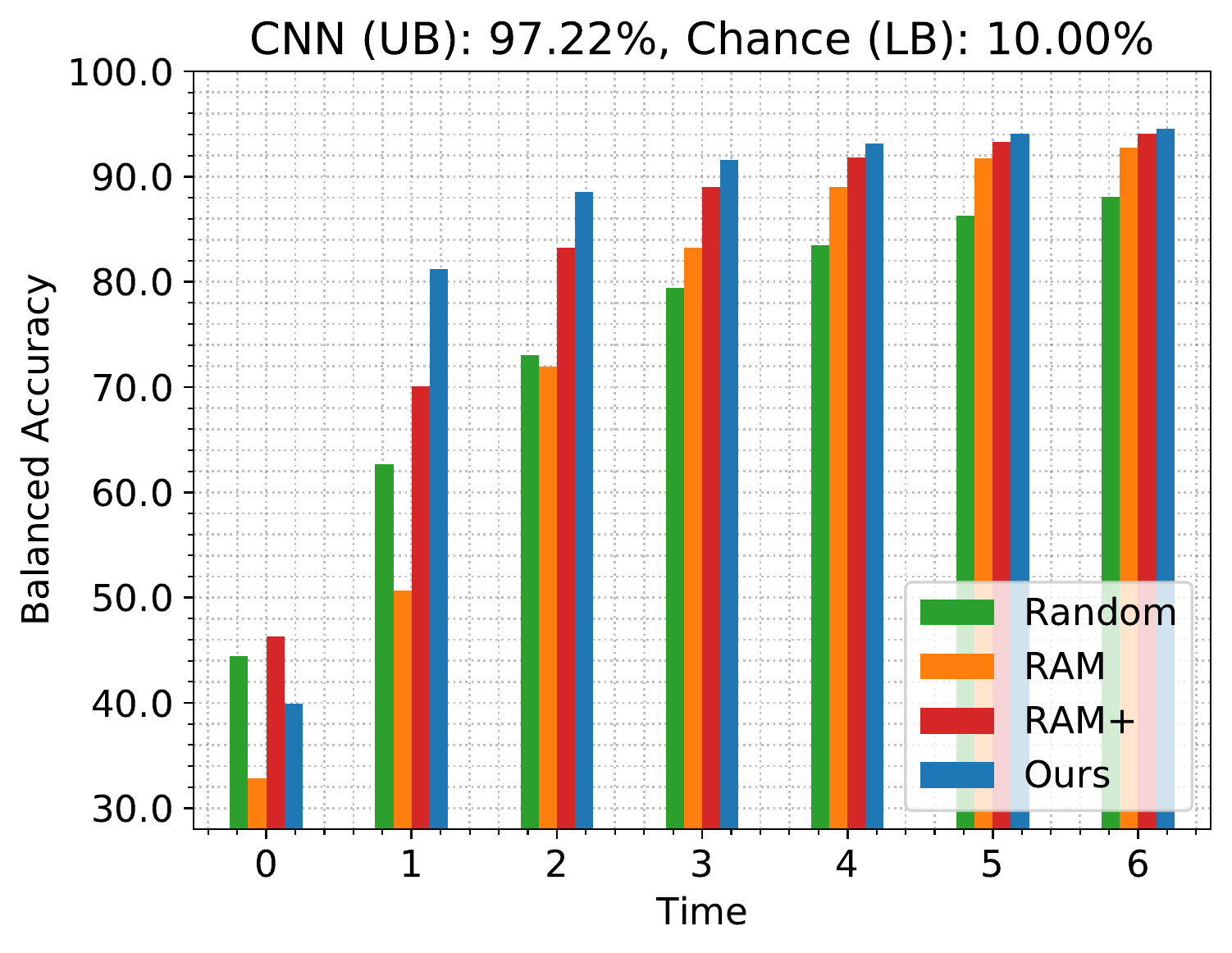}
    \centering{(b)}
    \end{minipage}
    \hfill
    \caption{(a) \textit{Accuracy with and without self-attention in recurrent network $R$ for CIFAR-10.} Results are averaged over ten runs. (b) Baseline comparison using Balanced accuracy for SVHN. We observe the same trend as the one observed in section 4.1 in main paper.}
    \label{fig:attn}
\end{figure}
\begin{figure}[t!]
    \begin{minipage}[c]{\linewidth}
    \includegraphics[width = \textwidth]{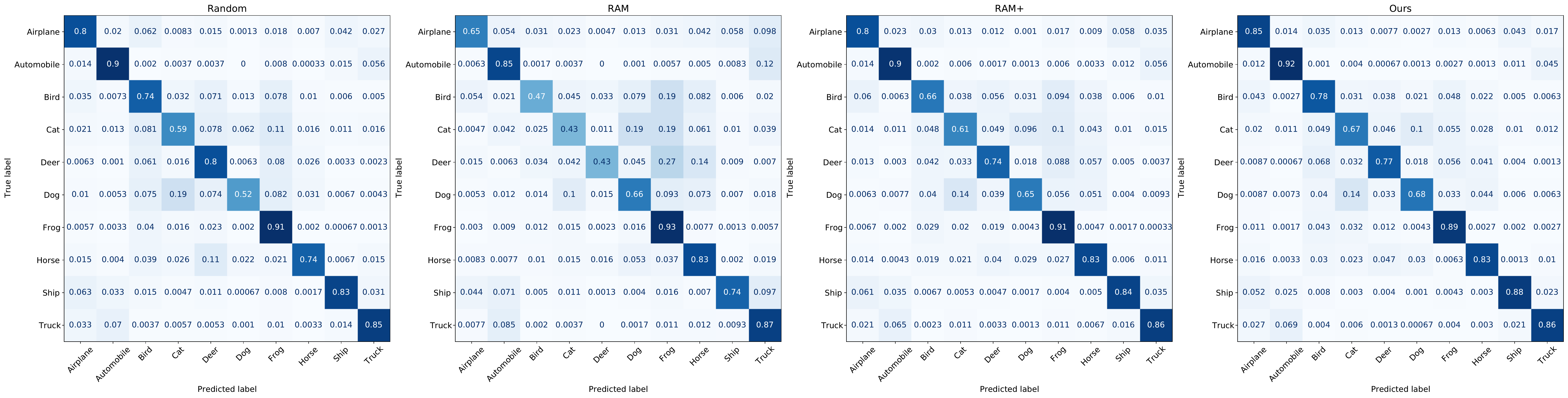}
    \end{minipage}
    \vfill
    \begin{minipage}[c]{\linewidth}
    \centering
    \includegraphics[width = 0.8\textwidth]{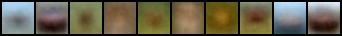}\\
    \hspace{-0.5cm}\centering{\small Airplane Automobile Bird \;\;\; Cat \;\;\; Deer \;\;\; Dog \;\;\; Frog \;\;\; Horse \;\; Ship \;\;\; Truck}
    \end{minipage}
    \caption{(Top) Confusion matrix at $t=6$ and (Bottom) average image per class from CIFAR-10.}
    \label{fig:conf_mat}
\end{figure}
\section{Additional Experiments}

\subsection{Accuracy as a function of area observed in an image} At a time, a hard attention model observes only a small region of the input image through a glimpse. As time progresses, the model observes more area in the image through multiple glimpses. We compare the accuracy of various models as a function of the area observed in an image (see Figure \ref{fig:areaacc}). RAM+ consistently outperforms RAM on various datasets. We observe that the Random baseline performs better than the RAM and RAM+ on comparatively more challenging datasets. As discussed in the main paper, the Random baseline dedicates an entire latent space to the classifier. Our model, RAM+, and RAM use a common latent space for a classifier and a Partial VAE or a controller. Finally, for any given value of the observed area, our model achieves the highest accuracy. The result suggests that our model captures glimpses on the regions that provide more useful information about the class label than the baseline methods.

\subsection{Entropy in the class-logits as a function of number of glimpses} An efficient hard attention model should become more confident in predicting a class label with small number of glimpses. We measure confidence using entropy in the predicted class-logits; lower entropy indicates higher confidence. Figure \ref{fig:conf} shows entropy in the class-logits predicted by various models with time. Irrespective of the glimpse acquisition strategy, all models reduce entropy in their predictions as they acquire more glimpses. In case of RAM, the trend is inconsistent for initial time steps. Recall that RAM optimizes loss only at the last time step, unlike other models that optimize losses for all time steps. Consistent with the previous analyses, RAM+ outperforms RAM on all datasets, and the Random baseline outperforms RAM and RAM+ on complex datasets. Our model achieves lower entropy with smaller number of glimpses compared to the baseline methods. The result indicates that our model acquires glimpses that help the most in reducing the uncertainty in the class-label prediction.

\subsection{Generalization to a large number of glimpses} Recall that all models are trained for seven glimpses. In Figure \ref{fig:largeT}, we assess their inference-time generalizability for a large number of glimpses by testing the models for T=49. Note that our method would have observed an entire image by then. We notice that, during the inference-time, RAM does not generalize well to a sequence of glimpses that are longer than the training time; \cite{elsayed2019saccader,sermanet2014attention,lu2019revisit} also make a similar observation. Our method and RAM+ show greater generalizability than RAM. While the Random baseline is the most generalizable, our method outperforms it with an optimal number of glimpses. The accuracy achieved by our model with an optimal number of glimpses is lower than the CNN; however, the CNN is trained exclusively on complete images, and our model is not. Note that the primary motivation for the attention mechanism is to achieve higher accuracy using limited time and constrained resources. If time and resources are available, one can instead collect a large number of random glimpses and use a non-attentive CNN.

\begin{figure}
    \centering
    \begin{minipage}[c]{0.82\linewidth}
    \includegraphics[width = \textwidth]{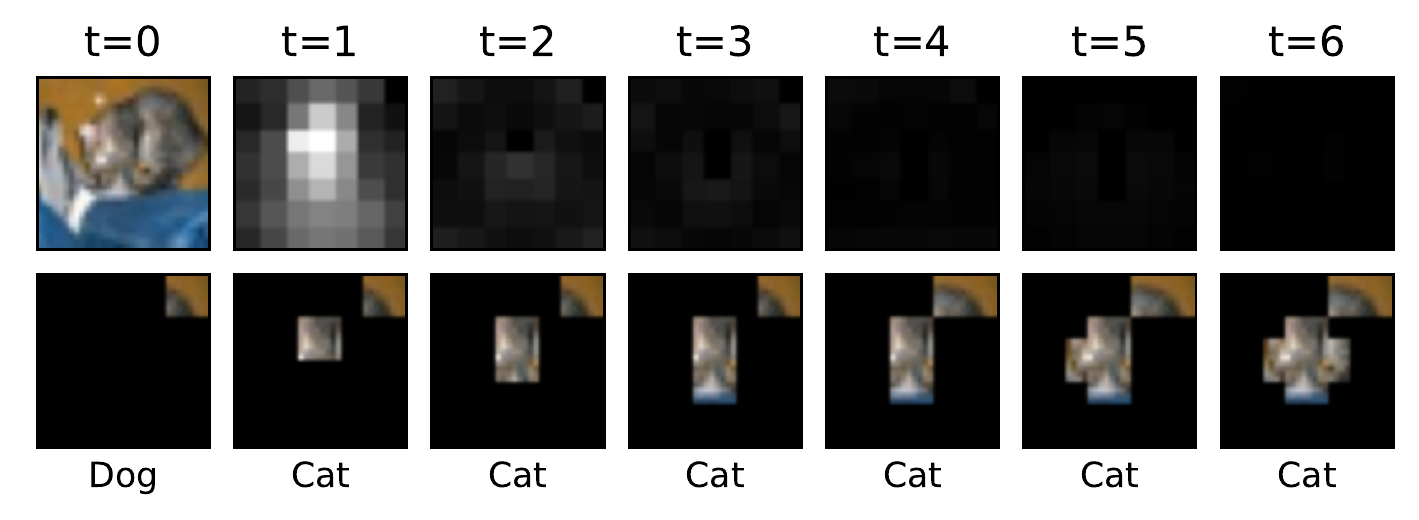}
    \end{minipage}
    \vfill
    \begin{minipage}[c]{0.82\linewidth}
    \includegraphics[width = \textwidth]{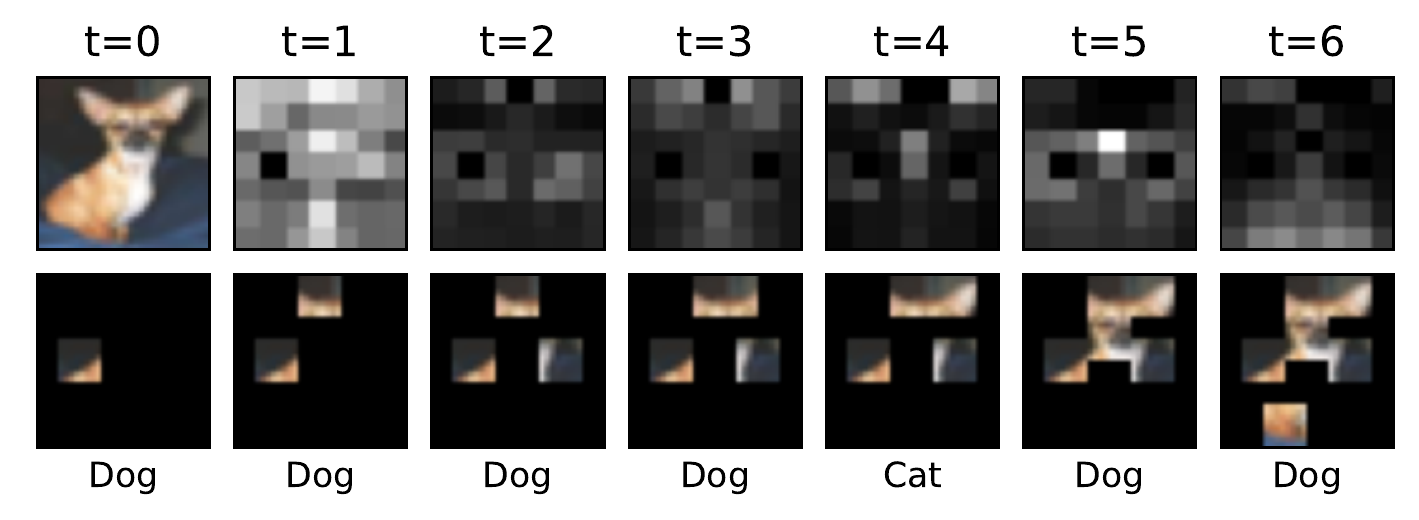}
    \end{minipage}
    \vfill
    \begin{minipage}[c]{0.82\linewidth}
    \includegraphics[width = \textwidth]{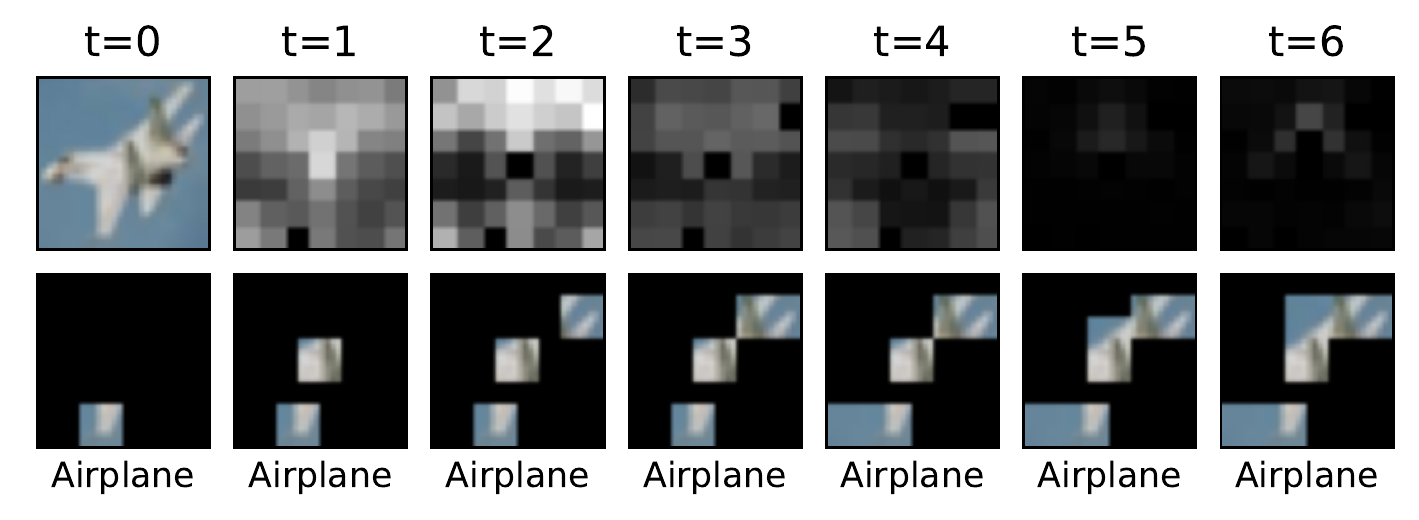}
    \end{minipage}
    \vfill
    \begin{minipage}[c]{0.82\linewidth}
    \includegraphics[width = \textwidth]{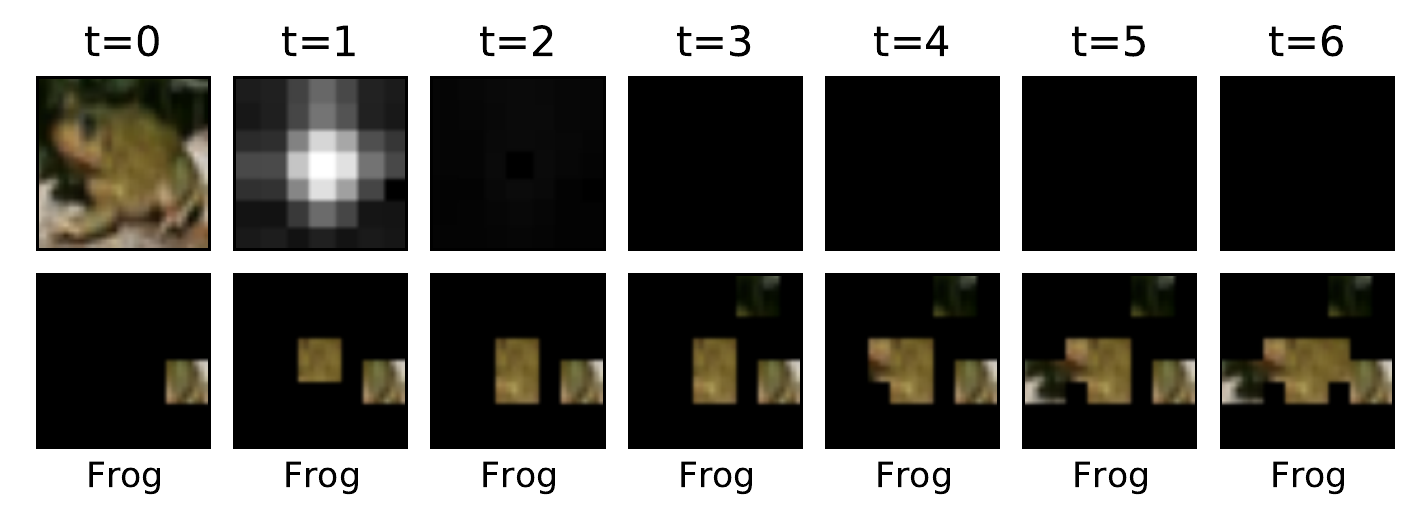}
    \end{minipage}
    \vfill
    \begin{minipage}[c]{0.82\linewidth}
    \includegraphics[width = \textwidth]{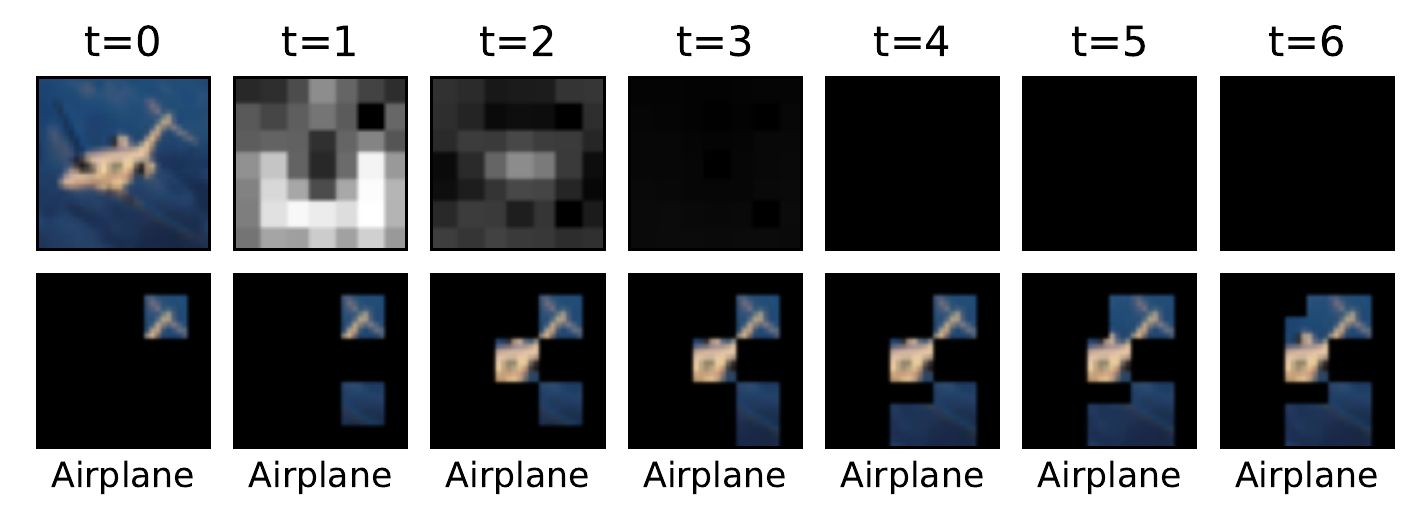}
    \end{minipage}
    \caption{\textit{Additional visual results on randomly chosen images from CIFAR-10 dataset}.}
    \label{fig:viz1}
\end{figure}
\begin{figure}
    \centering
    \begin{minipage}[c]{0.82\linewidth}
    \includegraphics[width = \textwidth]{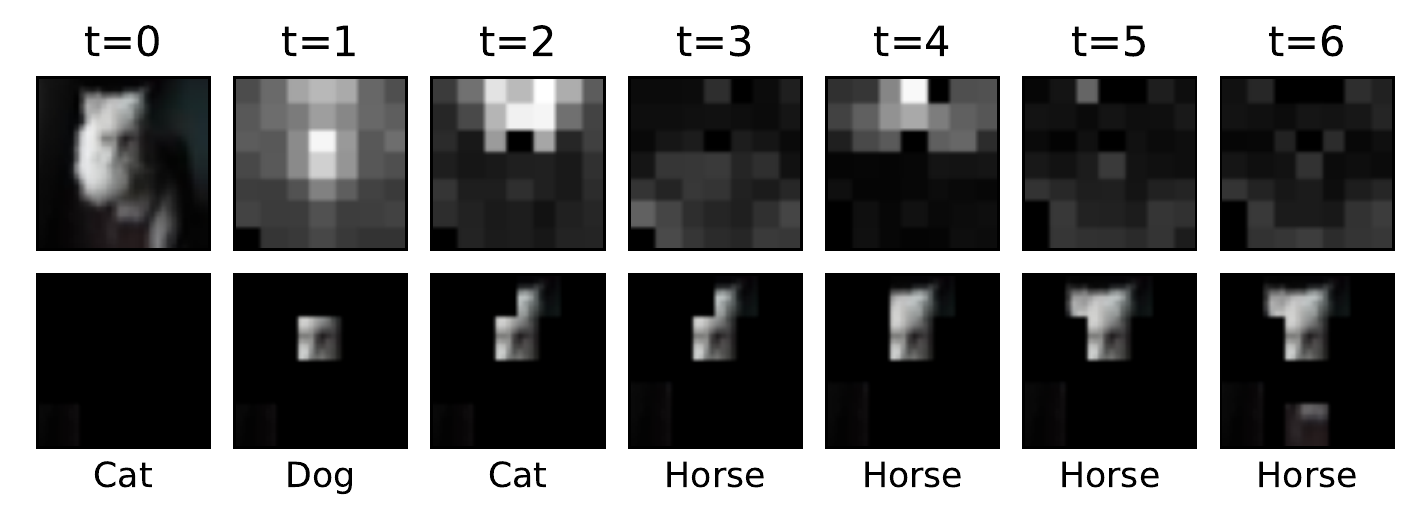}
    \end{minipage}
    \vfill
    \begin{minipage}[c]{0.82\linewidth}
    \includegraphics[width = \textwidth]{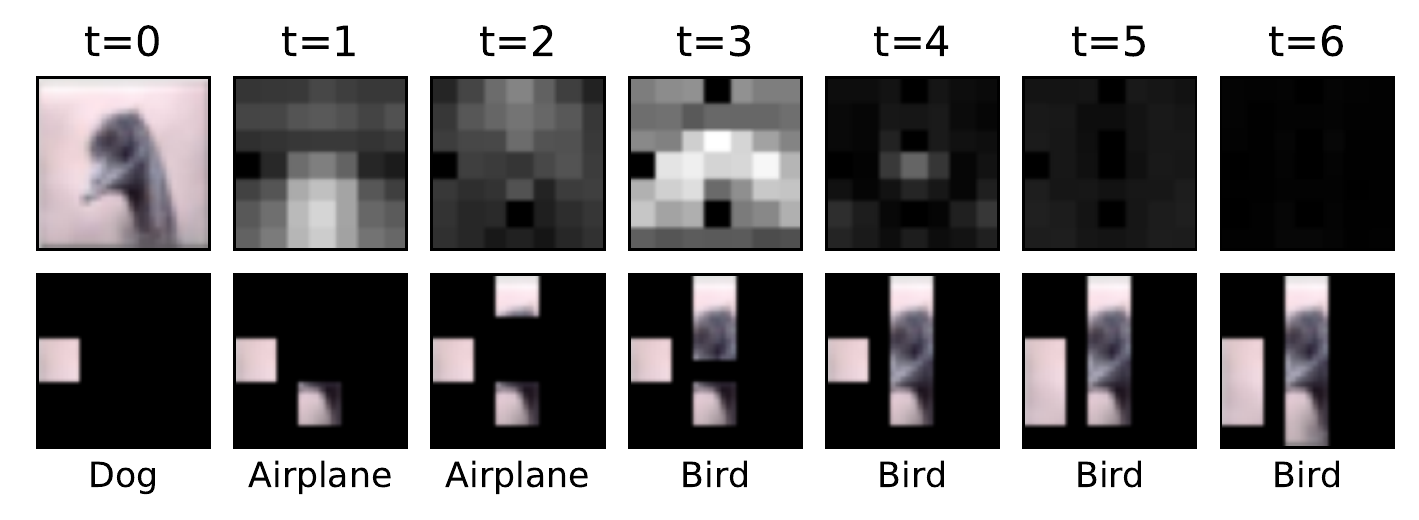}
    \end{minipage}
    \vfill
    \begin{minipage}[c]{0.82\linewidth}
    \includegraphics[width = \textwidth]{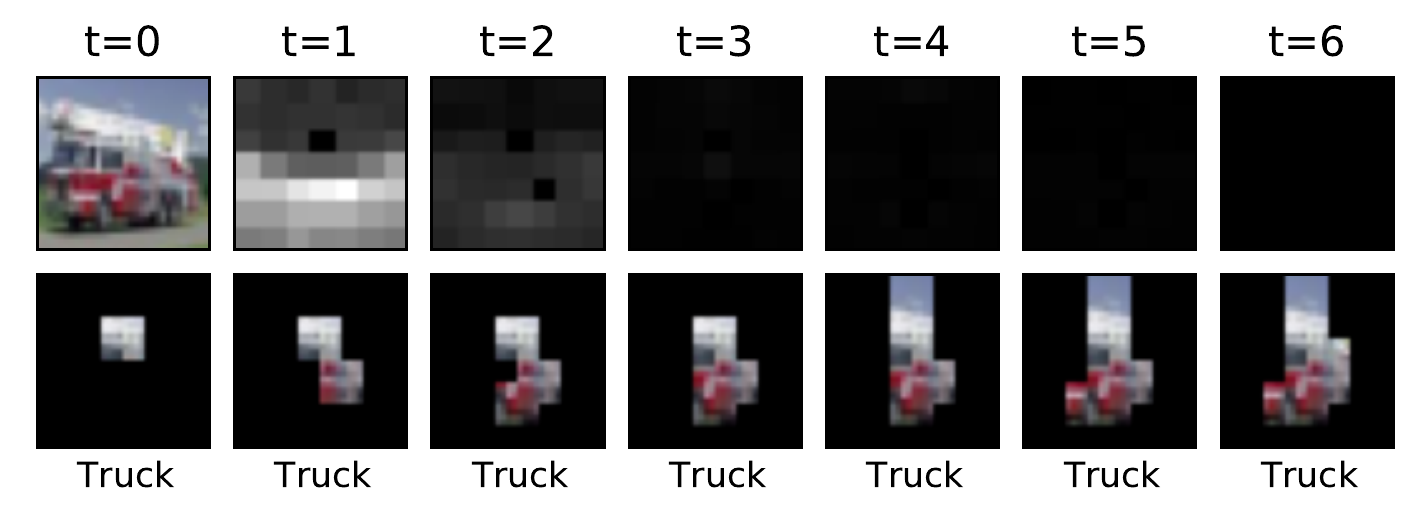}
    \end{minipage}
    \vfill
    \begin{minipage}[c]{0.82\linewidth}
    \includegraphics[width = \textwidth]{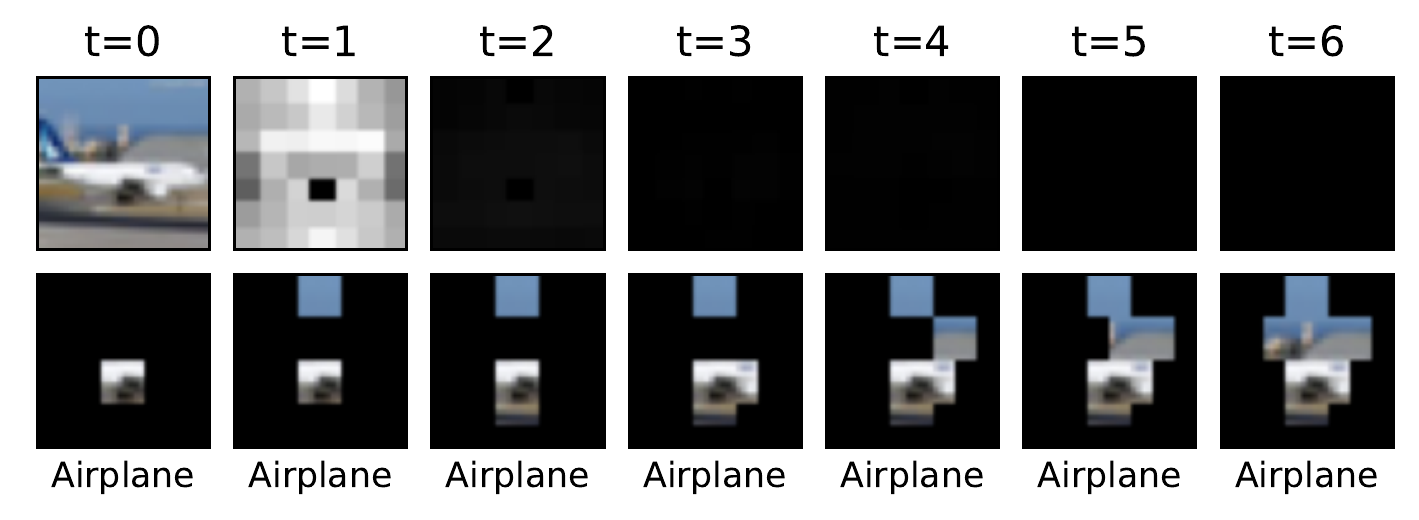}
    \end{minipage}
    \vfill
    \begin{minipage}[c]{0.82\linewidth}
    \includegraphics[width = \textwidth]{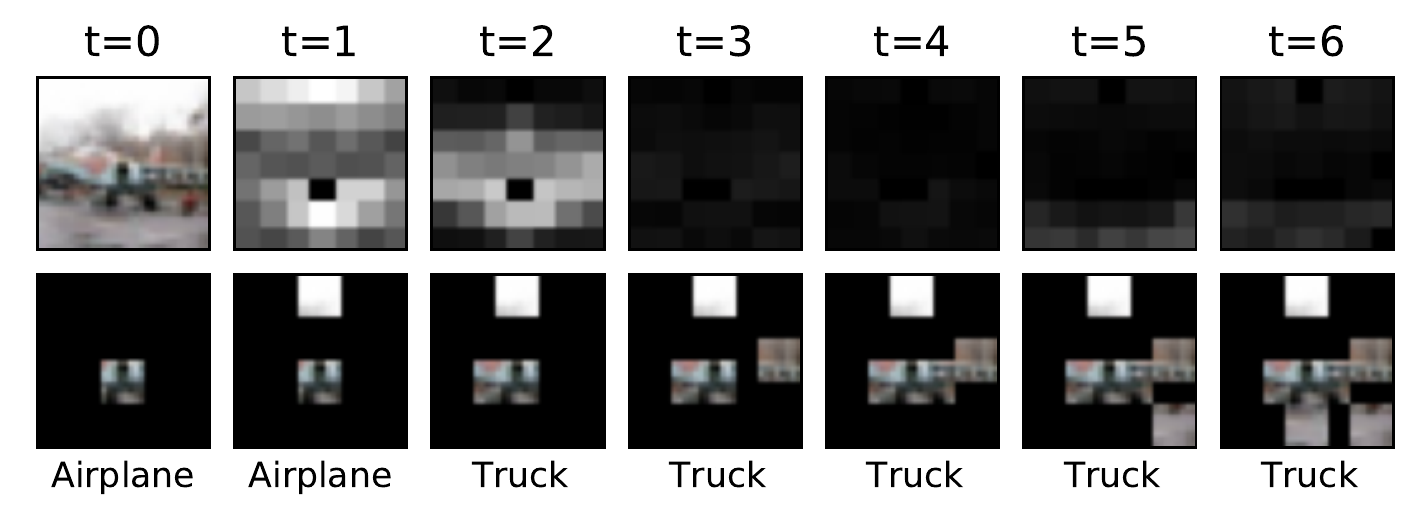}
    \end{minipage}
    \caption{\textit{Additional visual results on randomly chosen images from CIFAR-10 dataset}.}
    \label{fig:viz2}
\end{figure}
\begin{figure}
    \centering
    \begin{minipage}[c]{0.82\linewidth}
    \includegraphics[width = \textwidth]{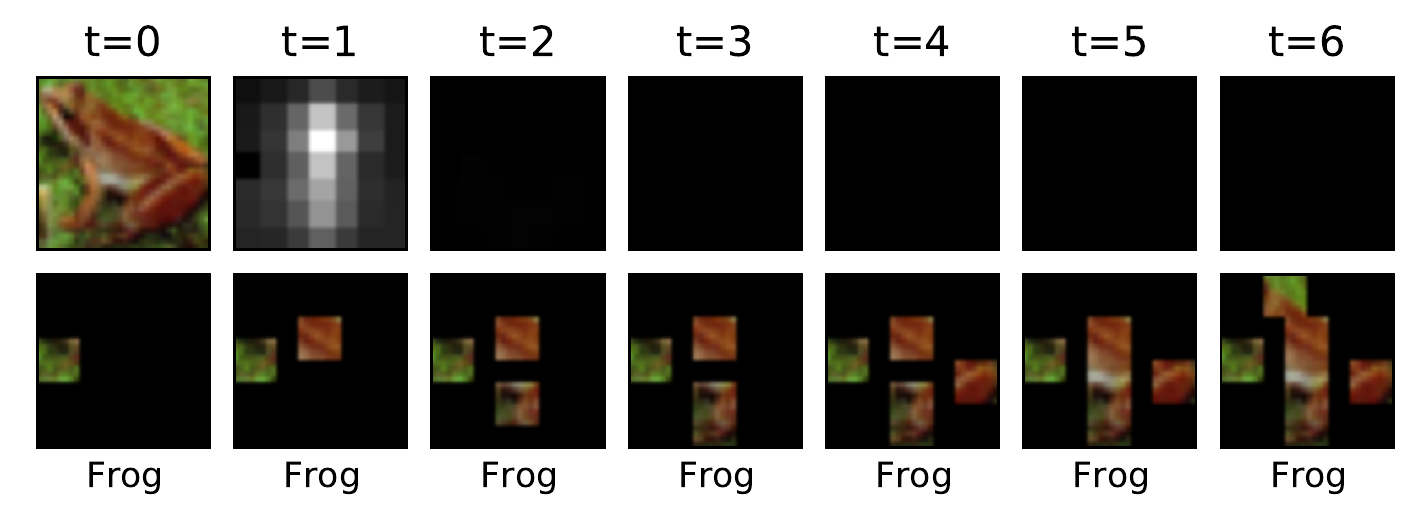}
    \end{minipage}
    \vfill
    \begin{minipage}[c]{0.82\linewidth}
    \includegraphics[width = \textwidth]{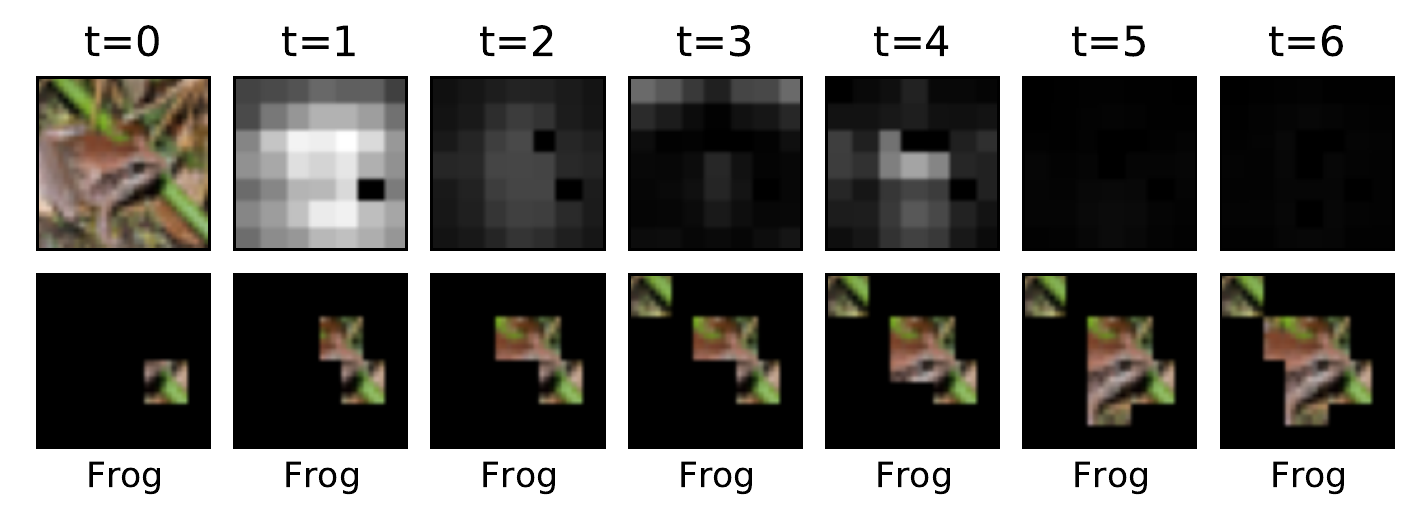}
    \end{minipage}
    \vfill
    \begin{minipage}[c]{0.82\linewidth}
    \includegraphics[width = \textwidth]{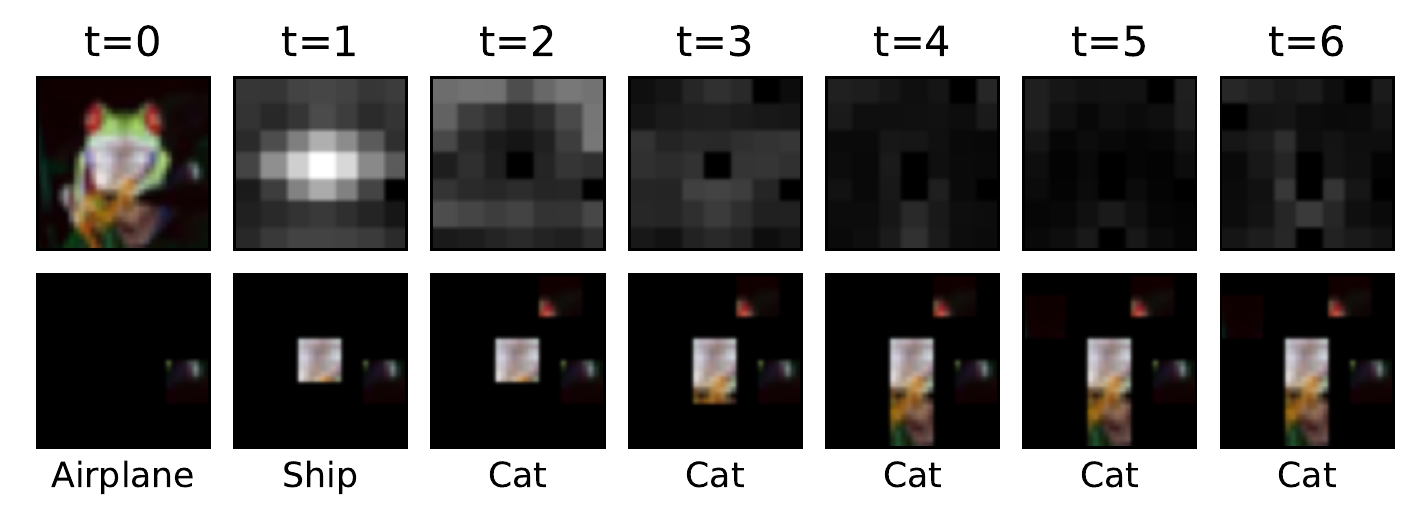}
    \end{minipage}
    \vfill
    \begin{minipage}[c]{0.82\linewidth}
    \includegraphics[width = \textwidth]{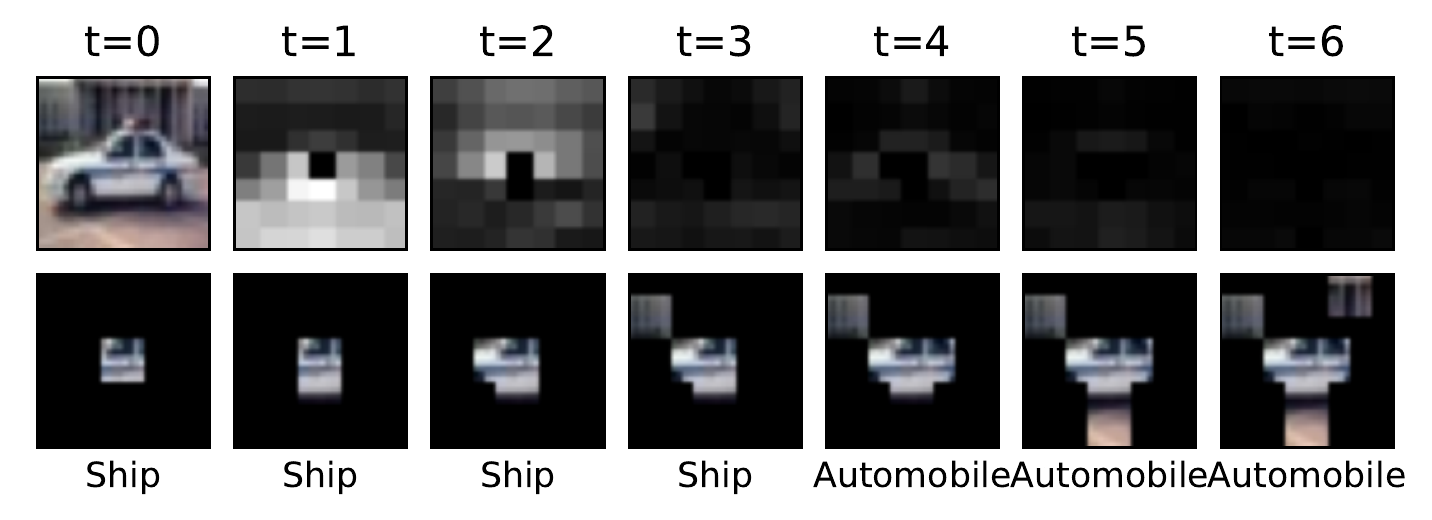}
    \end{minipage}
    \vfill
    \begin{minipage}[c]{0.82\linewidth}
    \includegraphics[width = \textwidth]{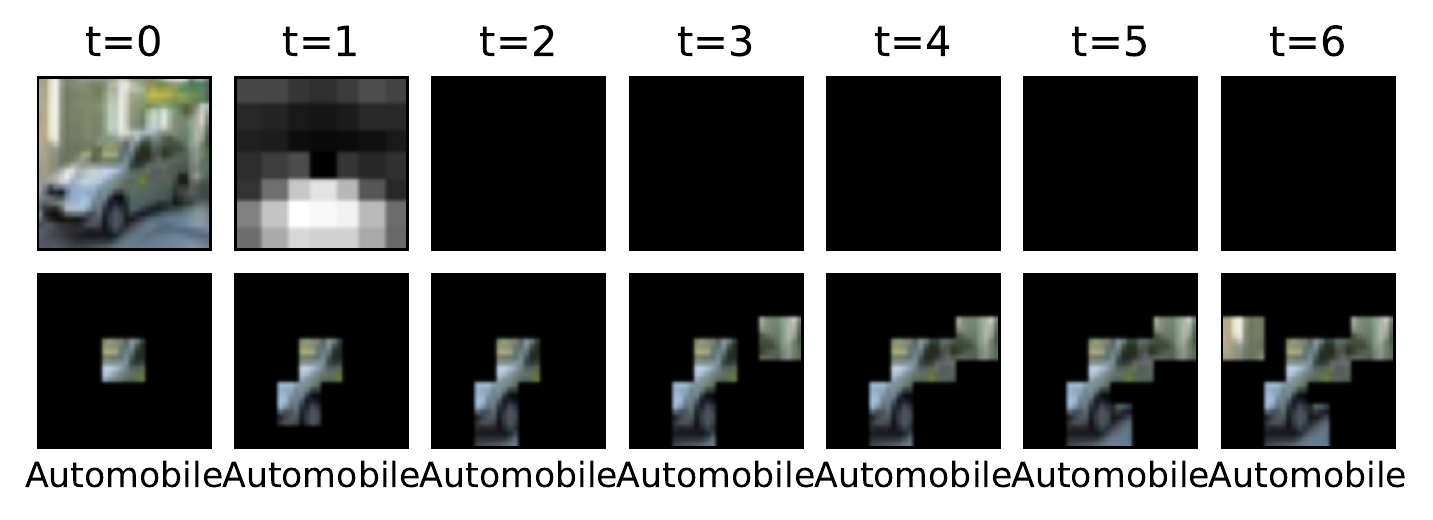}
    \end{minipage}
    \caption{\textit{Additional visual results on randomly chosen images from CIFAR-10 dataset}.}
    \label{fig:viz3}
\end{figure}
\begin{figure}
    \begin{minipage}[c]{\linewidth}
    \includegraphics[width = \textwidth]{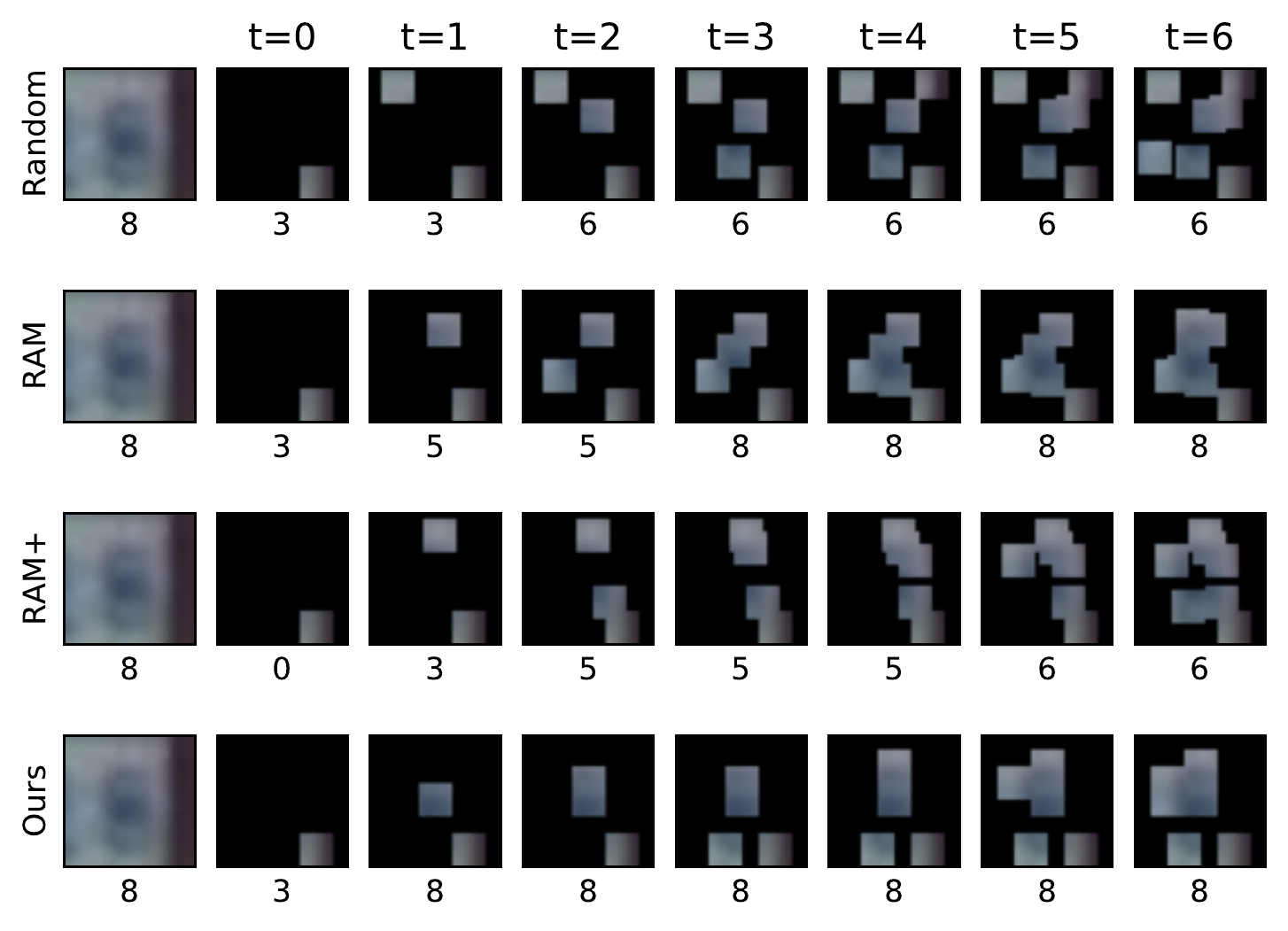}
    \end{minipage}
    \vfill
    \begin{minipage}[c]{\linewidth}
    \includegraphics[width = \textwidth]{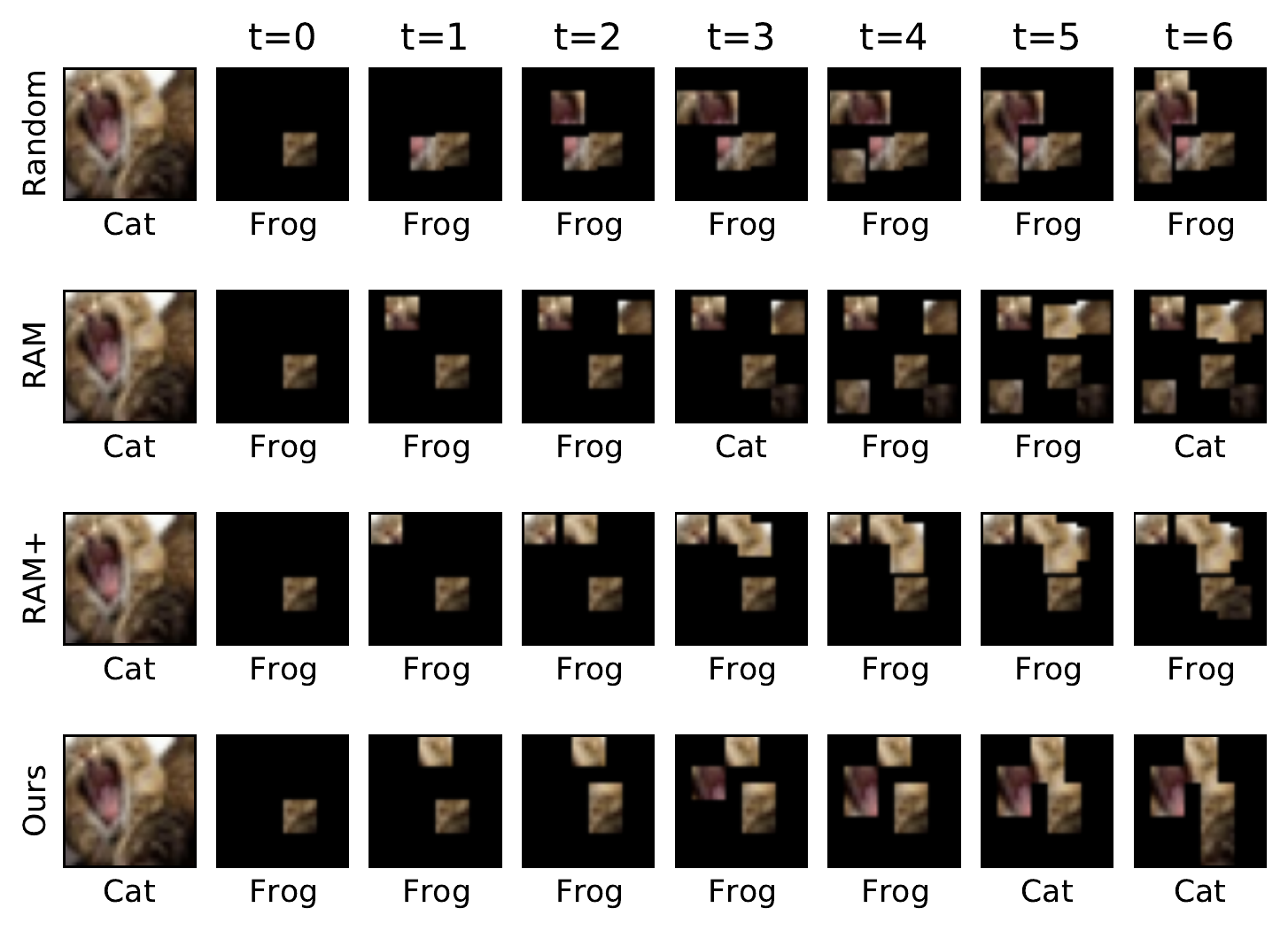}
    \end{minipage}
    \caption{Visualization of glimpses observed by different methods on images from (top) SVHN (bottom) CIFAR-10. Refer to section \ref{sec:viz} for explanation.}
    \label{fig:viz_all1}
\end{figure}
\begin{figure}
    \begin{minipage}[c]{\linewidth}
    \includegraphics[width = \textwidth]{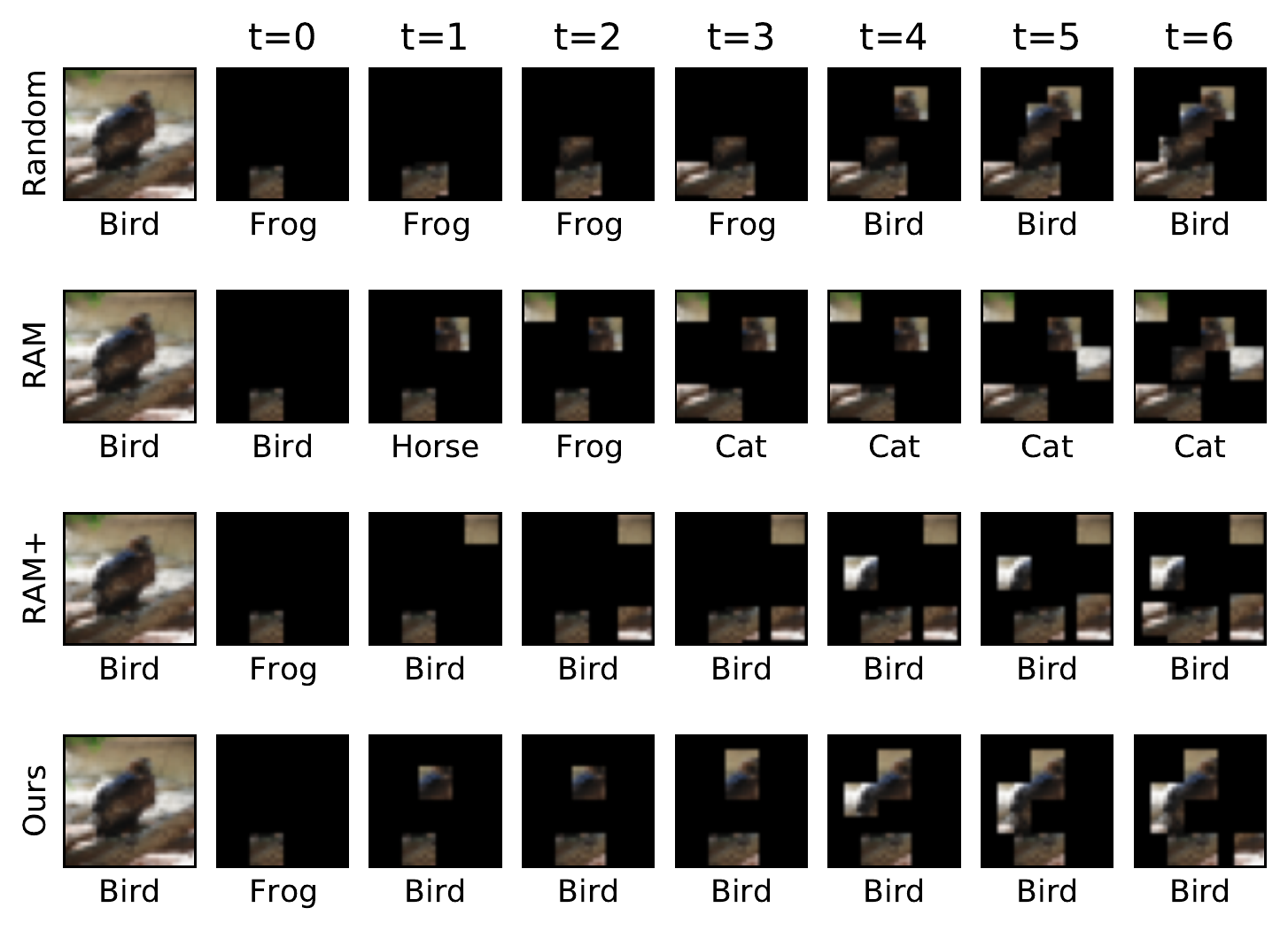}
    \end{minipage}
    \vfill
    \begin{minipage}[c]{\linewidth}
    \includegraphics[width = \textwidth]{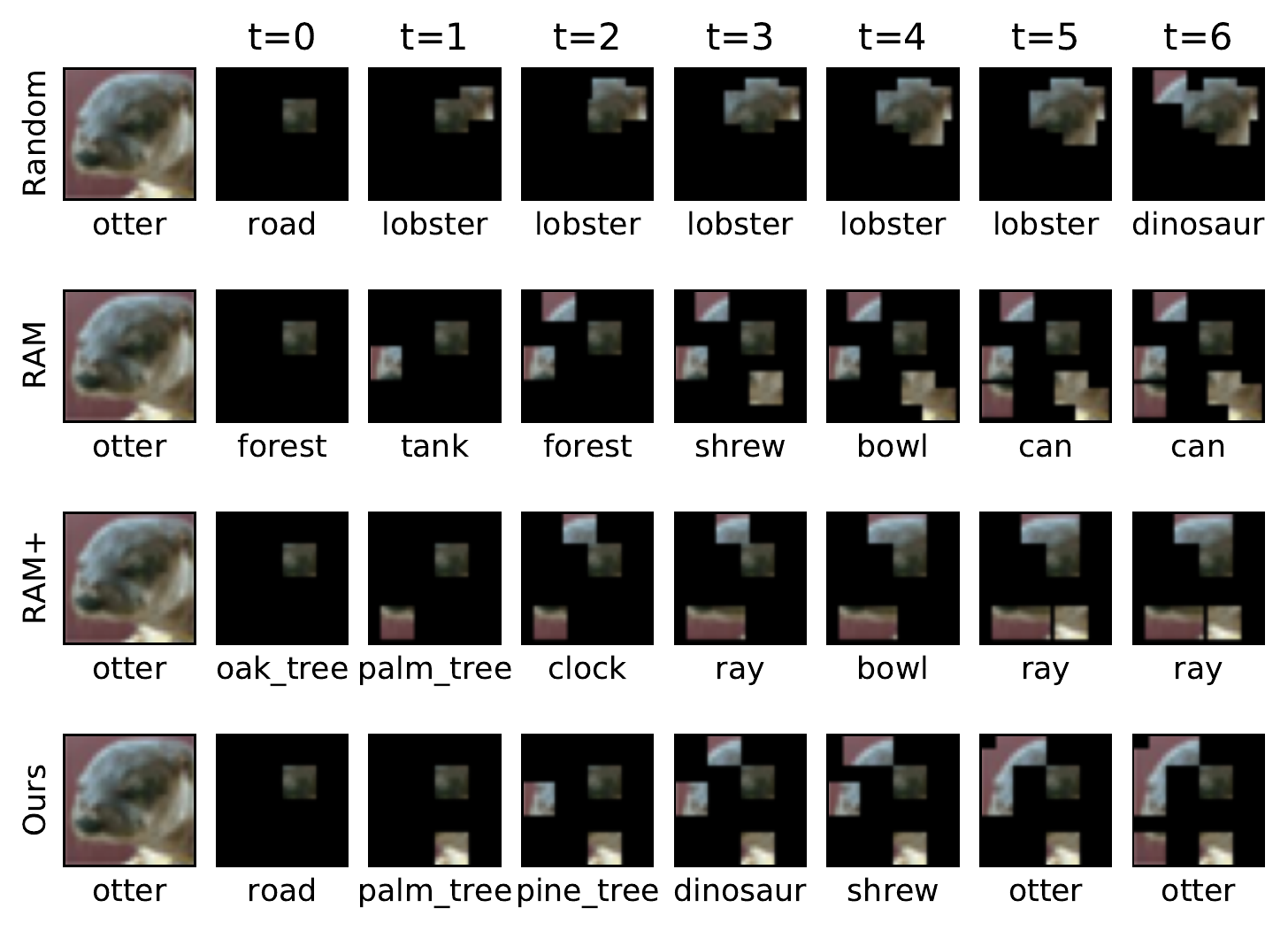}
    \end{minipage}
    \caption{Visualization of glimpses observed by different methods on images from (top) CINIC-10 (bottom) CIFAR-100. Refer to section \ref{sec:viz} for explanation.}
    \label{fig:viz_all2}
\end{figure}
\begin{figure}
    \begin{minipage}[c]{\linewidth}
    \includegraphics[width = \textwidth]{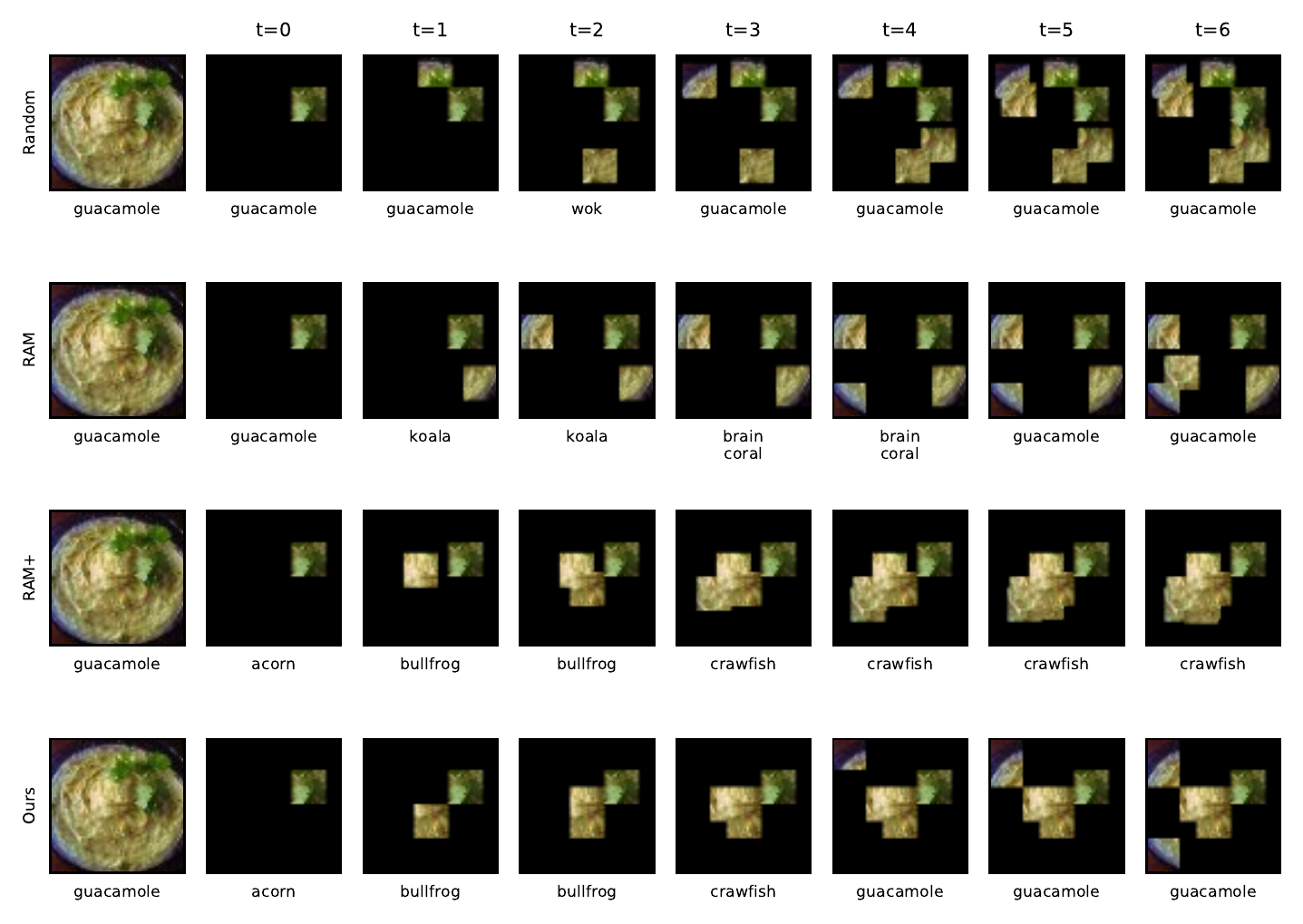}
    \end{minipage}
    \vfill
    \begin{minipage}[c]{\linewidth}
    \includegraphics[width = \textwidth]{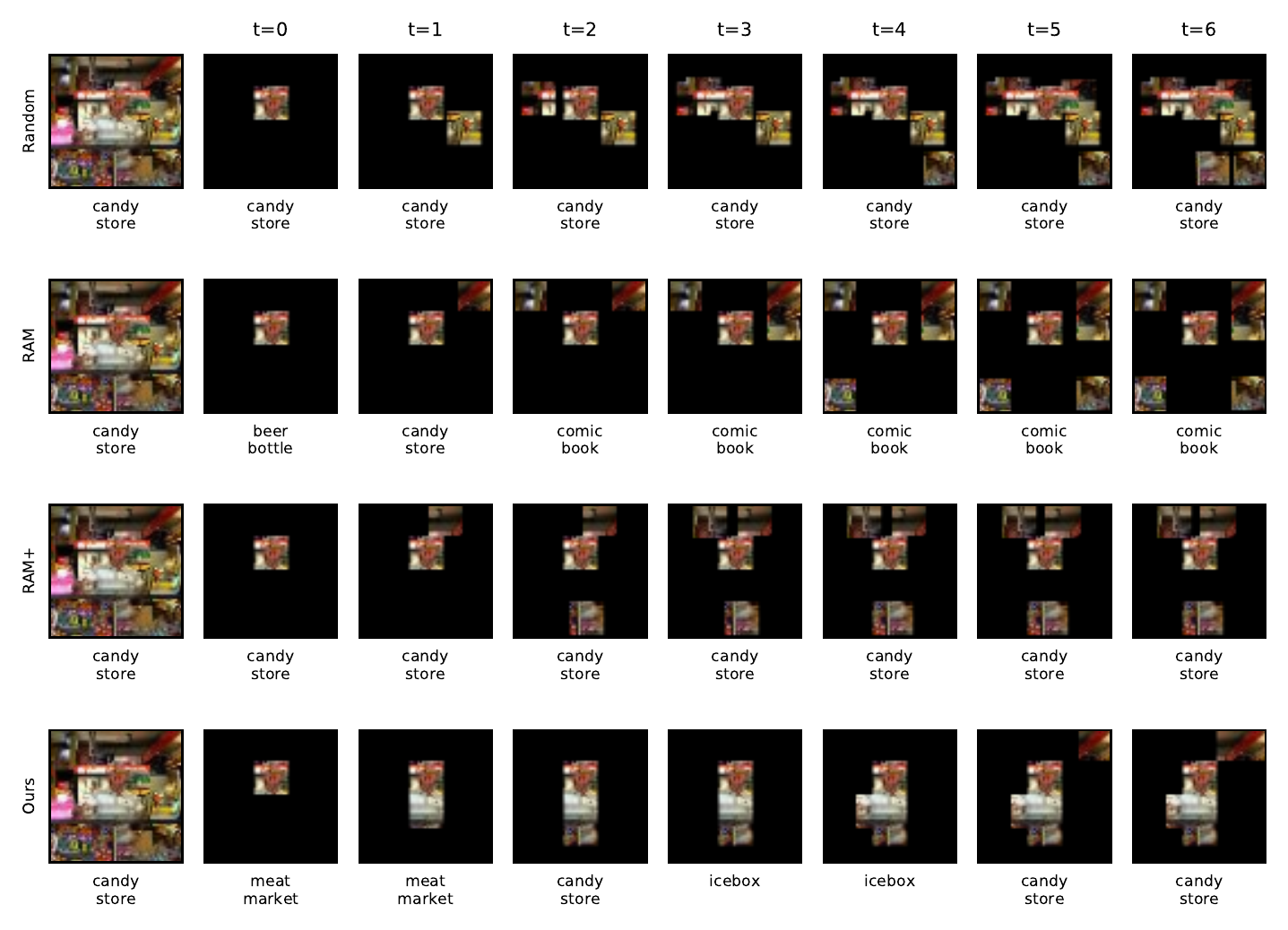}
    \end{minipage}
    \caption{Visualization of glimpses observed by different methods on two images from TinyImageNet. Refer to section \ref{sec:viz} for explanation.}
    \label{fig:viz_all3}
\end{figure}
\subsection{Visualization of $q(z|h_t)$ estimated with and without normalizing flows} Synthesizing a feature map of a complete image using only partial observations is an ill-posed problem with many solutions. We use normalizing flows in the encoder $S$ to capture a complex multimodal posterior $q(z|h_t)$ that helps the decoder predict multiple plausible feature maps for a complete image. We perform an ablation study to analyze the importance of using normalizing flows. In Figure \ref{fig:tsne}, we present TSNE \cite{van2008visualizing} projections of $q(z|h_t)$ estimated with and without the use of normalizing flows. We can observe that the normalizing flows capture a complex multimodal posterior. Capturing multiple modes leads to a more accurate estimation of EIG and consequently higher performance.

\subsection{Self-attention in $R$} Here, we experiment with self-attention in $R$. We adapt the method proposed in \cite{lin2017structured} for our problem. Figure \ref{fig:attn}(a) compares the performance of our model with and without self-attention. We observe only a marginal performance improvement due to self-attention, perhaps because the sequence of seven glimpses is relatively short. However, self-attention may achieve higher accuracy if the model is trained for longer sequences. Nevertheless, the results of this preliminary experiment are favorable, suggesting that exploring self-attention in $R$ is a promising direction for future works.

\subsection{Confusion matrix} In Figure \ref{fig:conf_mat}, we display the confusion matrix of various methods for the CIFAR-10 dataset at t=6. We also show the average image for each class. We observe that the models are able to discern classes with similar color schemes, suggesting that they rely on complex high level features instead of simple low level features such as pixel color. Furthermore, the Random baseline, the RAM, and the RAM+ over-represent category `frog', confusing it with many other categories. Our method does not suffer from this phenomenon.

\subsection{Additional visualization}
\label{sec:viz}
In Figure \ref{fig:viz1}-\ref{fig:viz3}, we visualize the $EIG$ maps and the glimpses observed by our model on CIFAR-10 images. The top rows in each plot show the entire image and the $EIG$ maps for $t=1$ to 6. The bottom rows in each plot show glimpses attended by our model. The model observes the first glimpse at a random location. Our model observes a glimpse of size $8\times 8$. The glimpses overlap with the stride of 4, resulting in a $7\times 7$ grid of glimpses. The EIG maps are of size $7 \times 7$ and are upsampled for the display. We display the entire image for reference; our model never observes the whole image. Additionally, in Figure \ref{fig:viz_all1}-\ref{fig:viz_all3}, we visualize a series of glimpses selected by various models for one image from all datasets. A complete image is shown for reference. Compared methods do not observe the complete image. Ground Truth label is shown in the first column, below the complete image. Labels predicted after observing various glimpses are shown in the columns marked with time $t$.

\section{Normalizing Flows}
Normalizing Flows map a Gaussian distribution to a complex multi-modal distribution using a series of differentiable and invertible functions. We use conditional normalizing flows that map samples from $q(z_0|h_t)$ (Gaussian distribution) to $q(z_{N}|h_t)$ (a complex distribution) using a series of invertible functions $\{f_0, f_1, \dots, f_{N-1}\}$ conditioned on $h_t$.
\begin{align}
    z_N &= f_{N-1} \circ \dots \circ f_1 \circ f_0(z_0); \text{   where   } z_0 \sim q(z_0|h_t)
\end{align}
The relation between $q(z_0|h_t)$ and $q(z_{N}|h_t)$ is established using a change of variable formula.
\begin{align}
    q(z_N|h_t) &= q(z_0|h_t) \prod_{n=0}^{N-1} |det(J_{f_n})|^{-1}
\end{align}
Where $J_f$ is a Jacobian of $f$. We use $N = 3\times n_s$, where values of $n_s$ are given in Table \ref{tab:arch}. For all $n\in\{0,3,\dots, N-3\}$, we define $f_n$ $f_{n+1}$ and $f_{n+2}$ using ActNorm \cite{kingma2018glow}, Flip \cite{dinh2016density} and Neural Spline Flows \cite{durkan2019neural}, respectively. Below, we provide a brief introduction on these three functions. To reduce clutter, we refer to $z_n$ as $z$ and $f_n$ as $f$. \\

\minisection{ActNorm \cite{kingma2018glow}.} An ActNorm layer performs an element-wise scaling and shifting of $z$.
\begin{align}
    f(z) = s \odot z + b
\end{align}
The scale parameter $s$ and the shift parameter $b$ are predicted by a neural network using $h_t$. We adapt a data-dependent initialization scheme for this network \cite{kingma2018glow}. Specifically, we initialize the above neural network such that the predicted $s$ and $b$ yield $f(z)$ with unit variance and zero mean for the first batch. We can compute $|det(J_f)| = \prod_i |s(i)|$.\\

\minisection{Flip \cite{dinh2016density}.} A Flip layer simply reverses elements of $z$, i.e. $f(z)=reversed(z)$. The Jacobian determinant $|det(J_f)|=1$.\\ 

\minisection{Neural Spline Flows (NSF) \cite{durkan2019neural}.} An NSF performs element-wise transformations on $z$. Specifically, it transforms an element $z(i)$ using a monotonic piece-wise spline function $f_i$ that is defined using parameters $\{\{w^{(k)}\}^K_{k=0} , \{v^{(k)}\}^K_{k=0}, \{\delta^{(k)}\}^{K-1}_{k=1}\}$. Refer to \cite{durkan2019neural} for more details on these parameters. In auto-regressive NSF, a neural network predicts the parameters of $f_i$ from the elements $z(j<i)$ and $h_t$. Then, we find a specific $k$ for which $w^{(k)} <z(i)<w^{(k+1)}$ and transform $z(i)$ as follows.
\begin{align}
  s &= \frac{v^{(k+1)}-v^{(k)}}{w^{(k+1)}-w^{(k)}}\\
  \xi(z(i)) &= \frac{z(i) - w^{(k)}}{w^{(k+1)}-w^{(k)}}\\
  f_i(\xi) &= v^{(k)}+\frac{(v^{(k+1)}-v^{(k)})[s\xi^2+\delta^{(k)}\xi(1-\xi)]}{s+[\delta^{(k+1)}+\delta^{(k)}-2s]\xi(1-\xi)}
\end{align}
The derivative of $f_i$ is defined as follows.
\begin{align}
    \frac{d f_i}{d z(i)} = \frac{s^2[\delta^{(k+1)}\xi^2+2s\xi(1-\xi)+\delta^{(k)}(1-\xi)^2]}{[s+[\delta^{(k+1)}+\delta^{(k)}-2s]\xi(1-\xi)]^2}
\end{align}
As NSF applies element-wise monotonic functions, $|det(J_f)| = \prod_i |\frac{d f_i}{d z(i)}|$.
\end{document}

%% file: math_commands.tex
\usepackage{amsmath,amsfonts,bm}

\def\eqref#1{equation~\ref{#1}}

\def\1{\bm{1}}

\DeclareMathAlphabet{\mathsfit}{\encodingdefault}{\sfdefault}{m}{sl}
\SetMathAlphabet{\mathsfit}{bold}{\encodingdefault}{\sfdefault}{bx}{n}

\newcommand{\KL}{D_{\mathrm{KL}}}

\DeclareMathOperator*{\argmax}{arg\,max}